%% file: main.tex
\documentclass[a4paper,12pt]{article}
\usepackage{times}
\usepackage[nottoc,notlof,notlot]{tocbibind} 

\usepackage[most]{tcolorbox} 

\input{math_commands.tex}

\usepackage[utf8]{inputenc}      
\usepackage[T1]{fontenc}         
\usepackage{hyperref}            
\usepackage{url}                 
\usepackage{natbib}

\usepackage[margin=3cm]{geometry} 
\usepackage{graphicx} 

\usepackage[labelfont=bf]{caption} 

\usepackage{amsmath}  
\usepackage{amssymb}
\usepackage{booktabs} 
\usepackage{longtable}
\usepackage{pifont}
\usepackage{setspace} 
\usepackage{subfigure}
\usepackage{svg}      

\usepackage[dvipsnames]{xcolor}

\usepackage{enumitem}
\usepackage{hwemoji}

\usepackage{verbatim}

\definecolor{cmap_blue}{rgb}{0.5664744329104193, 0.7687043444828915, 0.8685121107266437}
\definecolor{cmap_red}{rgb}{0.6922722029988466, 0.09227220299884659, 0.1677047289504037}

\newcommand{\cmark}{\ding{51}}
\newcommand{\xmark}{\ding{55}}

\usepackage[capitalize,noabbrev]{cleveref} 

\hypersetup{
    bookmarksopen=false,
    unicode=false,          
    pdftoolbar=true,        
    pdfmenubar=true,        
    pdffitwindow=false,     
    pdfstartview={FitH},    
    pdftitle={Quantifying construct validity in large language model evaluations}, 
    pdfauthor={Ryan Othniel Kearns},     
    pdfnewwindow=true,      
    colorlinks=true,       
    citebordercolor=0.74 0.85 0.91,
    linkbordercolor=0.69 0.09 0.17,
    urlbordercolor=0.07 0.29 0.52,        
    filecolor=black,      
    linkcolor=black,           
    urlcolor=black,
    menucolor=black,
    citecolor=black,
}

\usepackage{fancyhdr}
\newcommand{\HRule}{\rule{\linewidth}{0.5mm}}

\begin{document}

\begin{titlepage}

\begin{center}


\textsc{\LARGE University of Oxford}\\[1.5cm]


\HRule \\[0.4cm]
{\large \bfseries Quantifying construct validity in large language model evaluations}\\[0.4cm]

\HRule \\[1.5cm]

{\large Ryan Othniel Kearns} \\[0.6cm]

St Catherine's College \\[0.3cm]

Supervisor: Prof. Adam Mahdi

\vfill

Thesis submitted in partial fulfilment of the requirement for the degree of MSc in Social Data Science at the Oxford Internet Institute at the University of Oxford.

\vfill

\begin{minipage}{0.49\textwidth}
\begin{flushleft} 
{Trinity Term 2025}
\end{flushleft}
\end{minipage}
\begin{minipage}{0.49\textwidth}
\begin{flushright} 
{%
  \immediate\write18{texcount -1 -sum=1,1,0,0,0,0,0 -merge -q body.tex > body-words.sum }%
  \input{body-words.sum}%
  \vspace{1em}%
 words}
\end{flushright}
\end{minipage}

\end{center}
\end{titlepage}








\thispagestyle{empty}

\begin{abstract}
    \noindent The large language model research community often reports benchmark results as if they are synonymous with general model capabilities. For example, an LLM performing well on a mathematics benchmark is taken to have a general mathematical reasoning ability. However, benchmarks can have problems that distort performance, like test set contamination and annotator error. How can we know that a benchmark is a reliable indicator of some capability that we want to measure? This question concerns the construct validity of LLM benchmarks, and it requires separating benchmark results from capabilities when we model and predict LLM performance.

    \vspace{0.5em}

    \noindent Both social scientists and computer scientists propose formal models for identifying the capabilities underlying benchmark scores. Social scientists adapt latent factor models, used for modelling human cognitive capabilities, to explain LLM benchmark performance. Computer scientists adapt neural scaling laws, used for predicting LLM performance from model size, to model capabilities. These techniques approach the problem from opposite ends, and both can produce capability estimates that are more robust than individual benchmark scores. However, neither technique is satisfactory for construct validity. The social science models ignore scaling laws, and as a result, the capabilities they extract often proxy model size. The computer science models ignore measurement error, and as a result, the capabilities they extract are both uninterpretable and overfit to the observed benchmarks.

    \vspace{0.5em}

    \noindent This thesis presents the structured capabilities model, the first model to extract interpretable and generalisable capabilities from a large collection of LLM benchmark results. I fit this model and its two alternatives on a large sample of results from the OpenLLM Leaderboard. Compared to latent factor models, the structured capabilities model performs better according to parsimonious fit indices, and it removes a problematic dominant factor that appears to be a proxy for model scale. Compared to scaling law models, the structured capabilities model fits capability estimates that generalise to out-of-distribution benchmarks with higher predictive accuracy. These improvements are possible because neither existing approach separates model scale from capabilities in the appropriate way. Model scale should inform capabilities, as in scaling laws, and these capabilities should inform observed results up to measurement error, as in latent factor models. In combining these two insights, structured capabilities demonstrate better explanatory and predictive power for quantifying construct validity in LLM evaluations.
\end{abstract}

\clearpage
\thispagestyle{empty}

\tableofcontents

\thispagestyle{empty}
\listoffigures
\listoftables

\clearpage

\onehalfspacing
\fancyhead[L]{\leftmark}
\fancyhead[R]{\rightmark}
\pagenumbering{arabic} 

\input{body}

\clearpage

\singlespacing

\fancyhead[L]{REFERENCES}
\fancyhead[R]{}


\bibliography{references}
\bibliographystyle{bibstyle}

\clearpage

\fancyhead[L]{APPENDIX}

\appendix
\crefalias{section}{appendix}

\input{appendices/benchmarks}
\clearpage
\input{appendices/mathematical_notation}
\clearpage
\input{appendices/bbh_task_descriptions}
\clearpage
\input{appendices/derivation}
\clearpage
\input{appendices/bbh_logistic_correlations}

\end{document}

%% file: math_commands.tex
\usepackage{amsmath,amsfonts,bm}

\def\eqref#1{equation~\ref{#1}}

\def\1{\bm{1}}

\def\ra{{\textnormal{a}}}

\def\rx{{\textnormal{x}}}

\def\rva{{\mathbf{a}}}

\def\erva{{\textnormal{a}}}

\def\ervx{{\textnormal{x}}}

\def\rmA{{\mathbf{A}}}

\def\vmu{{\bm{\mu}}}
\def\vtheta{{\bm{\theta}}}
\def\va{{\bm{a}}}
\def\vb{{\bm{b}}}

\def\ve{{\bm{e}}}
\def\vf{{\bm{f}}}

\def\vl{{\bm{l}}}

\def\vn{{\bm{n}}}

\def\vs{{\bm{s}}}
\def\vt{{\bm{t}}}

\def\vv{{\bm{v}}}
\def\vw{{\bm{w}}}
\def\vx{{\bm{x}}}

\def\eva{{a}}

\def\evf{{f}}

\def\evx{{x}}

\def\mA{{\bm{A}}}
\def\mB{{\bm{B}}}

\def\mE{{\bm{E}}}
\def\mF{{\bm{F}}}

\def\mH{{\bm{H}}}
\def\mI{{\bm{I}}}
\def\mJ{{\bm{J}}}

\def\mL{{\bm{L}}}

\def\mT{{\bm{T}}}

\def\mW{{\bm{W}}}
\def\mX{{\bm{X}}}

\def\mSigma{{\bm{\Sigma}}}

\DeclareMathAlphabet{\mathsfit}{\encodingdefault}{\sfdefault}{m}{sl}
\SetMathAlphabet{\mathsfit}{bold}{\encodingdefault}{\sfdefault}{bx}{n}
\newcommand{\tens}[1]{\bm{\mathsfit{#1}}}
\def\tA{{\tens{A}}}

\def\tX{{\tens{X}}}

\def\gG{{\mathcal{G}}}

\def\sA{{\mathbb{A}}}
\def\sB{{\mathbb{B}}}

\def\sR{{\mathbb{R}}}
\def\sS{{\mathbb{S}}}

\def\emA{{A}}
\def\emB{{B}}

\def\emL{{L}}

\newcommand{\etens}[1]{\mathsfit{#1}}

\def\etA{{\etens{A}}}

\newcommand{\E}{\mathbb{E}}

\newcommand{\R}{\mathbb{R}}

\newcommand{\KL}{D_{\mathrm{KL}}}
\newcommand{\Var}{\mathrm{Var}}

\newcommand{\Cov}{\mathrm{Cov}}

\newcommand{\normltwo}{L^2}
\newcommand{\normlp}{L^p}

\newcommand{\parents}{Pa} 

%% file: body.tex
\input{sections/1_introduction}
\input{sections/2_background}
\input{sections/3_methods}
\input{sections/4_results}
\input{sections/5_discussion}
\input{sections/6_conclusion}

%% file: sections/1_introduction.tex
\section{Introduction}

\subsection{Large language models: A ``superhuman'' intelligence?}

When we claim that large language models (LLMs) possess `scientific knowledge,' or `mathematical reasoning abilities,' how do we justify these claims?\footnote{In this thesis, I intend `single quotations' to be read as scare quotes. ``Double quotations'' will indicate direct quotation from a cited source. I will also use \textbf{boldface} to signify definitions for terms and \emph{italics} for general emphasis.} One method is anecdotal: individuals can ask an LLM some science or mathematics questions and come away with an impression of the model's abilities. This justification is neither very scalable nor scientific. A more robust method, and the standard method for modern AI systems, is to use a \textbf{benchmark dataset}: a curated selection of questions and answers that gives an objective measure of an LLM's performance. Popular benchmarks measure LLM performance on mathematics problems~\citep{cobbe_training_2021, hendrycksMATH, glazer2024frontiermathbenchmarkevaluatingadvanced}, scientific questions~\citep{rein2024gpqa, wang2025mmlu}, commonsense language puzzles~\citep{talmor_commonsenseqa_2019, sakaguchi_winogrande_2021}, and code generation tasks~\citep{chen2021evaluatinglargelanguagemodels, jimenez2024swebench}. On many benchmarks, LLMs achieve ``superhuman'' performance, eclipsing the average human accuracy on these questions~\citep{wangSuperglueStickierBenchmark2019a}. In fact, LLMs perform at superhuman levels on so many benchmarks that some researchers worry we are running out of difficult tasks to benchmark. Others claim that LLMs demonstrate the ``sparks of artificial general intelligence''~\citep{bubeck_sparks_2023}, and recently published benchmarks feature dramatic names, like ``Humanity's Last Exam''~\citep{humanitys_last_exam}.

Benchmarks play an essential role in orienting progress in AI~\citep{liang_holistic_2023}. Most benchmarks are public artefacts, making LLM performance reproducible and independently verifiable. Their clear, objective metrics signal current model performance to policymakers and unsolved gaps to researchers~\citep{pottsReliableCharacterizationsNLP2021}. For the most part, AI researchers report benchmark results as if they are synonymous with model capabilities: if a model performs well on a benchmark of mathematics problems, it is taken to have a mathematical problem-solving or reasoning capability~\citep{kojima2022large, wei_emergent_2022, srivastava2023beyond}.

\subsection{The evaluation crisis}

Unfortunately, benchmarks often give unreliable accounts of LLM capabilities. Datasets can suffer contamination, which occurs when a model is exposed to benchmark data during training. Models can achieve unrealistic benchmark results by memorising data through such exposure~\citep{zhouDontMakeYour2023, yang_rethinking_2023}. Popular benchmarks can be gamified by influential model providers. For example, \citet{singh2025leaderboardillusion} show that selective reporting of commercial model results has distorted model comparisons on the Chatbot Arena leaderboard~\citep{chiang2024chatbotarenaopenplatform}. Benchmark scores can also give erroneous performance estimates simply due to human error when labelling the correct answers~\citep{kocijanDefeatWinogradSchema2023}.
Even different choices of evaluation metrics can lead to conflicting accounts of whether LLM capabilities are ``emergent'' in certain benchmark results~\citep{schaeffer_are_2023, burnell_rethink_2023}.
Finally, even when datasets are labelled correctly, reported fairly, and withheld from contamination, LLM performance can challenge our expectations. Several studies show that seemingly meaningless alterations to questions, like paraphrases, or changing names in arithmetic problems, can drastically shift benchmark performance \citep{jiang_peek_2024, mizrahi-etal-2024-state, mirzadeh_gsm-symbolic_2025}.

This situation is paradoxical. LLMs appear ``superhuman'' across many tasks, and there is strong anecdotal evidence that this technology represents a powerful and compelling advancement in AI. Yet our ability to benchmark these systems seems to suffer from a growing list of failures. Some researchers call the current moment an ``evaluation crisis'' for LLMs~\citep{karpathy_my_2025}.

An evaluation crisis affects not just the AI research community, but also policymakers and society at large. Misplaced estimates of these capabilities can cause medical harms if AI systems are deployed prematurely~\citep{bean2025clinicalknowledgellmsdoes} and can obscure the labour market impact of this transformative technology, hindering our possible economic and policy responses~\citep{freyGenerativeAIFuture2023}. Accurate capability measurements play an essential role in the fairness and governance of AI systems~\citep{jacobsMeasurementFairness2021, jacobs2021measurementgovernanceresponsibleai, europeanparliamentcounciloftheeuropeanunionClassificationGeneralPurposeAI2024}.

\subsection{Can construct validity solve the evaluation crisis?}\label{sec:can_construct_validity_solve}

Given all of the problems facing benchmarks, what can be done to secure reliable estimates of LLM capabilities? According to one emerging narrative in AI research, the solution can come from approaching the problem like social scientists \citep{wallachEvaluatingGenerativeAI2024, salaudeen2025measurementmeaningvaliditycenteredframework}. To see this possibility, we must first observe that problems like contamination, gamification, and annotator error affect the reliability of benchmarks—but reliable \emph{benchmarks} are not, in themselves, our ultimate goal. We seek reliable estimates of LLM \emph{capabilities}, and our benchmarks only serve to estimate these capabilities. Users and policymakers do not really care whether a system performs well on any particular mathematical benchmark. They care whether the system can \emph{do mathematics}, as ultimately this general capability determines if an LLM can file our taxes, price our insurance, or teach mathematics to our children. If we had only one mathematical benchmark, we would indeed be in trouble, but this is not our situation. We have many benchmarks that claim to measure mathematical reasoning. Even if individual benchmarks have problems, though many separate measurements, we should be able to improve our estimation of mathematical reasoning capabilities simply by the law of large numbers.

It is in this respect that the social sciences can help LLM evaluation. Our problem is similar to one faced in disciplines like psychology. Psychology studies abstract phenomena, like personality traits, and their measurement tools (like survey questions) are often noisy and underspecified for these phenomena. As a result, psychological measurements prioritise a condition called \textbf{construct validity}, which determines whether measurements accurately reflect the phenomena (or constructs) they intend to measure \citep{cronbach_construct_1955}. \textbf{Measurement theory}, the basis for construct validity, provides psychologists with formal models to assess and iteratively refine the validity of their measurement tools~\citep{clarkConstructingValidityNew2019}. Central modern psychological concepts, like the Big Five personality traits~\citep{digmanPersonalityStructureEmergence1990}, are predicated on a literature of construct valid measurements \citep{goldberg1992development}. Several AI position papers argue that LLM benchmarks should justify and improve their connection to their own abstract phenomena, like mathematical reasoning capabilities, in a similarly rigorous way \citep{rajiAIEverythingWhole2021, wangPsychometrics2023, wallachEvaluatingGenerativeAI2024, alaa_medical_2025, salaudeen2025measurementmeaningvaliditycenteredframework}. These formal standards could provide robustness against the varied issues of contamination, gamification, and human annotator error, giving our evaluations the explanatory and predictive power we need to trust LLM capabilities in the real world.

Construct validity offers us an escape from the evaluation crisis, though for it to work, we need to apply measurement theory's formal models to our LLM benchmark results. My thesis addresses a gap in this application. Both social scientists and computer scientists have published attempts at formal models of LLM capabilities, but to date no attempt approaches the problem with the proper interdisciplinary perspective.

\subsection{The gap}

Formal models for construct validity amount to opinionated versions of dimensionality reduction: they aim to relate a large number of indicators (benchmark scores) to a smaller number of constructs (capability estimates). Separate studies from social scientists and computer scientists try their hand at this modelling challenge with existing LLM benchmark results. The social scientists \citep{burnellRevealingStructureLanguage2023, ilicEvidenceInterrelatedCognitivelike2024} attempt to directly apply \textbf{latent factor models}, the models of human cognitive capabilities already in use for construct validity \citep{fabrigarExploratoryFactorAnalysis2012, klineSpecificationIdentificationConfirmatory2016}.
In parallel, computer scientists approach this problem with their own established methods for predicting LLM performance. \citet{ruan2024observational} introduce \textbf{observational scaling laws}, adapting the \textbf{neural scaling laws} that predict LLM performance as a function of their size \citep{kaplanScalingLawsNeural2020, henighan2020scalinglawsautoregressivegenerative, hoffmannTrainingComputeoptimalLarge2022}. In place of size, \citet{ruan2024observational} use principal component analysis (PCA) to reduce a collection of benchmark scores down to ``capability vectors'' that they plug into scaling laws to predict novel task performance.

Both latent factor models \citep{burnellRevealingStructureLanguage2023, ilicEvidenceInterrelatedCognitivelike2024} and observational scaling laws \citep{ruan2024observational} succeed at translating LLM benchmark results into estimates for LLM capabilities. Unfortunately, though, both present deficiencies making them unable to demonstrate construct validity.
These deficiencies mostly arise because they ignore something the other model does correctly. Latent factor models assume that the only variables are the observed benchmark scores and the latent capabilities behind them—they ignore the model size parameters at work in scaling laws. In human populations, there are no such scaling laws, and so pure latent models tend to be appropriate. However, as we shall see, scale is such a dominant predictor of LLM performance that it causes latent LLM capability estimates to effectively collapse into proxies for the LLM's parameter count. Modelling an observable design parameter is not ideal when the intention was to capture a latent capability. Yet, observational scaling laws fare no better with their capability estimates. \citeauthor{ruan2024observational}'s approach proposes to extract capabilities from benchmark scores with PCA, then to relate these capabilities to downstream performance with scaling laws. Yet, again, since scale is the dominant predictor of LLM performance, the principal component extracted proves to be another proxy for model scale. Moreover, since PCA explains all variance in a `greedy' fashion that we will explore, it proves to be suboptimal for producing capabilities that generalise to novel tasks.

There should be a way to improve upon these initial attempts. Despite originating in separate disciplines, we will show that their mathematical models differ by only a few small assumptions.
They also fail for the same reason: neither separates an LLM's scale from its capabilities in an appropriate way.

\begin{figure}[t]
    \centering
    \includegraphics[width=0.88\linewidth]{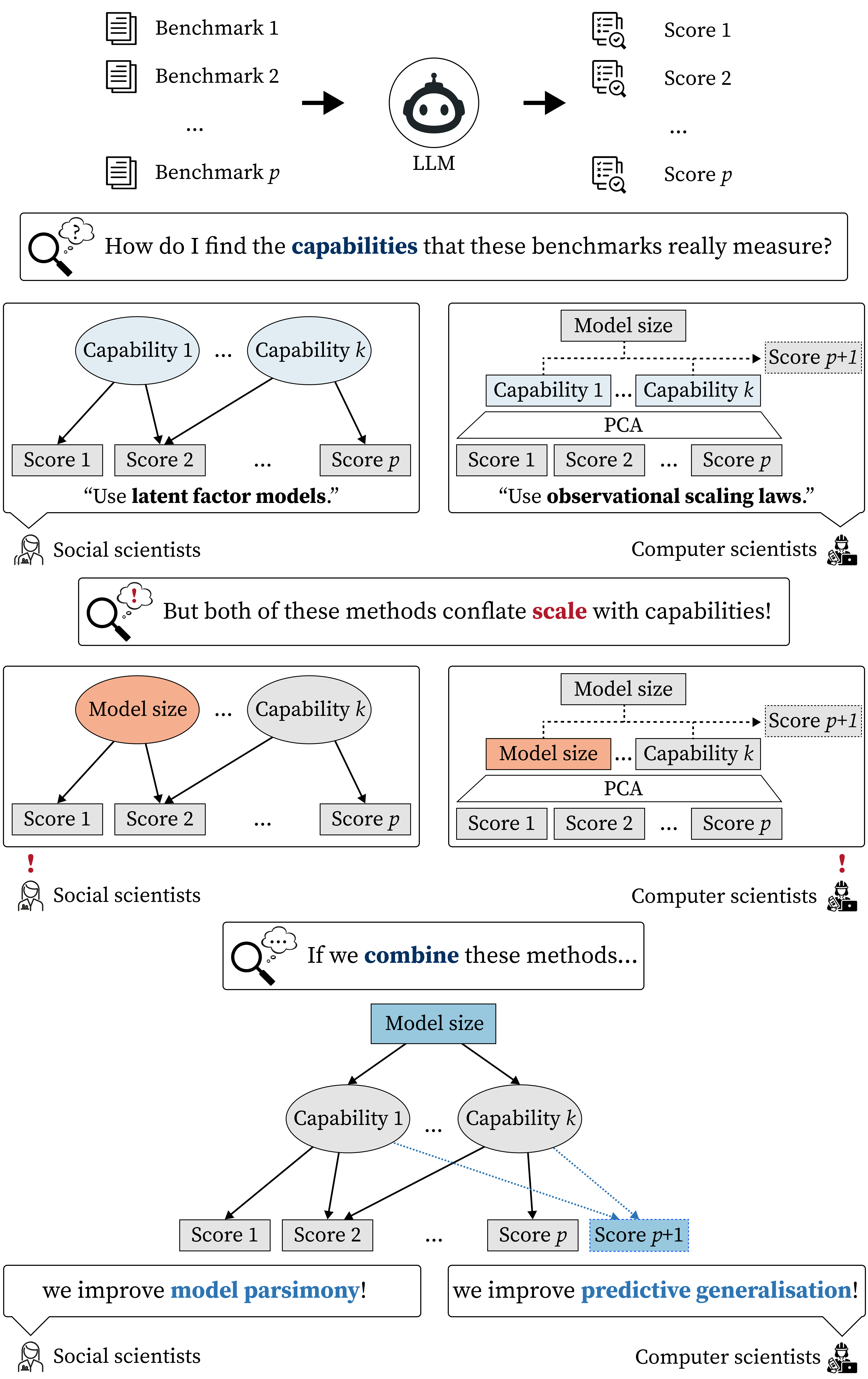}
    \caption[High-level picture of my contribution]{\textbf{High-level picture of my contribution}. My proposed model, the structured capabilities model, addresses problems in both latent factor and observational scaling law approaches. These existing approaches do not sufficiently separate model scale from capability estimates, which harms both their explanatory and their predictive power.}
    \label{fig:model_comparisons}
\end{figure}

\subsection{My contribution}

This thesis presents the \textbf{structured capabilities model}, the first model to extract interpretable and generalisable capabilities from a large collection of LLM benchmark results. The structured capabilities model combines latent factor models with observational scaling laws, avoiding the deficiencies that both approaches present on their own.

In fitting this model to benchmark results from the OpenLLM Leaderboard, I observe improvements that are communicable to both social scientists and computer scientists interested in modelling LLM capabilities. To social scientists interested in formal models of LLM capabilities, I demonstrate a more parsimonious model, with better fit statistics, than the common latent factor model for extracting capabilities from benchmark results. To computer scientists interested in predicting future model performance, I demonstrate improved out-of-distribution generalisation for benchmark score prediction compared to existing approaches based on observational scaling laws. My high-level approach and contributions are depicted in \cref{fig:model_comparisons}.



\clearpage

%% file: sections/2_background.tex
\section{Background}


\subsection{Scaling laws and LLM capabilities}

This thesis concerns valid measurements of LLM capabilities. We observe these capabilities in an LLM's `behaviour.' Despite appearances, we should not begin by assuming that LLM behaviour aligns with our intuitions for human behaviour. We first have to understand what LLMs \emph{are}, which begins with the language modelling task they are trained to perform~\citep{mccoy_embers_2023}.

A \textbf{language model} is any model for producing natural language \citep{blankWhatAreLarge2023}.\footnote{This definition encompasses models for the human production of language that are not machine learning models, e.g., formal models used in linguistics. A \emph{natural} language, like English, contrasts with a \emph{formal} language, like a mathematical or programming language—although this definition is porous, since we know large language models are capable of producing code!} In natural language processing (NLP), a subfield of AI, language models are machine learning systems optimised to predict the next word in a sequence. This predictive task is called the \textbf{language modelling task} \citep{jurafskyNgramLanguageModels2025}. Systems succeed at the language modelling task if their predicted next words match the distributional features of real natural language. As with other machine learning applications, success is determined by splitting an empirical distribution into separate segments for training and testing, updating the model's parameters through training on the former, then reporting predictive accuracy on the latter. This evaluation setting is called the ``common task'' method~\citep{liberman-2010-obituary} or the \textbf{Common Task Framework}, and has been a convention for evaluating progress in AI systems since its adoption by DARPA in the late 1980s~\citep{donoho50YearsData2017a}.

\begin{figure}[th]
    \centering
    \includegraphics[width=\linewidth]{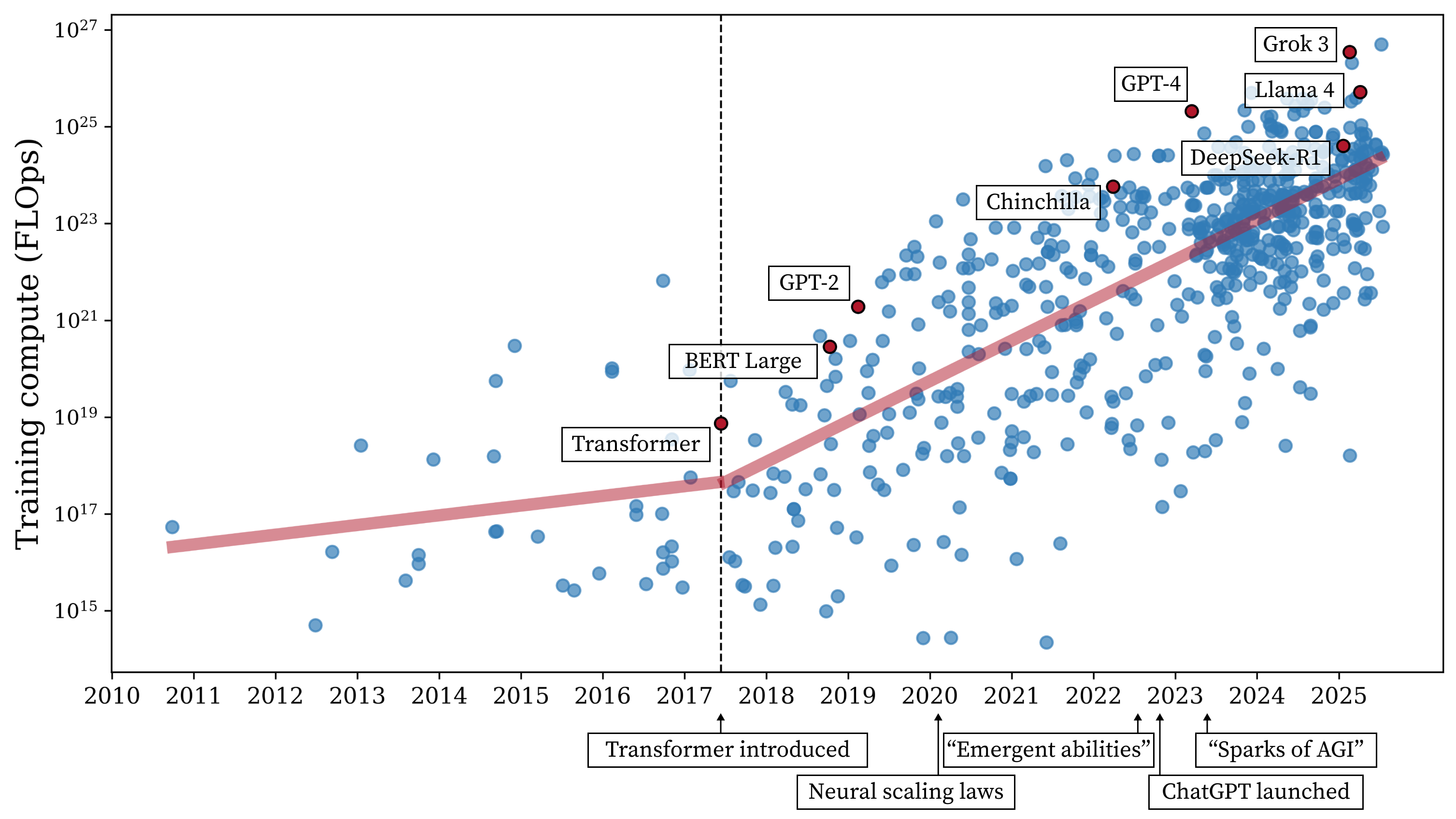}
    \caption[The scale of language model training, 2010 to 2025]{\textbf{The scale of language model training, 2010 to 2025}. The dashed black line shows the introduction of the Transformer in 2017. The solid red lines show trends for training compute before the Transformer (order of magnitude increase every 5 years) and after the Transformer (order of magnitude increase every 1.2 years). Influential LLMs are annotated on the plot with red points. Data from Epoch AI~\citep{EpochAIModels2025}.}
    \label{fig:scaling_laws}
\end{figure}

What makes a language model \textbf{large} is the number of parameters it can optimise for the language modelling task. LLMs are a contemporary phenomenon in NLP. Today, practically all language models are neural networks built with the Transformer architecture~\citep{vaswaniAttentionAllYou2017}. The largest of these models have billions and now possibly trillions of parameters. This development was unforeseen and unprecedented at the time of the Transformer's introduction, when \citet{vaswaniAttentionAllYou2017} proposed the architecture specifically for the task of machine translation.

Two factors explain the Transformer's dominance in modern NLP. First, the architecture allows training in a highly parallel fashion on consumer hardware, allowing model developers to scale extremely large networks~\citep{radfordLanguageModelsAre2019}. Second, massive natural language corpora, in the form of Internet text data from websites like Wikipedia and Reddit, are readily available for training~\citep{brown_language_2020, villalobosWillWeRun2024}. The fusion of these two resources, scalable compute and abundant data, drove an empirical breakthrough. For Transformer language models on general Internet text data, it happens that increasing the number of parameters, the training dataset size, and the amount of computation—in the correct ratios—yields smooth improvements at the language modelling task according to a power-law relationship. These findings are known as \textbf{neural scaling laws} \citep{kaplanScalingLawsNeural2020, henighan2020scalinglawsautoregressivegenerative, hoffmannTrainingComputeoptimalLarge2022}, and faith in their predictive power is largely responsible for the explosion in AI capital investment so far through the 2020s.

%
In their original formulation~\citep{kaplanScalingLawsNeural2020}, scaling laws are situated squarely in the language of the Common Task Framework. Increased scale at training time yields power-law improvements in predictive accuracy at testing time, specifically on data from the same distribution as seen in training. However, the original ambition of AI research lies in not merely \emph{test-set predictive accuracy}, but in \emph{the simulation of actually intelligent behaviour}~\citep{mccarthyProposalDartmouthSummer2006}. The distribution of Internet text is remarkably domain-general, and sequences of Internet text are not arbitrary data: they represent the outputs of human cognitive processes, like political argumentation, mathematical reasoning, or historical fact retrieval. An LLM improving at test-time predictive accuracy on this distribution is also improving at modelling these outputs of human cognitive processes. \citet{mccann2018naturallanguagedecathlonmultitask} propose that a wide range of natural language tasks beyond translation—sentiment analysis, summarisation, and inference, among others—can be reframed as language modelling tasks using question-answer pairs, and \citet{radfordLanguageModelsAre2019} show that scaling LLMs with Internet text data appears to improve performance on all of these tasks. All of a sudden, our AI systems appear to possess a broad and increasingly useful set of linguistic skills.
\citet{bommasani2022opportunitiesrisksfoundationmodels} name this new class of systems ``foundation models'' to reflect their generality of application,\footnote{It is worth a footnote's recognition that the term ``foundation model'' encompasses not just language models, but AI systems for other modalities, such as vision and speech modelling systems. \citet{henighan2020scalinglawsautoregressivegenerative}'s scaling law results apply to this general class of ``autoregressive generative models.'' For simplicity, I am presenting these results only as they apply to language models.} and provides some of the first commentary on the societal prospects of such domain-general AI systems.

The introduction of ``foundation models'' as a term represents an important step away from the Common Task Framework, and with it a focus on LLMs' test-time predictive accuracy. From here, attention shifts towards a focus on LLM \textit{capabilities} as something like genuine cognitive capabilities \citep{srivastava2023beyond}. Correspondingly, evaluation efforts shift away from distributionally-equivalent test partitions and towards more adversarial and task-specific datasets intended to encircle the limits of LLM performance~\citep{bowmanWhatWillIt2021}. Also around this time, critical discussion begins as to whether training for domain-general language modelling imbues any kind of true natural language ``understanding'' capability~\citep{bender_climbing_2020}, or whether LLMs' modelling the form of natural language represents a different capability, or no capability at all~\citep{stochastic_parrots}. These developments—the isolation of specific tasks for evaluation, and the critique of the connection between these tasks and true LLM capabilities—represent the beginning of the construct validity challenge for LLM evaluation~\citep{rajiAIEverythingWhole2021}.


The distinction between general scaling performance and specific capabilities intensifies with the claim that certain capabilities are not smoothly predictable with scale, but are instead ``emergent'' at certain large model sizes~\citep{ganguli_predictability_2022, wei_emergent_2022}. For example, smaller language models appear incapable of performing the \texttt{BIG-bench} modular arithmetic task, but accuracy on this task begins to suddenly increase after $10^{22}$ floating point operations in training~\citep[4]{wei_emergent_2022}. The same holds for \texttt{TruthfulQA}, a task involving the discernment of false statements~\citep{linTruthfulQAMeasuringHow2022}. \citet{ganguli_predictability_2022} pose this distinction as one between ``predictability'' and ``surprise'': scaling laws outline the general improvements to language modelling that researchers expect to arrive through scale, while ``emergent capabilities'' surprise even the designers of these systems.
\citeauthor{ganguli_predictability_2022} also highlight the policy implications of this finding: unpredictable capabilities pose a deployment and governance challenge, especially if the capabilities include deception or other possibly harmful qualities. These policy implications will soon be grounded in the way that LLM capabilities are—literally—written into law: Article 51 the European Union's Artificial Intelligence Act, entering into force in August 2025, specifies that
\begin{quote}
    [a] general-purpose AI model shall be classified as a general-purpose AI model with systemic risk if... it has high impact capabilities evaluated on the basis of appropriate technical tools and methodologies, including indicators and benchmarks.~\citep{europeanparliamentcounciloftheeuropeanunionClassificationGeneralPurposeAI2024}
\end{quote}

The scope of ``high impact capabilities'' is thus handed over for the technical experts to decide. We still face \citet{ganguli_predictability_2022}'s dichotomy—predictable improvements from scale on one hand, emergent capabilities on the other—and critiques in the form of \citet{bender_climbing_2020} and \citet{rajiAIEverythingWhole2021} about the validity of their evaluation techniques. It is time to address ``high impact capabilities'' using tools and techniques from the social sciences~\citep{ wallachEvaluatingGenerativeAI2024}.

\subsection{Quantifying construct validity}\label{sec:quantifying_construct_validity}

In social science language, an LLM capability is a \textbf{construct}—a theoretical quantity that cannot be directly measured (in the way a person's height can be directly measured).
Core to measurement theory in the social sciences is the separation between the measurements for constructs and the constructs themselves~\citep{cronbach_construct_1955, messickTestValidityMatter1998, salaudeen2025measurementmeaningvaliditycenteredframework}. The objects of study in the social sciences—the constructs—can be abstract. Psychology studies constructs like openness to experience; sociology studies constructs like radicalisation. A core assumption to the methodological practice of each field is that such constructs cannot be measured exhaustively or without error. Where these constructs are quantified, they are modelled to be latent variables whose values are informed by observable indicators, typically responses to surveys or behavioural tests. Ideally, an indicator would be perfectly calibrated to the construct it intends to measure. In reality, we do not expect a singular survey question to give a perfectly calibrated account of an individual's openness to experience or their radicalisation. Errors can arise from particularities of the question, like confusing phrasing (an \textbf{item-specific error}), or extraneous circumstances, like a participant being tired (a \textbf{measurement error}). However, many survey questions together can overcome these errors. Multiple measurements for the same construct should share some signal related to the construct apart from their own item-specific and measurement errors. More specifically, we expect measurements for the same construct (or theoretically similar constructs) to correlate, and to correlate more than measurements for separate constructs. Returning to our example from \cref{sec:can_construct_validity_solve}, there are many benchmarks for measuring mathematical reasoning—\texttt{MATH}, \texttt{FrontierMath}, \texttt{AIME 2024}, \texttt{MathBench}, and more.\footnote{In several places, I am going to list many benchmarks in succession, and it hinders readability to include all of the necessary inline citations. Readers should consult \Cref{app:benchmarks} for a description and citation of all benchmarks mentioned in this thesis.} For these benchmarks to be identifiably measuring the same thing—mathematical reasoning—there should be some degree of convergence in their scores. If \texttt{MATH} and \texttt{FrontierMath} both measure mathematical skill, and some LLM performs above average on \texttt{MATH}, we should expect it to perform above average on \texttt{FrontierMath}. However, this convergence in scores should be to some degree localised. If every benchmark score correlates highly with every other score—that is, if \texttt{MATH} performance is no better correlated to \texttt{FrontierMath} than any arbitrary benchmark for geography or history trivia—then we will struggle to justify that \texttt{MATH} and \texttt{FrontierMath} benchmark any specific `mathematical' attribute of LLMs in place of some more general `benchmark performance' attribute. In other words, even if an LLM can recognisably solve mathematical problems, we do not have the evidence to give a mathematical reasoning capability any independent theoretical standing.

If our collection of benchmarks has both of these criteria—\emph{convergent} correlations between scores, and \emph{discriminating power} in those correlations—we can better trust that our benchmarks collectively estimate some general capability. These guarantees put our evaluation techniques in a much more powerful position: with reference to an LLM's capabilities, we can \textit{explain} how the LLM is capable of certain tasks, and we can \textit{predict} whether the LLM will be capable of novel, related tasks. If we want fair and accurate AI capability regulation—or if we ever want an LLM to file our taxes—we need these guarantees. In measurement theory, these criteria are called \textbf{convergent} and \textbf{discriminant validity}, respectively, and they are the core of construct validity in its original formulation \citep{cronbach_construct_1955}.

The simplest mathematical model for expressing these criteria is the latent factor model \citep{everittIntroductionLatentVariable1984}. Suppose that we have $k$ latent constructs and, since we cannot measure them directly, have $p > k$ indicator variables we believe to measure these constructs. These indicator variables might be questions in a $p$-question survey. The basic latent factor model assumes that
\begin{equation}
    \vx = \mL\vf + \ve\label{eq:latent_factor_model}
\end{equation}
holds for each survey participant.\footnote{All of my mathematical notation will follow \citeauthor{goodfellow2016deep}'s textbook, \textit{Deep Learning} (\citeyear{goodfellow2016deep}). \cref{app:math_notation} contains a table of the notation copied over from the textbook.} Here $\vx$ is a $p\times 1$-vector of observed responses, $\vf$ is a $k\times 1$-vector of latent \textbf{factor scores}, $\mL$ is a $p\times k$ matrix of \textit{pattern coefficients} (also called \textit{factor loadings}), and $\ve$ is a $p\times 1$-vector of error terms~\citep{klineSpecificationIdentificationConfirmatory2016}. \cref{eq:latent_factor_model} states that observed responses ($\vx$) result from some influence ($\mL$) of underlying latent constructs ($\vf$) mediated by some error ($\ve$). This model is the simplest possible model for connecting measurements to the constructs they purport to measure. Given empirical survey data, we could estimate the latent factor values with techniques including likelihood optimisation. The survey demonstrates construct validity if the indicators and latent constructs have suitable convergent and discriminant correlations. If indicators $\evx_a$ and $\evx_b$ should both measure construct $\evf_c$, and indicator $\evx_d$ should not, we should expect both $\emL_{a, c} > 0$ and $\emL_{b, c} > 0$ (convergent validity) and $\emL_{d, c} < \emL_{a, c}$ and $\emL_{d, c} < \emL_{b, c}$ (discriminant validity) to result from the model estimates.

Factor analysis models can be \textbf{exploratory}~\citep{fabrigarExploratoryFactorAnalysis2012} or \textbf{confirmatory}~\citep{klineSpecificationIdentificationConfirmatory2016}. An exploratory model freely estimates all pattern coefficients in $\mL$, while a confirmatory model sets $\emL_{i, j} \gets 0$ for particular coefficients. Thus, in an exploratory model, constructs are mostly unsupervised discoveries made by the model, more akin to an unsupervised dimensionality reduction, whereas in a confirmatory model, the researcher hypothesises the constructs that inform specific indicators. \citet{burnellRevealingStructureLanguage2023} use an exploratory factor model, while \citet{ilicEvidenceInterrelatedCognitivelike2024} fit a confirmatory factor model, specifying their LLM capabilities in advance to resemble human cognitive constructs.

In LLM evaluation, we similarly want to separate indicators for constructs from the constructs themselves. Again, returning to our mathematics example: today's best LLMs recently achieved gold-medal performance on the International Mathematics Olympiad (IMO) \citep{luongAdvancedVersionGemini2025, caiGoogleClinchesMilestone2025}. This performance indicates a latent capability which, if observed in humans, we would call mathematical reasoning or problem-solving. Do these IMO gold-medallist LLMs have the same capability? We have seen that subjective arguments either way are controversial, but the formal definition of construct validity provides a foothold.
Given a population of LLMs and performances across different mathematical benchmarks, we could fit a latent factor model to help bolster the connection between the IMO and LLMs' mathematical capability.

We do have sizeable datasets of LLM performances on benchmarks, including \texttt{HELM} \citep{liang_holistic_2023}, Chatbot Arena~\citep{chiang2024chatbotarenaopenplatform}, and the OpenLLM Leaderboard~\citep{fourrierOpenLLMLeaderboard2024}. Unfortunately, there are reasons to suspect that the standard inferential statistics behind latent factor models will not apply to LLMs in the same way that they apply to human populations.
In general, when you conduct inferential statistics on a sample, you assume that sample to be representative of some greater population, such that your findings on the sample will generalise to the population.
However, the particular dynamics and history of LLMs suggests that the very idea of an LLM `population' could be flawed. One reason is survivorship bias: commercial model providers might only release LLMs that perform well in internal testing. Studies like \citet{singh2025leaderboardillusion} have already shown how such selective reporting can distort the statistical properties of benchmark results. Another reason is the temporal variance of such a population. While estimates of human capabilities, like IQ, do drift slowly over time~\citep{trahanFlynnEffectMetaanalysis2014}, the population of LLMs is more volatile and competitive, making it difficult to identify stable and reproducible factors~\citep{zhou_general_2025}. A third reason is the scaling laws reviewed above. Unlike humans, LLMs have observable design parameters that strongly influence their performance on most tasks. These scaling parameters are not capabilities, but they can affect significant variance in observed benchmark scores. Statistical models using latent factors to capture LLM capabilities should be wary to account for these effects, as they can easily capture proxies for scaling parameters by accident.

\subsection{Related work}\label{sec:related_work}

\subsubsection*{Latent factor models of LLM capabilities}

In the first application of latent factor modelling for LLM capabilities, \citet{burnellRevealingStructureLanguage2023} fit an exploratory factor model using performance on the \texttt{HELM} benchmark, utilsing a sample of $29$ LLMs. Models in their sample range in size from $410$ million parameters (Cohere small v20220720~\citep{cohereIntroductionLargeLanguage2022}) to $530$ billion parameters (TNLG v2~\citep{smith2022usingdeepspeedmegatrontrain}) and were released between June 2021 and January 2023. The first author, a cognitive scientist, annotated each of the 34 \texttt{HELM} tasks with the capability the task elicited, including cognitive abilities like ``Inductive reasoning,'' ``Domain knowledge,'' and ``Comprehension'' (5). Fitting three latent factors to explain their benchmark results, \citeauthor{burnellRevealingStructureLanguage2023}'s model fit statistics are below the conventionally acceptable threshold in social science,\footnote{Specifically, \citeauthor{burnellRevealingStructureLanguage2023}'s model achieves $\mathrm{CFI} = 0.70, \mathrm{TLI} = 0.61, \mathrm{RMSEA} = 0.26$, and latent factor models are considered acceptable at $\mathrm{CFI} \geq 0.90, \mathrm{TLI} \geq 0.90, \mathrm{RMSEA} \leq 0.10$ and excellent at $\mathrm{CFI} \geq 0.95, \mathrm{TLI} \geq 0.95, \mathrm{RMSEA} \leq 0.06$~\citep{finchInvestigationPersonalityStructure1997, huFitIndicesCovariance1998, huCutoffCriteriaFit1999, clarkConstructingValidityNew2019}. While the specific interpretations of these statistics is not necessary for this overview, they will be described in \cref{sec:validation_experimentation} for validating my own models.} though the authors note that they expect this from the small size of their sample. Nonetheless, the authors identify and discuss three separable factors representing LLM capabilities. The first, which they label ``Language modeling,'' loads\footnote{For latent factor models, to say that a factor (or capability) $j$ ``loads'' on a benchmark $i$ means that the model fit a significant positive coefficient at $\emL_{i, j}$ in the pattern coefficient matrix $\mL$, as I introduced in \cref{sec:quantifying_construct_validity}. A larger loading means a larger coefficient $\emL_{i, j}$, which also means that factor $j$ explains more of the observed variance in benchmark $i$.} on tasks all annotated as ``Language modeling,'' including sentence completion for The Pile \citep{gao2020pile} and the International Corpus of English \citep{greenbaum1991ice}. The second, which they label ``Reasoning,'' loads on benchmarks annotated ``Mathematical reasoning,'' like \texttt{GSM8K} and \texttt{MATH}, and ``Inductive reasoning,'' like \texttt{BBQ}, among others. The final factor they label ``Comprehension,'' and it loads on a diverse set of annotated cognitive capabilities, including ``Domain knowledge'' (\texttt{HellaSwag}; \texttt{WikiFact}), ``Comprehension'' (\texttt{NaturalQuestions}; \texttt{XSUM}), ``Commonsense reasoning,'' (\texttt{OpenbookQA}) and ``Deductive reasoning'' (\texttt{bAbI}; \texttt{Dyck}). The ``Comprehension'' capability explains the largest proportion of communal variance in the measured scores ($33\%$) and it correlates the largest with other factors ($0.43$ with Language modeling; $0.51$ with Reasoning) and with the $\log$ of the LLM parameter count ($0.70$).\footnote{All correlations are Pearson correlations.} Notably, \citeauthor{burnellRevealingStructureLanguage2023} also observe a largely positive correlation manifold across their benchmarks, with the mean inter-task correlation $\overline{r} = 0.56$, meaning that each pair of benchmark results is moderately correlated on average.

\citet{ilicEvidenceInterrelatedCognitivelike2024} conduct a second experiment with latent factor models, this time using a confirmatory factor analytic approach. Their experiment uses the same LLM sample as ours: HuggingFace models evaluated on the OpenLLM Leaderboard, a much larger collection than \citet{burnellRevealingStructureLanguage2023}'s \texttt{HELM} results. They begin with the leaderboard's full population as of 8 March 2024—$3,862$ models—but reduce this number to $591$ using a combination of clustering techniques like DBSCAN \citep{DBSCAN} and subjective criteria to remove redundant models. \citeauthor{ilicEvidenceInterrelatedCognitivelike2024} hypothesise cognitive factors inherited from the Cattell-Horn-Carroll theory of human cognitive abilities~\citep{cattellHorn1978, carrollPsychometricsIntelligencePublic1997, schneiderCattellHornCarroll2018}, including fluid reasoning, quantitative knowledge, reading/writing, and domain-specific knowledge. They allocate different OpenLLM Leaderboard tasks to these different cognitive capabilities using descriptions of the tasks. Ultimately, the authors are unable to fit an acceptable model partitioning OpenLLM Leaderboard subtasks according to these capability dimensions. They instead find a two-factor model with acceptable fit.\footnote{Their model fit statistics are $\mathrm{CFI} = 0.984, \mathrm{TLI} = 0.977, \mathrm{RMSEA} = 0.089$.} This model includes one mixed factor for reading/writing and domain-specific knowledge, plus a factor that loads heavily on every benchmark indicator. The authors name this factor an ``Artificial General Ability'' (AGA) factor, in reference to the famous ``$g$-factor'' hypothesised to be a general measure of human intelligence \citep{spearmanGeneralIntelligenceObjectively1904}. This AGA factor has a large average loading of $0.81$ and accounts for $65.6\%$ of the variance in benchmark results. The authors fit a LOESS regression for the AGA factor against model parameter size, finding a ``curvilinear'' relationship explaining $25\%$ of variance in the factor value~(7). Like \citet{burnellRevealingStructureLanguage2023}, this study finds a large positive correlation manifold with mean inter-test correlation $\overline{r} = 0.73$ across benchmarks.

\subsubsection*{Observational scaling laws}

In parallel, \citet{ruan2024observational} develop a method for estimating LLM capabilities that generalises neural scaling laws. The original scaling laws apply to test-set predictive error—specifically, the cross-entropy loss on the test set \citep{kaplanScalingLawsNeural2020}. This error is an unbounded positive quantity where lower is better. Our subject matter, benchmarks, are instead scored on the interval $[0, 1]$ where higher is better. Previous studies resolve this discrepancy \citep{finnvedenExtrapolatingGPTNPerformance2020, owen2024predictable}, adapting scaling laws to predict individual benchmark scores. \citet{owen2024predictable} finds that a logistic link function between model scale and benchmark performance is the most accurate. For a model $i$ trained with $c_i$ total floating point operations (FLOps), \citet{owen2024predictable} estimates a benchmark score $b_i\in[0, 1]$ as
\begin{equation*}
    b_i = \sigma (\beta\log(c_i) + \alpha).
\end{equation*}

\citet{ruan2024observational}'s observational scaling laws take the same logistic form as the benchmark scaling laws in \citet{owen2024predictable}. However, in place of scale, observational scaling laws model benchmark performance for model $i$ as a function of $k$ distinct capabilities via a ``capability vector'' $\vs_i\in\sR^k$, such that
\begin{equation*}
    b_i \approx \sigma(\beta^\top \vs_i + \alpha)
\end{equation*}
instead. As with scaling laws, each model's capability vector is a function of the model's scale, except that the vector coefficients for converting scale into capabilities are shared across the entire model family. If model $i$ belongs to family $f$, $\vs_i$ is given by
\begin{equation*}
    \vs_i \approx \vt_f\log(c_i) + \vv_f.
\end{equation*}
For example, Llama 3.1 8B and Llama 3.1 405B would share the vectors $\vt_f$ and $\vv_f$ in virtue of belonging to the same ``Llama 3.1'' model family, despite having different values of training compute $c_i$. What makes these scaling laws ``observational'' is that the capability vectors are estimated using a shared, low-dimensional representation of observed benchmark scores. In traditional scaling laws, the coefficients mapping model scale to test-set error are estimated individually for each model family. 

\citeauthor{ruan2024observational} collect $7$ benchmark scores for $77$ models using data from the OpenLLM Leaderboard and EvalPlus \citep{liu2023is}.
To convert these scores to capabilities, they directly set the capability vectors to be the first $k=3$ principal components of their benchmark score matrix.
Specifically, if $\vb_i$ is a $p\times 1$-vector of model $i$'s benchmark scores, \citeauthor{ruan2024observational} set
\begin{equation}
    \vs_i \gets \mW\vb_i\label{eq:pca},
\end{equation}
where $\mW$ is a $k\times p$-matrix of principal component weights for all benchmark scores across the sample of $77$ LLMs. The authors find that $k=3$ delivers the best results, equal to the number of latent factors in \citet{burnellRevealingStructureLanguage2023}. These first three components explain $96.7\%$ of the variance in \citeauthor{ruan2024observational}'s observed results, with the first component explaining \textasciitilde$80\%$. They show that their derived principal component scores can be used to fit observational scaling laws for complex tasks beyond the $7$ benchmarks included in their score matrix, including some of the ``emergent'' capabilities in \citet{wei_emergent_2022}. Their method uses only scores from the lower end of the performance distribution, allowing their model to extrapolate future performance. These observational scaling laws outperform conventional, compute- and parameter count-based scaling laws in terms of predictive accuracy.

\subsubsection*{Latent factor models and PCA}

The latent factor modelling approaches, \citet{burnell_rethink_2023} and \citet{ilicEvidenceInterrelatedCognitivelike2024}, seem to have mostly explanatory goals—they aim to find clear representations of LLM capabilities that reflect benchmark data. By contrast, \citeauthor{ruan2024observational} have a mostly predictive goal—they aim to find representations of LLM capabilities that generalise to other benchmarks.

Nonetheless, there is a remarkable similarity between the latent factor models of capabilities in \citet{burnellRevealingStructureLanguage2023} and \citet{ilicEvidenceInterrelatedCognitivelike2024} (\cref{eq:latent_factor_model}) and \citeauthor{ruan2024observational}'s PCA decomposition approach (\cref{eq:pca}). Say that $\mB$ is the $p\times m$-matrix representing $p$ benchmarks scores for a population of $m$ LLMs. In matrix formulation, the latent factor model assumes
\begin{equation}
    \mB = \mL\mF + \mE,\label{eq:latent_factor_matrix_form}
\end{equation}
where pattern coefficients $\mL$ are unchanged from \cref{eq:latent_factor_model}, $\mF$ is a $k\times m$ matrix alternative for factor scores $\vf$, and $\mE$ is a $p\times m$-matrix alternative for error terms $\ve$. PCA instead utilises the decomposition
\begin{subequations}
    \begin{align*}
        \mB^\top\mW&= \mT^\top \\
        \mB &= \mW\mT,
\end{align*}
\end{subequations}
where $\mW$ is a $p\times p$ matrix of principal component weights and $\mT$ is a $p\times m$ matrix of principal component scores. The columns of $\mW$ are the eigenvectors of $\mB^\top\mB$, and the first column vector $\mW_{:, 1}$ maximises the explained variance in $\mB$ by satisfying
\begin{equation*}
    \mW_{:, 1} = \arg\max_{\lVert\vw\rVert = 1}\sum_i (\mB_{:, i}\cdot\vw)^2
\end{equation*}
and is therefore the first principal component~\citep{gewersPrincipalComponentAnalysis2022}. For dimensionality reduction, we can use the approximation
\begin{equation}
    \mB\approx \mW_{:, :k}\mT_{:k, :}\label{eq:pca_approx}
\end{equation}
which represents $\mB$ using the top $k$ principal components.\footnote{The term $\mW_{:, :k}$ represents a minor abuse of \citet{goodfellow2016deep}'s matrix notation. The intended reading is Pythonic: $\mW_{:, :k}$ represents all rows and the first $k$ columns of the matrix $\mW$. It is the $p\times k$ slice of a $p\times p$ matrix.} This is how \citet{ruan2024observational} achieve their version of pattern coefficients in the $p\times k$ matrix $\mW_{:, :k}$. Since $\mW_{:, :k}$ resembles the pattern coefficient matrix $\mL$, we can see \cref{eq:pca} as equivalent to \cref{eq:latent_factor_model} with $\mE = \mathbf{0}$.\footnote{This reformulation applies with the caveat that the pattern coefficient vectors $\mW_{:, j}$ for each `factor' $j$ are orthonormal, which is not necessary for general latent factor models.} In other words, construed as a model of common factors, PCA implicitly assumes that all variance is common variance and that there are no item-specific errors, including measurement errors. In the context of LLM benchmarks, this assumption states that LLM performances are not distorted by the unique phrasing of instructions or output format requirements in individual benchmarks. In contrast, the latent factor model allows for item-specific errors with the matrix $\mE$. To be precise, the model will estimate $p$ separate covariance parameters for each benchmark, then represent each row of $\mE$ using normally-distributed random errors according to these covariances~\citep{klineSpecificationIdentificationConfirmatory2016}. If benchmarks have specific formatting quirks that cause performance errors—for example, prompt phrasings that we know influence performance \citep{mizrahi-etal-2024-state, voronov-etal-2024-mind}—this assumption provides a `home' for the error in the statistical model. It does not force the capability estimates to pick up this error variance, while PCA decomposition does.

\subsection{Research questions}

I have now explained latent factor models and observational scaling laws in some mathematical detail. Both techniques can take a collection of LLM benchmark scores and estimate the capabilities behind these scores. Given benchmark scores $\mB$, latent factor models provide factor scores $\mF$ as capabilities, while observational scaling laws provide principal component scores $\mT_{:k, :}$. It should also be apparent that each approach deploys some domain knowledge that the other is missing. Observational scaling laws leverage model scale, including the results from \citet{finnvedenExtrapolatingGPTNPerformance2020} and \citet{owen2024predictable} that show that benchmark performance is decently explained by model scale on its own. Latent factor models ignore scaling laws, so they lose out on this predictive utility. Yet, latent factor models appear to capture a correct assumption about the individual error signals in benchmarks: the model's error terms capture these signals, providing the factors with cleaner estimates of general capabilities.

From these observations, I would like to determine if these are real deficiencies in these models, and if the supposed advantages of each approach can be conferred to the other. I propose two research questions:
\begin{enumerate}[label=\Alph*.]
    \item Does providing latent factor models with information about model scale improve their fit and explanatory power?
    \item Does providing observational scaling laws with latent factor capability estimates improve their predictions for novel benchmarks?
\end{enumerate}

\clearpage

%% file: sections/3_methods.tex
\section{Methods}

\subsection{Experimental design}\label{sec:experimental_design}

In this section, I propose two experiments to address my two research questions. The first experiment compares latent factor models against an alternative model; the second experiment compares observational scaling laws against an alternative model. The alternative model in each case is my structured capabilities model, a hybrid of the two. I will define the structured capabilities model in \cref{sec:techniques_models}.

\subsubsection*{Experiment A: Parameter size as an upstream cause of LLM capabilities}

Following \citet{burnellRevealingStructureLanguage2023} and \citet{ilicEvidenceInterrelatedCognitivelike2024}, my model uses factor analytic methods from the social sciences to decompose observed benchmark results into common capability scores. Adopting factor analysis from social science provides a measured and proven explanatory basis for the estimated capabilities. However, I believe that the omission of learnings from the AI scaling law literature makes these models deficient in some testable ways.

First, both \citeauthor{burnellRevealingStructureLanguage2023} and \citeauthor{ilicEvidenceInterrelatedCognitivelike2024}'s experiments share the `pure measurement' modelling assumption embodied in \cref{eq:latent_factor_matrix_form}, where the only variables influencing benchmark scores are the latent capability parameters $\mF$ and $\mL$ and the error terms $\mE$. Both experiments report a large positive correlation between benchmarks and a dominant singular latent factor that correlates strongly with model size. These results look suspiciously like the models are fitting proxies for model size in place of an actual capability.
I believe that these results suggest the pure measurement methodology is flawed, and that model size should be considered an \textit{exogenous upstream cause} of LLM capabilities within the model itself.\footnote{While I use the phrase ``exogenous upstream cause'' as a motivation for my modelling approach, the structured capabilities model does not allow for causal inference. We discuss this distinction in the limitations in \cref{sec:limitations}.} To test this hypothesis, I will compare a pure exploratory factor analysis (EFA) model before and after factors are constrained to depend on a single upstream variable: the $\log$ of the number of LLM model parameters, a measure of model scale used in \citet{owen2024predictable}'s scaling laws. I hypothesise that including the single upstream variable will provide a better explanation of my benchmark data. I suspect I will see the same dominant factor in my pure EFA model that occurs in \citet{burnellRevealingStructureLanguage2023} and \citet{ilicEvidenceInterrelatedCognitivelike2024}, and that in the presence of model size, this factor will be revealed as an illusory proxy for scale.


Second, standard factor analysis models as in \citeauthor{burnellRevealingStructureLanguage2023} and \citeauthor{ilicEvidenceInterrelatedCognitivelike2024} explain indicator variables (e.g., benchmark performances) as \textit{linear} responses to upstream latent factors. However, we know from \citet{owen2024predictable} that linear functions fare poorly for predicting benchmark performance. If I hypothesise that model size is a common upstream cause of model capabilities, I should also expect that the shape of this response is not linear, but logistic. I will transform my benchmark data to model a logistic relationship to see if the model fit improves.

\subsubsection*{Experiment B: Latent factors as generalisable LLM capabilities}

The structured capabilities model proposed above resembles observational scaling laws: it explains LLM capabilities with using model size, and it does so with logistic relationships. However, the approach differs in estimating capability scores as latent factor scores instead of principal component weights.

In \cref{sec:related_work}, I covered how latent factor and PCA decomposition encode different assumptions about communal and item-specific variance. The social sciences advocate measurement models like factor analysis models over data-driven techniques like PCA for the explanation of common causes. This preference is grounded in the improved hypothesis testing abilities and theoretical grounding of the measurement models. However, I also suspect that latent factor modelling offers an improvement over PCA more desirable in computer science literature, which is an improved \textit{generalisation} of capabilities to unseen, out-of-distribution benchmarks.

I suspect improved generalisation for two reasons. First, as covered in \cref{sec:related_work}, PCA implicitly estimates the item-specific error to be $0$ for every indicator variable.
Isolating no benchmark-specific variance will presumably overfit capability estimates to the particular benchmarks included in the PCA decomposition, harming the generalisation of these capabilities to novel settings.
Second, as principal components must be extracted \textit{before} modelling the contribution of model size, observational scaling laws are forced to incorporate variance due to model size in the principal component scores. These components are supposed to indicate capabilities independent from model size, yet seeing the results from \citet{ruan2024observational}, I expect the first principal component to proxy model size just like the dominant factors in the latent factor models.

I will test this hypothesis using a similar experimental setup to \citet{ruan2024observational}. I will fit both the observational scaling law and structured capabilities models to a subset of benchmarks, then I will use the estimated capabilities to extrapolate performance on a held-out benchmark. I will use prediction error as a measure of each technique's generalisation.

\subsection{Data}\label{sec:data}

\subsubsection*{The OpenLLM Leaderboard}

I source benchmark results from
the OpenLLM Leaderboard, a popular leaderboard providing standardised and open-source evaluation of LLMs~\citep{fourrierOpenLLMLeaderboard2024}. The v2 rendition of the OpenLLM Leaderboard evaluates LLMs on six popular benchmarks: \texttt{BIG-Bench Hard} (\texttt{BBH}), \texttt{IFEval}, \texttt{MATH}, \texttt{GPQA}, \texttt{MuSR}, and \texttt{MMLU-PRO}. As of 13 March 2025, the OpenLLM Leaderboard v2 is no longer taking submissions. The creators cite its obsolescence in the face of improving LLM reasoning and agentic capabilities, which their chosen six benchmarks do not assess~\citep{fourrierItsBeenWild2025}. At the time of closure, the v2 Leaderboard reported evaluations for $4,576$ LLMs across the aforementioned six benchmarks.

The OpenLLM Leaderboard v2 represents a complete and thorough experiment in public LLM evaluation. The $4,576$ LLMs represent $64$ distinct architectures and range in size from below $1$ to above $140$ billion parameters. All evaluated models are open-source and the results of evaluation are publicly inspectable. The leaderboard itself is a popular public resource within the AI community: as of July 2025 it remains the largest `Space' on the HuggingFace website, having amassed \textasciitilde 13,300 likes~\citep{huggingFaceSpaces}. Many academic studies use the leaderboard to understand LLM performance, including both of \citet{ilicEvidenceInterrelatedCognitivelike2024} and \citet{ruan2024observational}. Alongside \texttt{HELM}~\citep{liang_holistic_2023} and Chatbot Arena~\citep{chiang2024chatbotarenaopenplatform}, the OpenLLM Leaderboard is one of the most popular sources for LLM evaluation results, and almost certainly the largest public database of such results.

Following claims about gamification and unfair reporting on other popular leaderboards \citep{singh2025leaderboardillusion}, one might be wary that the leaderboard's popularity comes at the expense of its scientific utility. However, the OpenLLM Leaderboard team takes several measures to ensure the fairness of their reporting. All OpenLLM Leaderboard evaluations are conducted using a public fork of the Eleuther AI Language Model Evaluation Harness~\citep{eval-harness}, ensuring independent reproducibility of results. Once submitted for evaluation, an LLM's results are made public and immutable to discourage selective reporting~\citep{fourrierPerformancesArePlateauing2024}.

\subsubsection*{BIG-Bench Hard}

\begin{table}[t]
    \centering
    \begin{tabular}{lrr}
        \toprule
        \bf Subtask & \bf \# Questions & \bf \# Multiple-choice options \\
        \midrule
        Boolean expressions & 250 & 2 \\
        Causal judgement & 187 & 2 \\
        Date understanding & 250 & 6 \\
        Disambiguation QA & 250 & 3 \\
        Formal fallacies & 250 & 2 \\
        Geometric shapes & 250 & 11 \\
        Hyperbaton & 250 & 2 \\2
        Logical deduction & 750 & 7 \\
        Movie recommendation & 250 & 6 \\
        Navigate & 250 & 2 \\
        Object counting & 250 & 19 \\
        Penguins in a table & 146 & 5 \\
        Snarks & 178 & 2 \\
        Sports understanding & 250 & 2 \\
        Temporal sequences & 250 & 4 \\
        Tracking shuffled objects & 750 & 7 \\
        Reasoning about colored objects & 250 & 18 \\
        Ruin names & 250 & 6 \\
        Web of lies & 250 & 2 \\
         \bottomrule
    \end{tabular}
    \caption[BIG-Bench Hard subtasks in our dataset]{\textbf{BIG-Bench Hard subtasks in my dataset}.}
    \label{tab:bbh_subtasks}
\end{table}

Of the six benchmarks on the OpenLLM Leaderboard, I take scores on the \texttt{BBH} benchmark for my experiments. \texttt{BBH}, short for BIG-Bench Hard~\citep{BBH}, is a curated subset of the most challenging tasks from the \texttt{BIG-bench} dataset \citep{srivastava2023beyond}. \texttt{BIG-bench} is the result of a broad community effort spanning teams from OpenAI and Google as well as the greater AI and software communities. 450 authors, representing 132 institutions, contributed 204 tasks \citep[2]{srivastava2023beyond}. These tasks are diverse in both difficulty and scope, including problems drawn from biology and physics, common-sense reasoning and logic puzzles, and problems designed to test social bias. All \texttt{BIG-bench} tasks are evaluated by a team of relevant human experts to both validate problem construction and provide a human baseline.

To create \texttt{BBH}, \citeauthor{BBH} select 23 \texttt{BIG-bench} tasks where no LLM outperformed the average human rater, filtering additionally for sufficiently large sample sizes (>$100$ examples) and straightforward evaluation metrics (either exact match or multiple choice). The authors take a random subsample of up to 250 questions from each subtask, resulting in a benchmark dataset with 23 partitions and 6,511 total questions. These tasks are similarly diverse. For example, ``Geometric shapes'' asks the LLM to predict the shape drawn from some SVG code, while ``Boolean expressions'' gives the LLM a logical formula to parse, and ``Snarks'' asks the LLM to detect sarcasm. \cref{tab:bbh_subtasks} lists the subtasks that I use, and \cref{app:bbh_tasks} provides descriptions and example problems.

\texttt{BBH} is a popular benchmark that has enjoyed considerable attention in the scaling law literature. \citet{wei_emergent_2022}'s claims of ``emergent capabilities'' in LLMs used \texttt{BBH} subtasks for their experiments, and \citet{owen2024predictable}'s scaling laws predicted performance on \texttt{BBH} subtasks. In addition, \citet{ruan2024observational} used \texttt{BBH} data to fit their observational scaling laws, while \citet{burnellRevealingStructureLanguage2023} and \citet{ilicEvidenceInterrelatedCognitivelike2024} used related collections of benchmark tasks (\texttt{HELM} and mostly \texttt{MMLU}, respectively).

\subsubsection*{Data collection}

The OpenLLM Leaderboard itself reports only one aggregate \texttt{BBH} score. The leaderboard attempts to normalise each task by difficulty, though in a previous experiment I showed that subtasks remain a significant fixed effect for individual question scores, even after their contributions are normalised \citep{1089786TestletEffectsDifficulty2025}. I therefore collect and aggregate subtask accuracies from the raw evaluation results available on HuggingFace.

On the OpenLLM Leaderboard, four original \texttt{BBH} tasks are not reported for all models (Dyck languages, Multi-step arithmetic, Salient translation error detection, and Word sorting). Two are split into three separate tasks based on their complexity level (Logical deduction and Tracking shuffled objects). I recombine these results, yielding a final dataset of $p=19$ benchmark indicator variables instead of $23$. I also find that $181$ of the $4,576$ models included in the leaderboard are missing complete results. Finding no relationship to model size, publication date, or overall performance to explain these missing values, I simply drop these models from the dataset, yielding a final sample size of $m=4,395$ LLMs.

\subsubsection*{Formal notation and scoring}

As in \cref{sec:related_work}, I will write the $p\times m$ matrix $\mB$ to represent the performance across $p=19$ benchmarks for each of $m=4,395$ models. Each benchmark score is a real number in the interval $[0, 1]$ representing the average accuracy across all subtask questions. All \texttt{BBH} tasks on the OpenLLM Leaderboard are presented as multiple choice questions. Subtasks range from having 2 to 19 possible options (see \cref{tab:bbh_subtasks}). The evaluation harness scores the model according to the multiple choice letter that the model assigns the highest probability, meaning that LLMs never fail to provide an answer.

For each of the $m$ models, I also collect the available metadata for the model's parameter size. I denote parameter size with the variable $n$, such that model $j$ with benchmark scores $\mB_{:, j}$ has $n_j$ parameters.

\subsection{Techniques and models}\label{sec:techniques_models}

\subsubsection*{The logistic model}

Specifying the logistic model for experiment A requires unpacking the assumptions behind \citet{owen2024predictable}'s logistic model in more detail. Benchmark scores $\emB_{i, j}$ are bounded by $[0, 1]$, but since LLMs always provide one answer, even random guessing achieves an expected accuracy above $0$ on multiple choice questions. Thus, \citeauthor{owen2024predictable} does not predict raw scores with a form as simple as $\emB_{i, j} = \sigma(\alpha + \beta\log(n_j))$, because this relationship will predict $\emB_{i, j} \to 0$ for the smallest models. Instead, for a subtask $i$ with $\#\mathrm{MC}$ multiple-choice options, the worst models should achieve an expected accuracy of $c_i:= 1 / \#\mathrm{MC}$ through random guessing. \citet{owen2024predictable} fixes this problem by subtracting $c_i$ from observed performances and fitting a logistic function with a linear offset to compensate for adjusting the range. This model takes the form
\begin{equation}
    \emB_{i, j} = \sigma\big(k_i(\log(n_j)-x_{0_i})\big) + y_{0_i} + c_i.\label{eq:logistic_fixed_offset}
\end{equation}

\begin{figure}[t]
    \centering
    \includegraphics[width=0.75\textwidth]{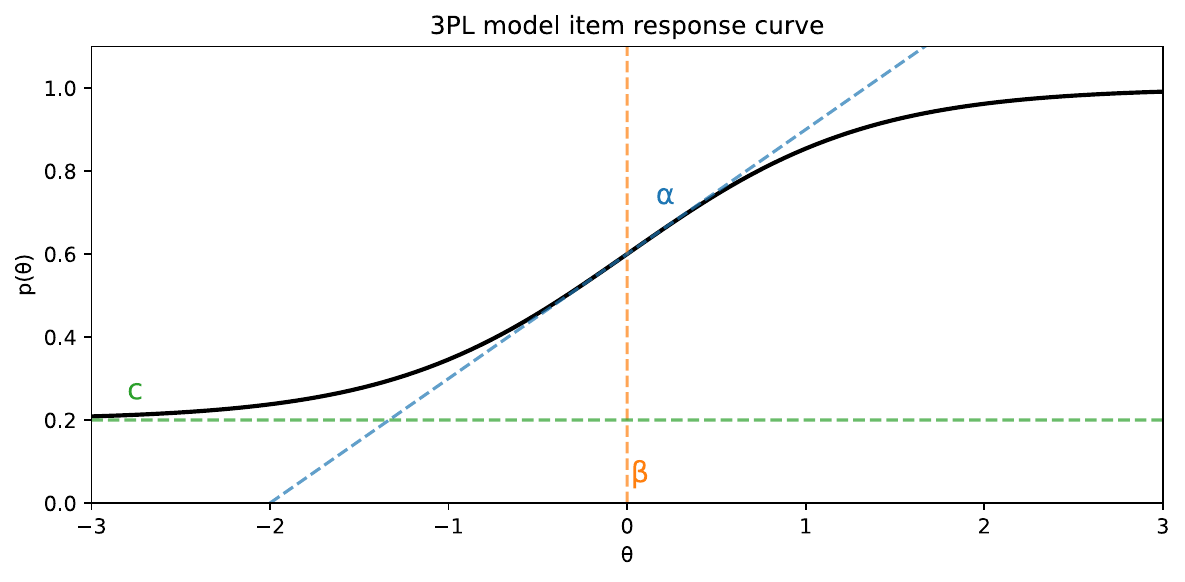}
    \caption[Illustrated three-parameter item response curve]{\textbf{Illustrated three-parameter item response curve}. This curve represents \cref{eq:3pl_irt}. ``Difficulty'' ($\beta = 0$) represents how difficult an item is relative to others by affecting a lateral transformation in $p(\theta)$ across all ability scores. ``Discrimination'' ($\alpha = 1$) represents the item's ability to differentiate a narrow range of ability scores by affecting the slope of the sigmoidal curve $\frac{d}{d\theta} p(\theta)$. ``Guessing probability'' ($c = 1/5 = 0.2$) represents the rate of success at $\lim_{\theta\to 0}p(\theta)$, which for multiple-choice questions will be the odds of guessing correctly from $1/c$ options.}
    \label{fig:3pl_curve}
\end{figure}

However, I want to adapt a more elegant model from the social sciences to achieve the same estimation within the original interval $[0, 1]$. This model is the three-parameter logistic model from Item Response Theory~\citep{embretsonItemResponseTheory2013}, which gives the probability of correctly answering a singular multiple choice question, accounting for random guessing. Typically, this model depends on an individual $j$'s latent \textit{ability score}, $\theta_j$:
\begin{equation}
    p_i(\theta_j) = c_i + \frac{1 - c_i}{1 + \exp[-\alpha_i(\theta_j - \beta_i)]}\label{eq:3pl_irt}
\end{equation}
In the case that $i$ represents a singular test item (that is, a singular multiple choice question), the parameters $\beta_i$, $\alpha_i$, and $c_i$ have specific interpretations. \cref{fig:3pl_curve} illustrates the effects of these parameters on the item's \textit{response curve}, $p_i$. Observing the similarity between \citet{owen2024predictable}'s \cref{eq:logistic_fixed_offset} and item response theory's \cref{eq:3pl_irt}, I would like to substitute $\log(n_j)$ for $\theta_j$ in \cref{eq:3pl_irt} to predict the entire subtask score $\emB_{i, j}$. Even though \cref{eq:3pl_irt} is defined for singular questions, I can adapt it to entire average scores with some simplifying assumptions.

Individual multiple choice question scores are dichotomous: the model either answers correctly or it does not. Thus, in item response theory, a model $j$'s score on a given question $i$ is a Bernoulli random variable with $p = p_i(\theta_j)$, where $p$ reflects the question's difficulty. On the simplifying assumption that all $q$ questions in each subtask are equally difficult, if we weigh all $q$ questions equally, then a model's total score is given by
\begin{equation*}
    \frac{\mathrm{Binom}(q, p_i(\theta_j))}{q},
\end{equation*}
with expected value
\begin{equation*}
    \E\bigg[\frac{\mathrm{Binom}(q, p_i(\theta_j))}{q}\bigg] = p_i(\theta_j),
\end{equation*}
meaning that \cref{eq:3pl_irt} can be utilised to represent model $j$'s expected score across an entire subtask as a function of the ability parameter $\theta_j$. Thus, we can simply swap $\log(n_j)$ in for $\theta_j$ to get a logistic model similar to \citet{owen2024predictable}'s:
\begin{equation}
    \emB_{i, j} = c_i + \frac{1 - c_i}{1 + \exp[-\alpha_i(\log(n_j) - \beta_i)]}.\label{eq:logistic_scaling_law_model}
\end{equation}

\cref{eq:logistic_scaling_law_model} has the advantage of ranging over the original interval $[0, 1]$ without any offset, plus the nice interpretations for $\beta_i$, $\alpha_i$, and $c_i$ described in \cref{fig:3pl_curve} as applied to \texttt{BBH} subtasks.

\subsubsection*{The structured capabilities model}

At this stage, I can formally define the structured capabilities model that I will use for the two experiments. The structured capabilities model resembles the fully latent model in \cref{eq:latent_factor_matrix_form}, except it allows model size to directly influence the latent capability estimates. The model replaces the factor value matrix $\mF$ like so:
\begin{equation}
    \mB = \mL[\vw_n\log\vn] + \mE\label{eq:structural_regression}
\end{equation}
Here $\vn = [n_1,\dots, n_m]$ is a vector of each model's parameter size and $\log$ is the natural logarithm applied elementwise. $\vw_n$ is a $k\times 1$-vector of regression coefficients that determine how model size affects each LLM capability. Under this model, each LLM $j$ will have a vector of capability scores equal to $\vw_n\log(n_j) + \epsilon$, where $\epsilon$ is a residual error term that occurs because of $\mE$. Thus, model size affects each model capability, as in a scaling law, but model- and capability-specific errors affect this relationship, as in a measurement model.

This relationship is not yet entirely faithful to scaling laws, however, since the relationship between model size and performance is linear. I hypothesised that a logistic relationship will better fit the data. Above, I provided an adapted three-parameter logistic model (\cref{eq:logistic_scaling_law_model}) that uses model size $n_j$ as the predictor for subtask performance. Now, I would like to use the structured capability estimates from \cref{eq:structural_regression} in place of model size.

\cref{eq:structural_regression} is a type of structural equation model, and the best available software package for estimating such a model is the \texttt{lavaan} package in R~\citep{rosseelLavaanPackageStructural2012}. Ideally, I could model each vector of subtask scores $\mB_{i, :}$ directly with the logistic relationship
\begin{equation}
    \mB_{i, :} = c_i + \frac{1 - c_i}{1 + \exp[-\alpha_i(\mL_{i, :}[\vw_n\log\vn] + \mE_{i, :}-\beta_i)]}.\label{eq:direct_logistic_relationship}
\end{equation}
Unfortunately, \texttt{lavaan} does not support this kind of nonlinearity in its model syntax. This model also resembles a multidimensional item-response theory model \citep{reckaseMultidimensionalItemResponse2009}, which handles nonlinearity and is definable in R with the \texttt{mirt} package~\citep{chalmersMirtMultidimensionalItem2012}. To my knowledge, though, this package cannot handle the structural coefficients that I need.
Thus, I instead apply the inverse operation to the data matrix $\mB$, yielding a transformed matrix $\mB'$. It is equivalent to \cref{eq:direct_logistic_relationship} to represent each row $\mB_{i, :}$ with

\begin{equation}
    \mB'_{i, :} = \beta_i + \frac{1}{\alpha_i}\log\bigg[\frac{\mB_{i, :} - c_i}{1 - \mB_{i, :}}\bigg],\label{eq:transformation}
\end{equation}
and then to use the structural equation
\begin{equation}
    \mB' = \mL[\vw_n\log\vn] + \mE\label{eq:logistic_structural_regression}
\end{equation}
to model the logistic relationship between capabilities and benchmark scores.\footnote{In \cref{app:derivation}, I provide the derivation for \cref{eq:transformation} beginning from \cref{eq:direct_logistic_relationship}.} I apply the transformation from \cref{eq:transformation} on the OpenLLM Leaderboard data matrix before loading the data into R, where I can then model the linear response in \cref{eq:logistic_structural_regression} with \texttt{lavaan} syntax. This final equation represents the structured capabilities model that I propose to combine latent factor models and observational scaling laws.

\subsection{Validation and experimentation}\label{sec:validation_experimentation}

\subsubsection*{Experiment A: Parameter size as an upstream cause of LLM capabilities}

My hypotheses for experiment A are testable by comparing the structured capabilities model against a pure latent factor model. In \cref{sec:techniques_models}, I show how to introduce the structural parameter for model size (\cref{eq:structural_regression}), and also how to transform the data matrix to model the logistic relationship between capabilities and performance (\cref{eq:transformation} and \cref{eq:logistic_structural_regression}). To consolidate my equations into one place, I will be comparing the following four models to test these two alterations:
\begin{subequations}
    \begin{align}
        \mB &= \mL\mF + \mE\label{eq:exp_a_1} \\
        \mB &= \mL[\vw_n\log\vn] + \mE\label{eq:exp_a_2} \\
        \mB' &= \mL\mF + \mE\label{eq:exp_a_3} \\
        \mB' &= \mL[\vw_n\log\vn] + \mE\label{eq:exp_a_4}
    \end{align}
\end{subequations}

I believe that the pure latent factor models, \cref{eq:exp_a_1} and \cref{eq:exp_a_3}, will fit a dominant factor serving as a proxy for model size. As these measurement models are unconstrained, likelihood optimisation should compel them to fit a parameter to the feature explaining the greatest variance in the data, and this feature will turn out to be model size. I believe that this occurs for the pure measurement models in the studies by \citet{burnellRevealingStructureLanguage2023} and \citet{ilicEvidenceInterrelatedCognitivelike2024}, specifically \citeauthor{burnellRevealingStructureLanguage2023}'s ``Comprehension'' factor and \citeauthor{ilicEvidenceInterrelatedCognitivelike2024}'s ``Artificial General Ability'' factor. In this model comparison, the effect should be visually apparent in the values of the pattern coefficient matrix $\mL$. The pure measurement models should estimate factors with a high pattern coefficients on all indicators; that is, large values of $\mL_{:, k'}$ for the offending factor $k'$. I can compare the variance explained by each factor using the sum of squared factor loadings $\lVert \mL_{:, k'}\rVert$ \citep{klineSpecificationIdentificationConfirmatory2016}, where I expect a single factor to dominate the explained variance.

I propose specifically that the structural models \cref{eq:exp_a_2} and \cref{eq:exp_a_4} will be better explanatory models in virtue of being more \emph{parsimonious} than their unrestricted equivalents. `Parsimonious' has a precise definition for statistical models—a more likely model with a simpler structure is more parsimonious. This criterion balances model fit with model complexity. My hypothesis is directly testable by comparing parsimonious fit indices between each pair of models. I will compare the Akaike~\citep{akaikeIC} and Bayesian~\citep{schwarzEstimatingDimensionModel1978} information criteria ($\mathrm{AIC}$ and $\mathrm{BIC}$), each of which balances model fit against complexity to quantify parsimony. These are defined as
\begin{subequations}
    \begin{align*}
        \mathrm{AIC} &:= 2d - 2\ln(\hat{\ell}) \\
        \mathrm{BIC} &:= d\ln(m) - 2\ln(\hat{\ell}),
    \end{align*}
\end{subequations}
where $d$ is the number of parameters estimated by the model, $m$ is the sample size, and $\hat{\ell}$ is the maximised value of the likelihood function for the model. Lower values of both criteria are better. The latent factor models \cref{eq:exp_a_1} and \cref{eq:exp_a_3} estimate $k$ factor loadings for each of the $p$ benchmarks, plus $p$ unique error variances for each benchmark and $\binom{k}{2}$ covariance terms between factor pairs. So, these models estimate a total of
\begin{equation*}
    d = k\times p + p + \binom{k}{2}\text{ parameters}.
\end{equation*}
Our structured models \cref{eq:exp_a_2} and \cref{eq:exp_a_4} estimate the same parameters, plus an additional $k$ for the regression coefficients in $\vw_n$. For these structural models to be more parsimonious, the improvement to the likelihood $\hat{\ell}$ must be great enough to offset this increase in parameters $d$.

I also hypothesise that a logistic relationship between LLM capabilities and benchmark scores should outperform a linear relationship, which is the typical standard for latent factor models. If my software packages allowed me to model the logistic relationship directly on the raw data $\mB$, I could compare the model fits with a likelihood-ratio test. Unfortunately, since I must instead apply the inverse transformation in \cref{eq:transformation} to get $\mB'$, the likelihoods will not be directly comparable, since transforming $\mB$ to $\mB'$ changes the null model and the likelihood landscape. Therefore, I can only compare the model fit using the conventional heuristics for global fit indices, like the $\chi^2$ value, the comparative fit index ($\mathrm{CFI}$~\citep{bentler1990comparative}), the root-mean squared error of approximation ($\mathrm{RMSEA}$~\citep{steiger1990structural}), and the standardized root mean squared residual ($\mathrm{SRMR}$~\citep{huCutoffCriteriaFit1999}). $\mathrm{CFI}$ is a comparative metric that compares a model against its null model, with values bound between $0$ and $1$ where higher values are better. $\mathrm{RMSEA}$ and $\mathrm{RSMR}$ are both estimation error measures where values closer to $0$ are better. While these metrics respond differently to features of the experimental setup, like sample size,\footnote{For example, $\mathrm{RSMR}$ appears more robust in situations where residuals are correlated \citep{montoyaPoorFitModel2021} and scales differently with sample size compared to $\mathrm{RMSEA}$~\citep{shiAssessingFitOrdinal2020}.} social science literature generally advises that models are acceptable at $\mathrm{CFI} \geq 0.9; \mathrm{RMSEA} \leq 0.10; \mathrm{SRMR} \leq 0.10$, and excellent at $\mathrm{CFI} \geq 0.95; \mathrm{RMSEA} \leq 0.06; \mathrm{SRMR} \leq 0.06$~\citep{finchInvestigationPersonalityStructure1997, huFitIndicesCovariance1998, huCutoffCriteriaFit1999, clarkConstructingValidityNew2019}.

\subsubsection*{Experiment B: Latent factors as generalisable LLM capabilities}

In experiment A above, I test my model's assumptions that differ from latent factor models: the structural parameter for model size and the logistic relationship between capabilities and benchmark scores. In experiment B, I want to test how the model's assumptions compare to observational scaling laws. This involves comparing latent variables to principal components as representations of LLM capability dimensions. For two reasons, I cannot use a comparative fit test for this comparison. First, while latent factors provide a testable model of the observed benchmark data, PCA is not testable. Second, computer science literature tends to prefer improvements to predictive accuracy over explanatory measures like fit indices, so I will prefer to demonstrate improvement in predictive accuracy.

I will compare the two models using an out-of-fold prediction procedure. For each of $p'\in\{1,\dots, 19\}$, I withhold $\mB'_{p', :}$, the transformed scores for \texttt{BBH} subtask number $p'$. I can fit both a structural model and a PCA decomposition on this data:\footnote{Note that the PCA approach uses $\mB_{i\neq p', :}$, not $\mB'_{i\neq p', :}$. \citet{ruan2024observational}'s observational scaling laws fit the PCA decomposition directly to the raw benchmark scores, then predict downstream performances with a logistic link function. To be equivalent, my approach fits the principal components using $\mB$, then predicts the transformed scores $\mB'$.}
\begin{subequations}
    \begin{align}
        \mB'_{i\neq p', :} &= \mL[\vw_n\log\vn] + \mE\label{eq:structured_capabilities_exp_b} \\
        \mB_{i\neq p', :} &= \mW_{:, :k}\mT_{:k, :}
    \end{align}
\end{subequations}

The structured capabilities model \cref{eq:structured_capabilities_exp_b} fits a $k\times m$ matrix of model capability scores equal to $\vw_n\log\vn + \epsilon$. The PCA model fits a $k\times m$ matrix of principal component `capability' scores equal to $\mT_{:k, :}$. Which estimate of capabilities generalises better to the unseen subtask $p'$? To answer, I will split the held-out scores $\mB'_{p', :}$ into separate training and test sets, $\vb_{p'}^\mathrm{train}$ and $\vb_{p'}^\mathrm{test}$, where $\vb_{p'}^\mathrm{train}$ consists of the bottom $80\%$ of scores. I will use the training set to fit $k$-length vectors of coefficients via linear regression:
\begin{subequations}
    \begin{align}
        \vb_{p'}^\mathrm{train} &= \vl_{p'} \vw_n\log\vn\label{eq:exp_b_latent} \\
        \vb_{p'}^\mathrm{train} &= \vw_{p'} \mT_{:k, :}\label{eq:exp_b_pca}
    \end{align}
\end{subequations}

The vectors $\vl_{p'}$ and $\vw_{p'}$, like the row vectors of $\mL$ and $\mW_{:, :k}$, encode the importance of each of the $k$ capabilities for subtask $p'$. I can use each of these and the learned capability scores to predict the held-out performances $\vb_{p'}^\mathrm{test}$. To match the presentation in \citet{ruan2024observational}, I will compare each approach using its mean-squared error ($\mathrm{MSE}$). I will also compare a simple scaling law approach, per \citet{owen2024predictable}, as a baseline predictive model:
\begin{equation}
    \vb_{p'}^\mathrm{train} = \alpha + \beta\log\vn\label{eq:exp_b_scale}
\end{equation}

\clearpage

%% file: sections/4_results.tex
\section{Results}

\subsection{Preliminary data analysis}\label{sec:preliminary_analysis}

\subsubsection*{Inter-item correlations}

\begin{figure}[p]
    \centering
    \includegraphics[width=\linewidth]{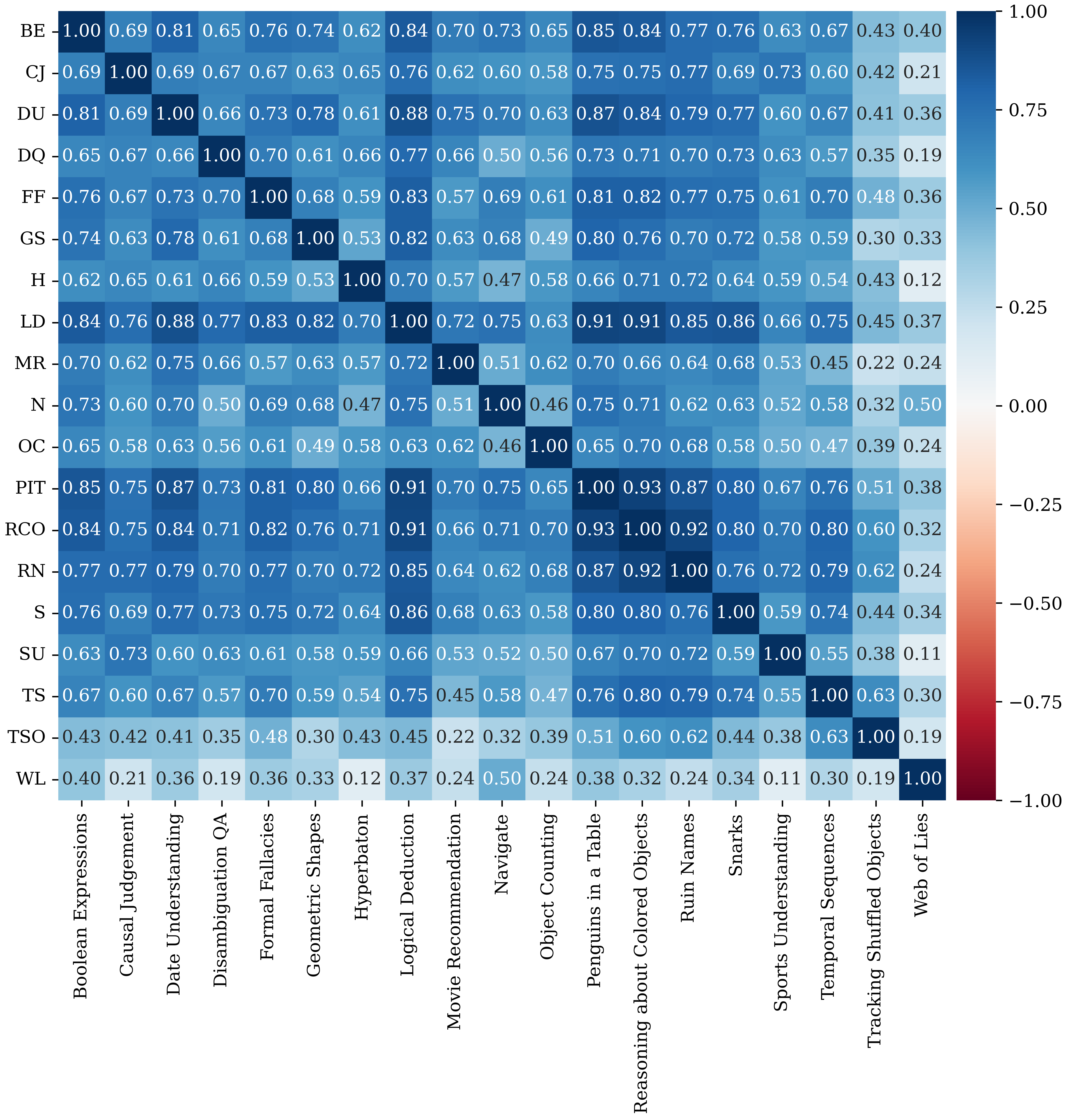}
    \caption[Raw \texttt{BBH} score correlation matrix]{\textbf{Raw \texttt{BBH} score correlation matrix}. The correlation metric is Spearman's rank correlation $\rho$. The labels along the $y$-axis are acronym abbreviations for the full subtask names, which are given along the $x$-axis.}
    \label{fig:raw_correlation_matrix}
\end{figure}

For an initial investigation into the relationship between subtask performances, I plot the inter-item correlation between subtask scores in \cref{fig:raw_correlation_matrix}. Consistent with \citet{burnellRevealingStructureLanguage2023} and \citet{ilicEvidenceInterrelatedCognitivelike2024}, there is a strong positive manifold to my dataset—the average inter-item Spearman correlation is $\overline{\rho} = 0.64$.\footnote{\citet{burnellRevealingStructureLanguage2023} and \citet{ilicEvidenceInterrelatedCognitivelike2024} both report Pearson instead of Spearman correlation. However, I am not interested in any parametric relationships at this stage. \citet{owen2024predictable}'s sigmoidal scaling laws suggest that model subtask performances might not relate in a linear fashion, and in any case, I am interested in whether the same models perform better across all subtasks, which is a nonparametric relationship.} Similar tasks like Penguins in a table and Reasoning about colored objects, both of which involve recalling attributes from structured lists, have correlations as high as $\rho = 0.93$. At the opposite extreme, dissimilar tasks like Sports understanding and Web of lies have correlations as low as $\rho = 0.11$, though due to the large sample size, all correlations are significant.

\subsubsection*{Predicting subtask performance with model size}

\begin{figure}[t]
    \centering
    \includegraphics[width=\linewidth]{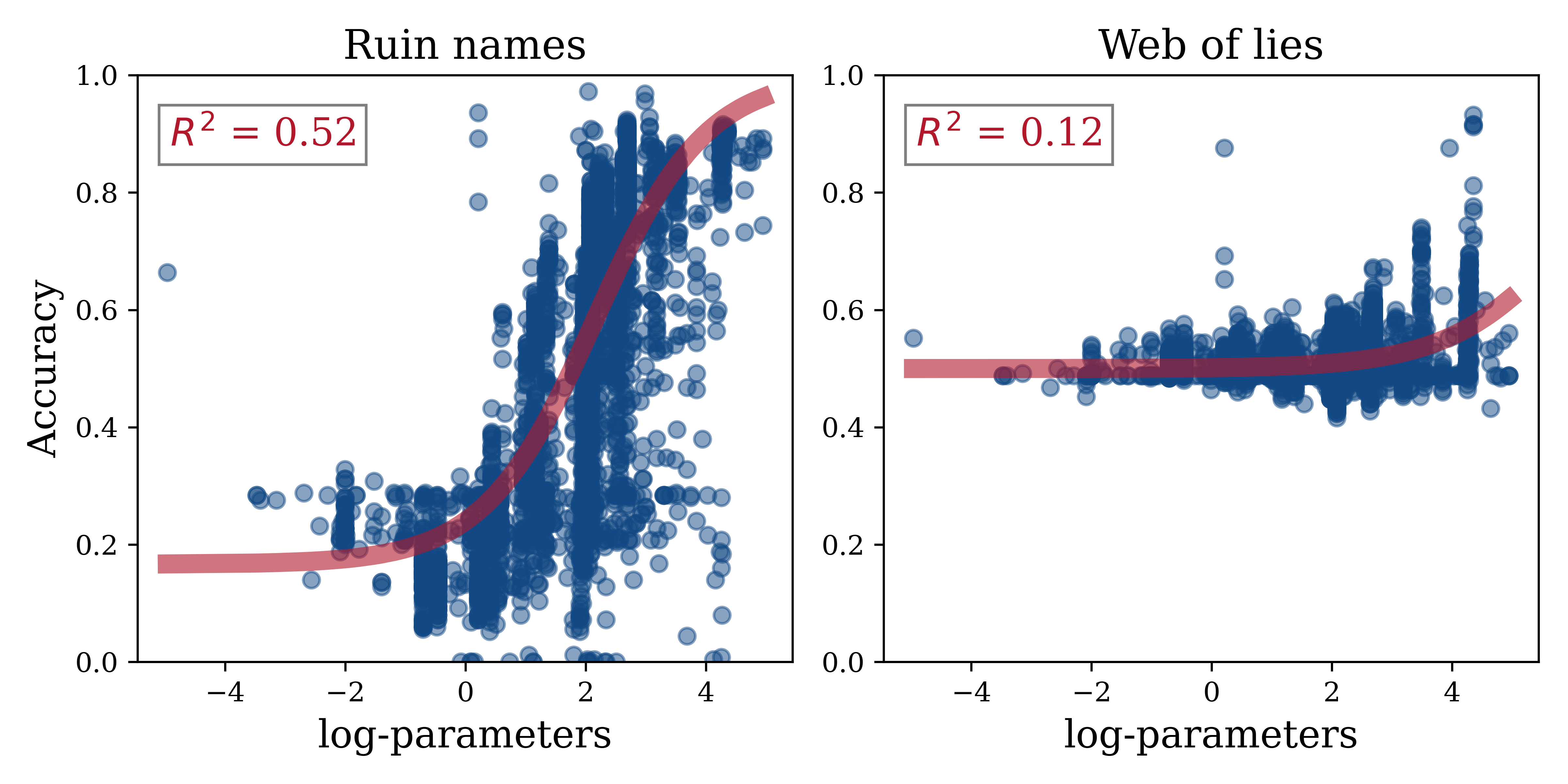}
    \caption[Logistic scaling law fits for two subtasks]{\textbf{Logistic scaling law fits for two subtasks}. Ruin names has the highest observed correlation with the logistic fit ($R^2 = 0.52$); Web of lies has the lowest correlation ($R^2 = 0.12$).}
    \label{fig:side_by_side}
\end{figure}

Next, to observe the baseline of \citet{owen2024predictable}'s logistic scaling law model on my dataset, I fit \cref{eq:logistic_scaling_law_model} to each collection of raw subtask scores $\mB_{i, :}$. $R^2$ values range from very mild values—$0.12$ for the Web of lies task—to moderate values—$0.52$ for the Ruin names task. Tasks also show a spread of different difficulty $\beta_i$ and discrimination coefficients $\alpha_i$, demonstrating different rates of improvement from scale. \cref{fig:side_by_side} provides a side-by-side comparison of Web of lies and Ruin names, and \cref{app:all_logistic_correlations} displays the logistic fits for all $19$ subtasks.

\subsubsection*{Inter-item residual correlations}

\begin{figure}[p]
    \centering
    \includegraphics[width=\linewidth]{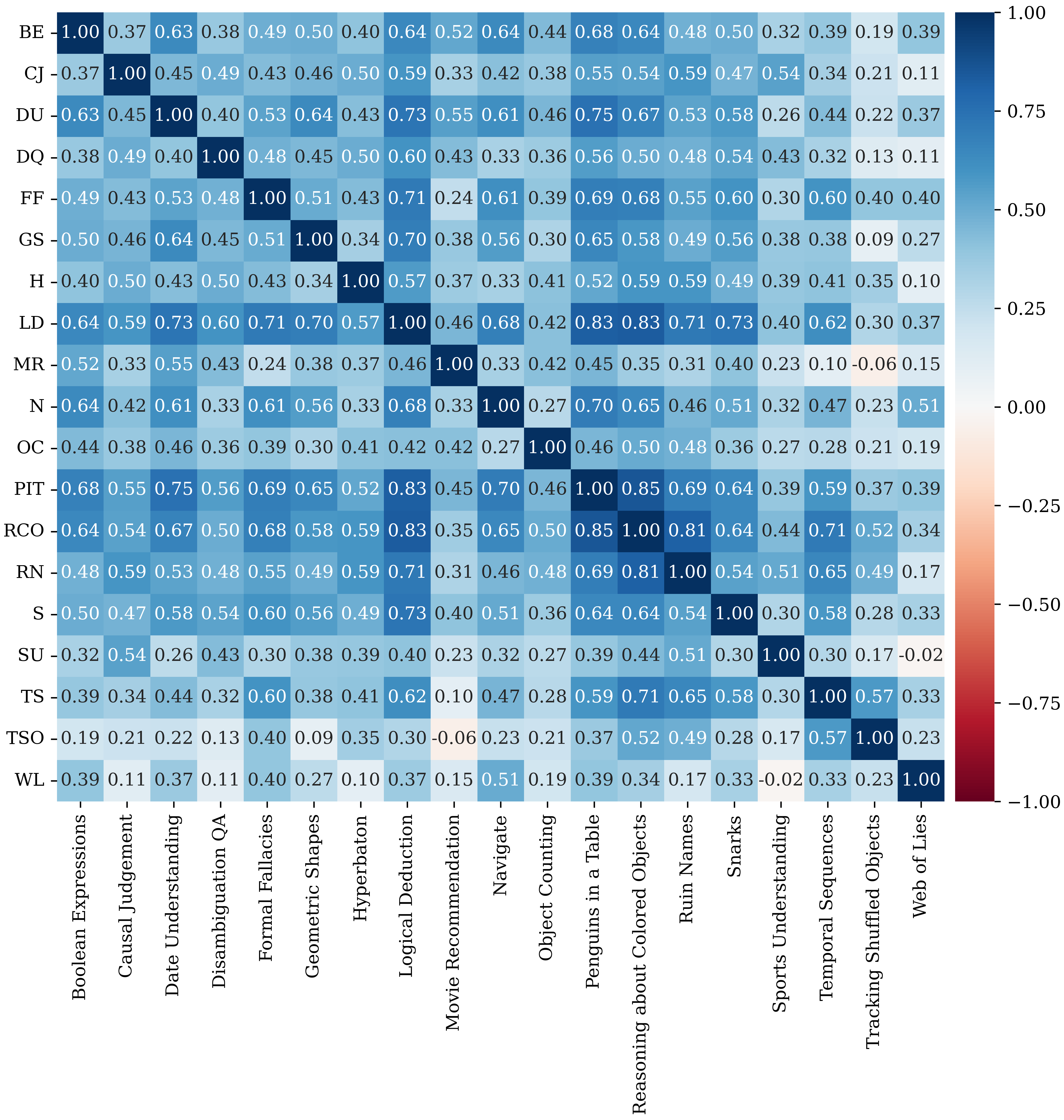}
    \caption[\texttt{BBH} score residuals correlation matrix]{\textbf{\texttt{BBH} score residuals correlation matrix}. Score residuals are taken from the logistic fits against the $\log$-parameter size displayed in \cref{fig:subtask_performances_logistic_fit}. The correlation metric is Spearman's rank correlation coefficient, $\rho$. The labels along the $y$-axis are acronym abbreviations for the full subtask names, which are given along the $x$-axis.}
    \label{fig:residual_correlation_matrix}
\end{figure}

My next step investigates correlations between benchmarks alongside the impact of model scale. I reinvestigate the inter-item correlation by rank-correlating the residuals of the estimates from \cref{eq:logistic_scaling_law_model}. In subtracting the best-fit explanation of performance according to $\log$-parameter size, I can effectively control for the parameter size, and can informally depict the residual relationship between subtasks. I display the updated correlation matrix in \cref{fig:residual_correlation_matrix}. The average inter-item correlation reduces from $\overline{\rho} = 0.64$ to $\overline{\rho} = 0.48$, and some item pairs are now slightly anti-correlated. However, $\overline{\rho} = 0.48$ remains a moderately positive correlation manifold.


\subsubsection*{Parallel analysis for the number of factors}

\begin{figure}[p]
    \centering
    \subfigure[Scree plot for raw score matrix, recommending to retain $5$ factors.]{
        \includegraphics[width=0.99\linewidth]{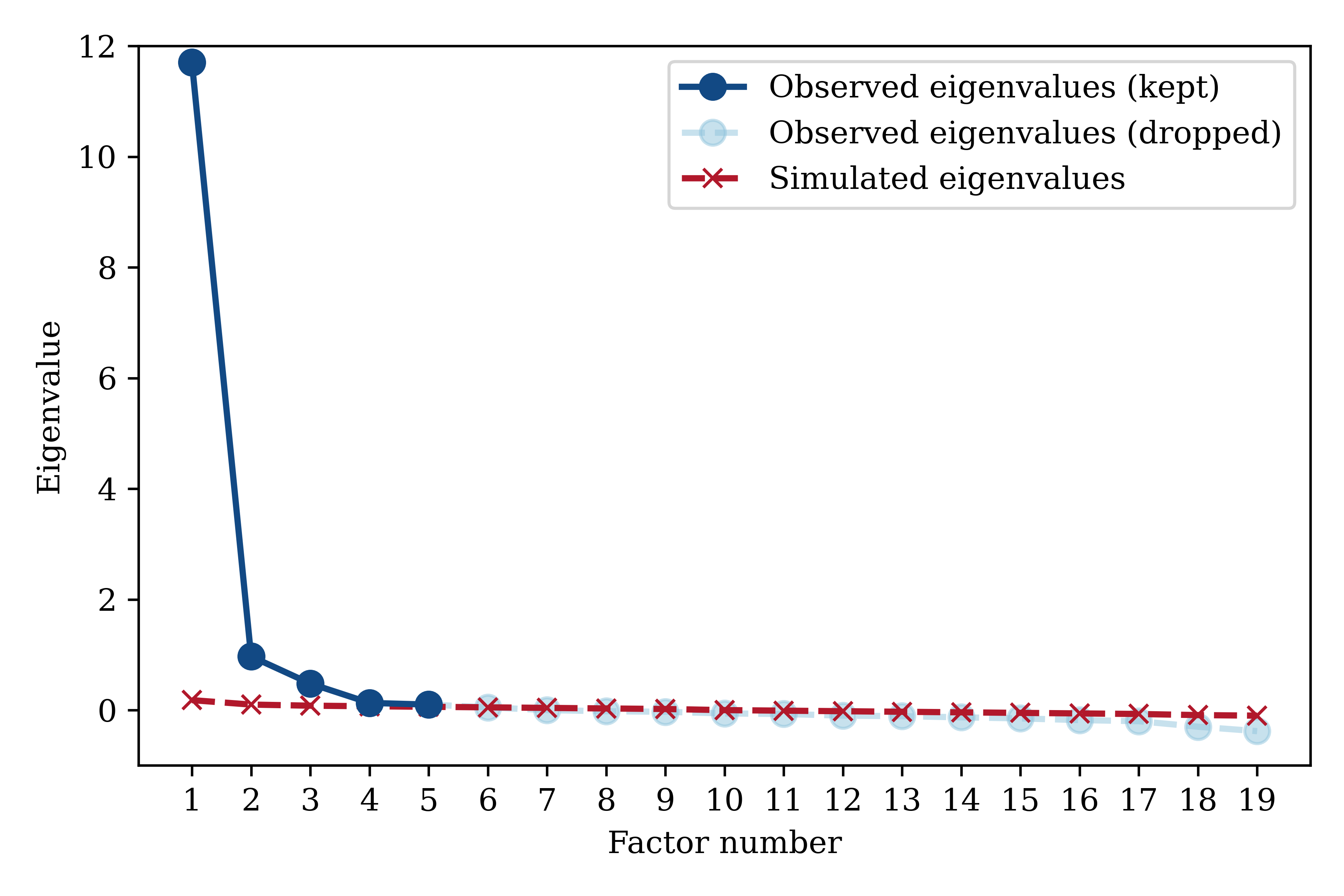}
    }
    \subfigure[Scree plot for score residuals after controlling for parameter size, recommending to retain $6$ factors.]{
        \includegraphics[width=0.99\linewidth]{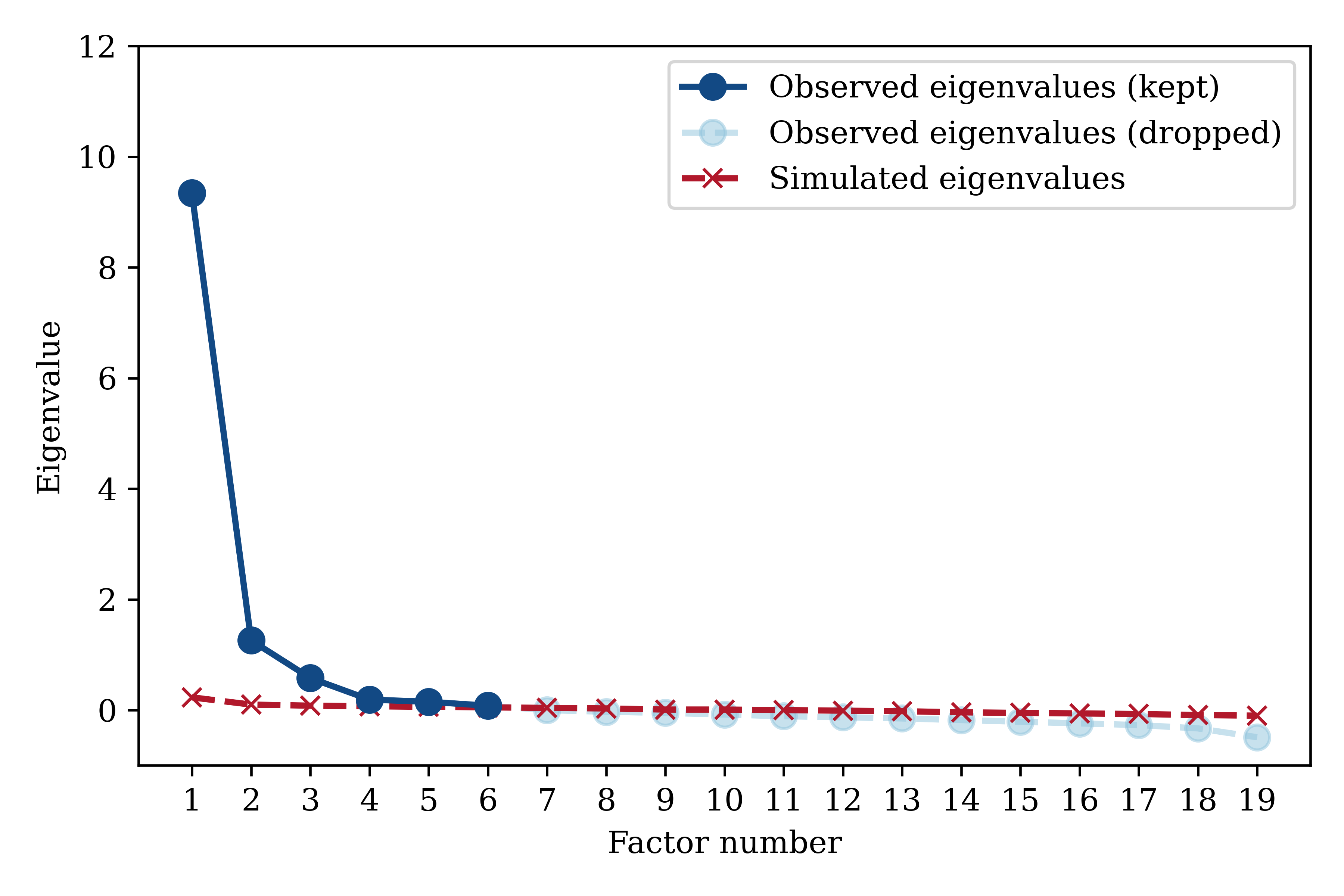}
    }
    \caption[Scree plots for \texttt{BBH} score matrices]{\textbf{Scree plots for \texttt{BBH} score matrices}.}
    \label{fig:scree_plots}
\end{figure}

I next conduct Horn's parallel analysis on the raw score and residual matrices from the previous processing steps. Horn's parallel analysis compares the eigenvalues of each real data matrix to the eigenvalues of Monte Carlo-simulated matrices of the same size. On the assumption that the Monte Carlo-simulated eigenvalues represent a component of the variance attributable to sampling error, subtracting these and retaining the positive ranks can provide an estimate for the correct number of latent factors for the dataset~\citep{hornRationaleTestNumber1965}. In other words, this process calculates an estimate for the `dimensionality' of the benchmark data, as in the number of separable factors that may be contributing to observed scores. I utilise this technique to determine $k$, the number of latent factors my models will estimate. \citet{burnellRevealingStructureLanguage2023} uses the same technique to determine their number of latent factors. As pictured in \cref{fig:scree_plots}~(a), this analysis suggests the retention of $5$ factors for the raw scores. In \cref{fig:scree_plots}~(b), the residual data provides support for an additional sixth factor, and the principal eigenvalue drops from $11.70$ to $9.34$.

Since the sixth factor for the score residuals is marginally above the level of the sixth Monte Carlo-simulated eigenvalue, and since not all of my models will control for parameter size, I select $k=5$ as the number of latent factors for all subsequent experiments.

\subsection{Comparison to exploratory factor analysis}\label{sec:results_efa_comparison}

\begin{table}[t]
{
    \centering
    \small
    \begin{tabular}{ccc|cccccc}\toprule
        \bf Eq. & \bf Str. & \bf Log. & \bf $\chi^2$ $\downarrow$ & \bf CFI $\uparrow$ & \bf RMSEA $\downarrow$ & \bf SRMR $\downarrow$ & \bf AIC $\downarrow$ & \bf BIC $\downarrow$ \\ \midrule
        (\ref{eq:exp_a_1}) & \xmark & \xmark & \colorbox{white}{2984} & \colorbox{white}{0.9737} & \colorbox{white}{0.0876} & \colorbox{white}{0.0134} & \colorbox{white}{-217242} & \colorbox{white}{-216578} \\
        (\ref{eq:exp_a_2}) & \cmark & \xmark & \colorbox{white}{3615} & \colorbox{white}{0.9680} & \colorbox{white}{0.0894} & \colorbox{white}{0.0163} & \colorbox{cmap_blue}{\bf -220369} & \colorbox{cmap_blue}{\bf -219673} \\\midrule
        (\ref{eq:exp_a_3}) & \xmark & \cmark & \colorbox{cmap_blue}{\bf 1547} & \colorbox{cmap_blue}{\bf 0.9836} & \colorbox{cmap_blue}{\bf 0.0623} & \colorbox{cmap_blue}{\bf 0.0110} & \colorbox{white}{222038} & \colorbox{white}{222703} \\
        (\ref{eq:exp_a_4}) & \cmark & \cmark & \colorbox{white}{2127} & \colorbox{white}{0.9782} & \colorbox{white}{0.0679} & \colorbox{white}{0.0135} & \colorbox{cmap_blue}{\bf 218743} & \colorbox{cmap_blue}{\bf 219439} \\
        \bottomrule
    \end{tabular}
    \caption[Fit measures for the four tested EFA models]{\textbf{Fit measures for the four tested EFA models}. The \textbf{Str.} column indicates whether the model includes model parameter size $\log(n)$ as a structural regression parameter. The \textbf{Log.} column indicates whether the dataset uses the transformed matrix $\mB'$ in place of $\mB$. The fourth row, with \cref{eq:exp_a_4}, shows the structured capabilities model. Highlighted boldface values for $\chi^2$, $\mathrm{CFI}$, $\mathrm{RMSEA}$, and $\mathrm{SRMR}$ indicate the best overall model, since these absolute model fit metrics share the same range. Highlighted boldface values for $\mathrm{AIC}$ and $\mathrm{BIC}$ indicate the most parsimonious model for a given data matrix, either $\mB$ or $\mB'$, where the likelihood is comparable.}
    \label{tab:efa_results}
    }
\end{table}

For experiment A, \cref{tab:efa_results} summarises the fit statistics for the four models I compare. All four models provide acceptable fit according to the conventional interpretation from \citet{huCutoffCriteriaFit1999}, with $\mathrm{CFI} \geq 0.95$, $\mathrm{RMSEA} \leq 0.10$, and $\mathrm{SRMR} \leq 0.06$. The best-fit model is (\ref{eq:exp_a_3}), which utilises the logit-transformed data $\mB'$ but not the structural parameter $\vw_n\log\vn$. Modelling the transformed data $\mB'$ in place of $\mB$ results in better fit statistics, as shown by the reduction in $\chi^2$, $\mathrm{RMSEA}$, and $\mathrm{SRMR}$ and the increase in $\mathrm{CFI}$ from models (\ref{eq:exp_a_1}) and (\ref{eq:exp_a_2}) to models (\ref{eq:exp_a_3}) and (\ref{eq:exp_a_4}). The fit measures are better for pure measurement models (\ref{eq:exp_a_1}) and (\ref{eq:exp_a_3}) compared to their structured counterparts (\ref{eq:exp_a_2}) and (\ref{eq:exp_a_4}). However, in  both cases, the structured models perform better according to parsimonious fit indices $\mathrm{AIC}$ and $\mathrm{BIC}$.

\begin{figure}[p]
    \centering
    \subfigure[Heatmap of pattern coefficients per factor.]{
        \includegraphics[width=0.98\textwidth]{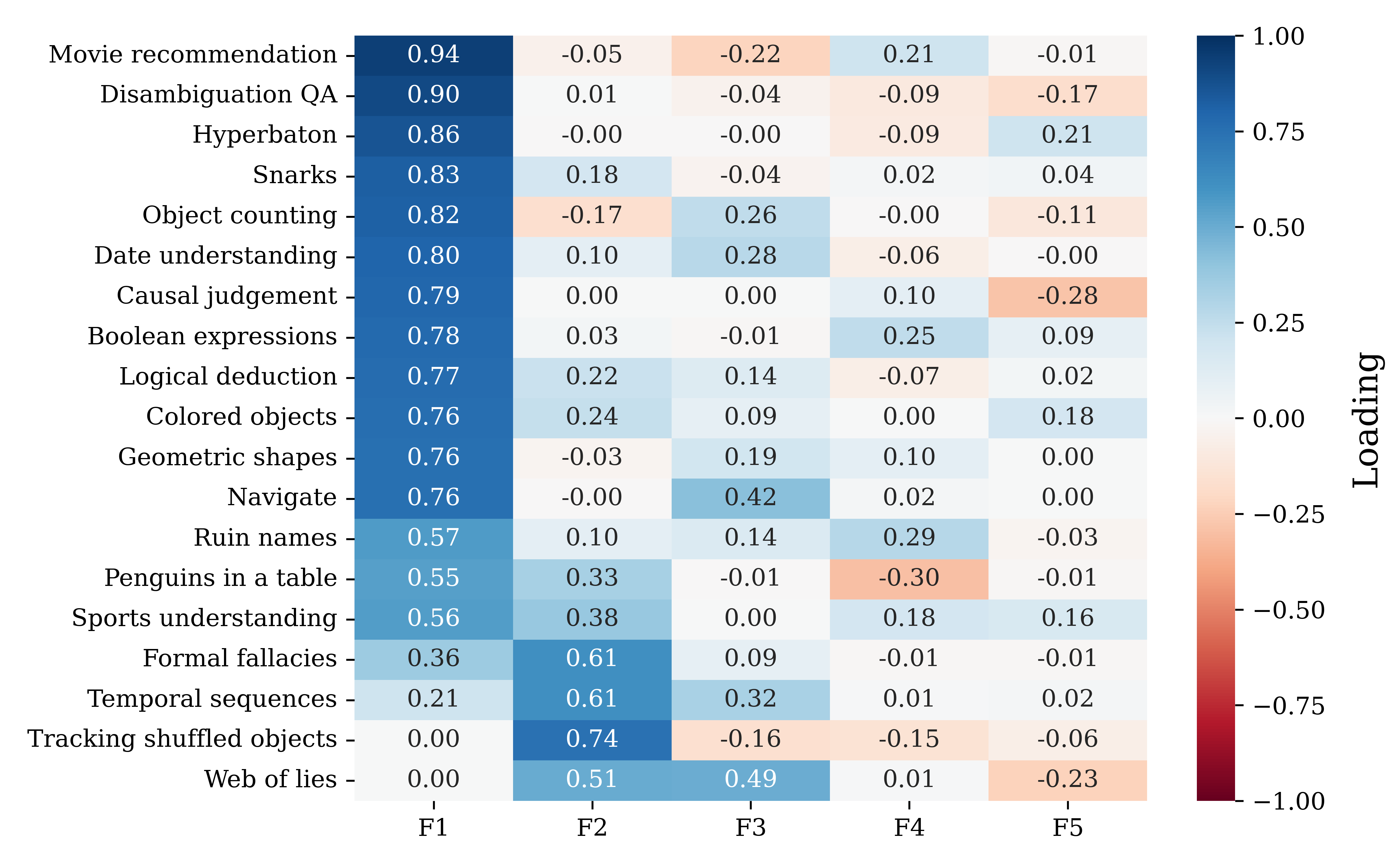}
    }
    \subfigure[Proportions of communal variance explained by each factor.]{
        \includegraphics[width=0.98\textwidth]{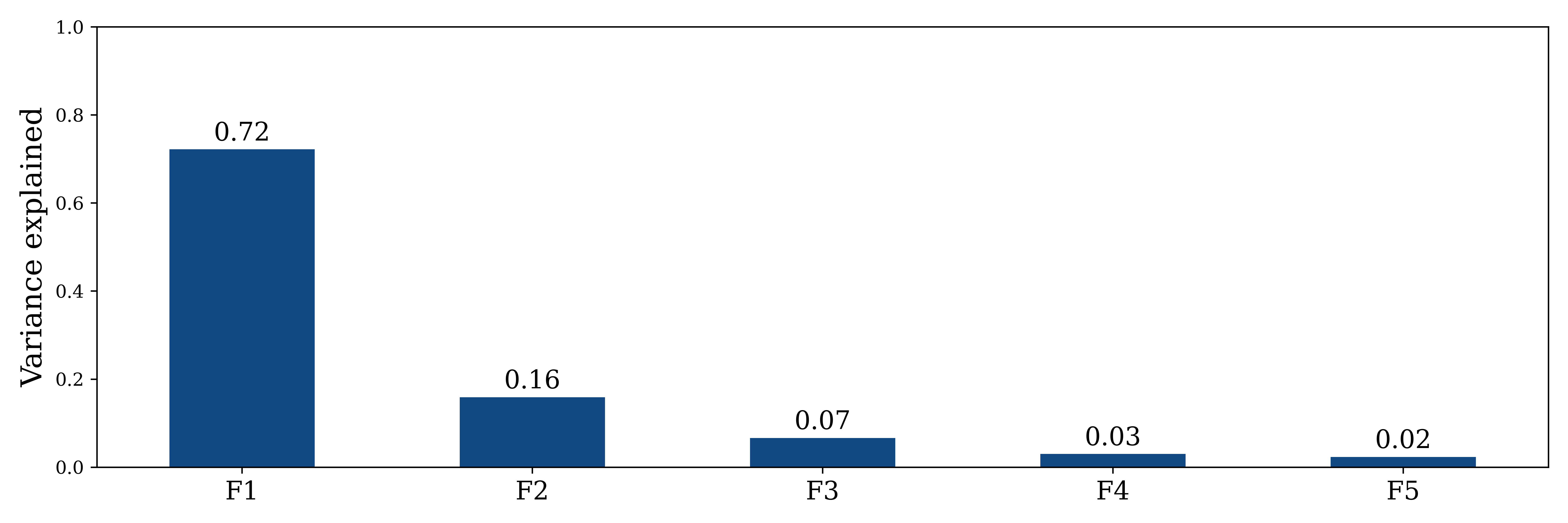}
    }
    \subfigure[Logistic relationship between $\log$-parameter size and factor scores for each factor.]{
        \includegraphics[width=0.98\textwidth]{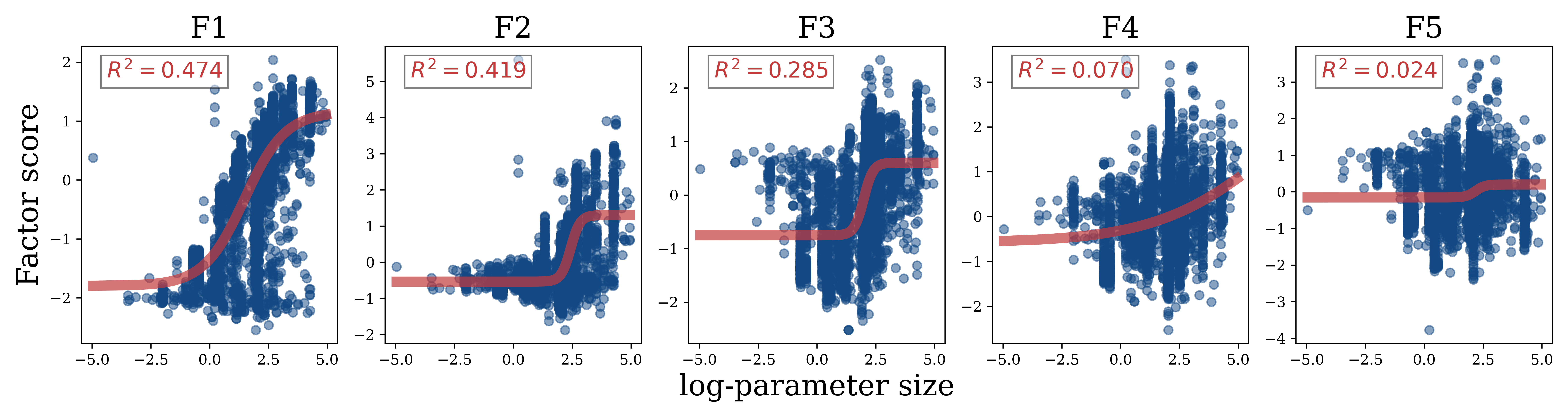}
    }
    \caption[Linear latent factor model results]{\textbf{Linear latent factor model results}. This model corresponds to \cref{eq:exp_a_1} from Experiment A. Factor 1 dominates, explaining the largest proportion of the total variance ($0.72$) and correlating strongest with parameter size ($R^2 = 0.474$). Factor 2 loads strongest on some more challenging subtasks (like Tracking shuffled objects) and also correlates strongly with model size ($R^2 = 0.419$).}
    \label{fig:factor_comparisons_1}
\end{figure}

\begin{figure}[p]
    \centering
    \subfigure[Heatmap of pattern coefficients per factor.]{
        \includegraphics[width=0.98\textwidth]{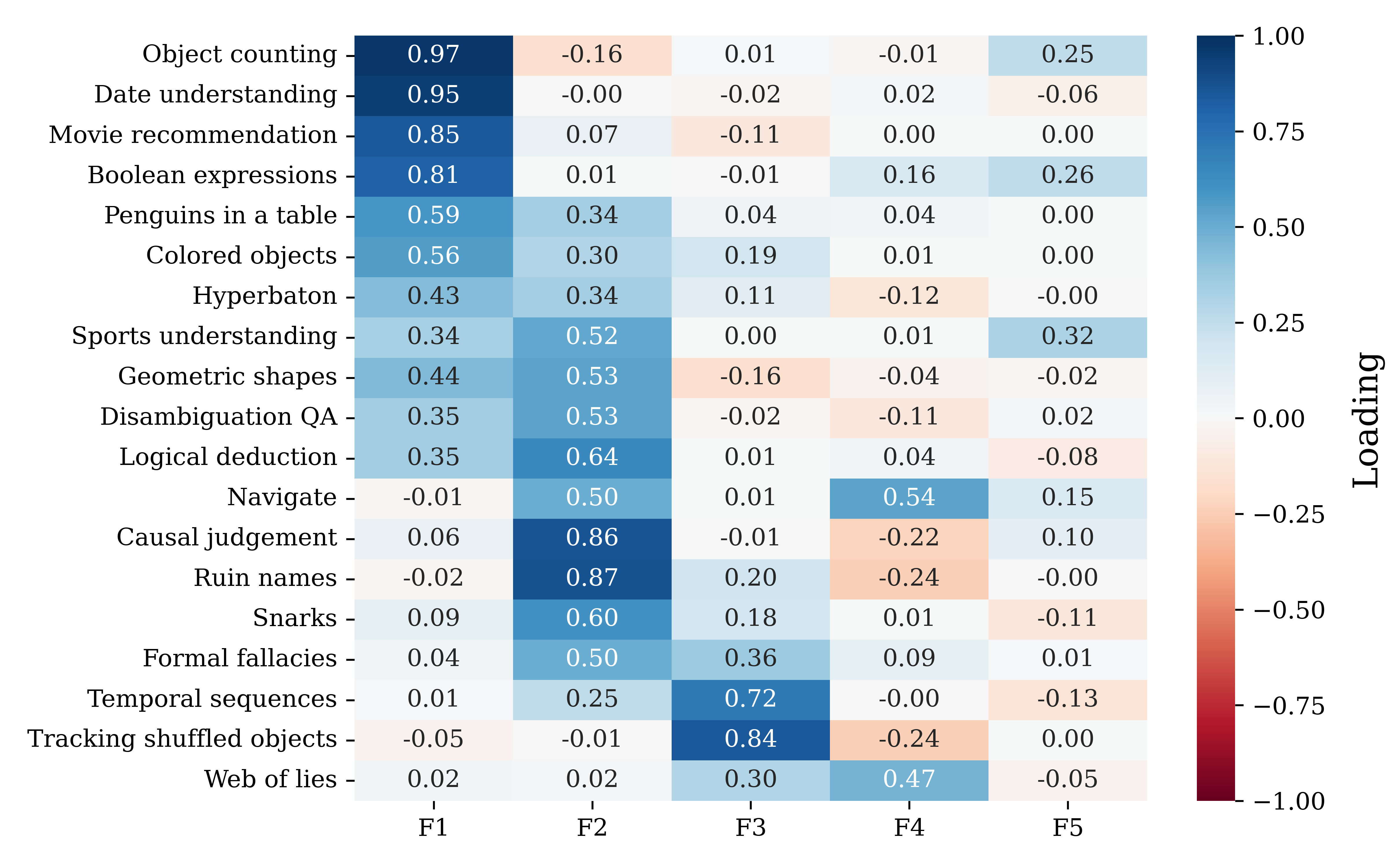}
    }
    \subfigure[Proportions of communal variance explained by each factor.]{
        \includegraphics[width=0.98\textwidth]{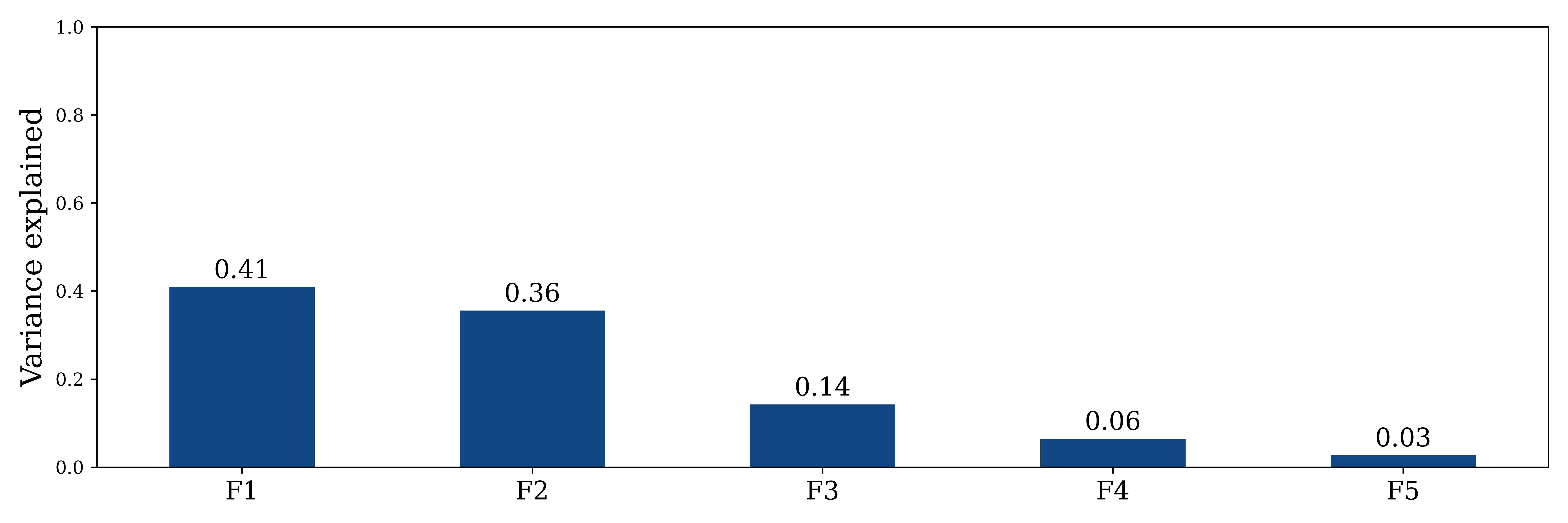}
    }
    \subfigure[Logistic relationship between $\log$-parameter size and factor scores for each factor.]{
        \includegraphics[width=0.98\textwidth]{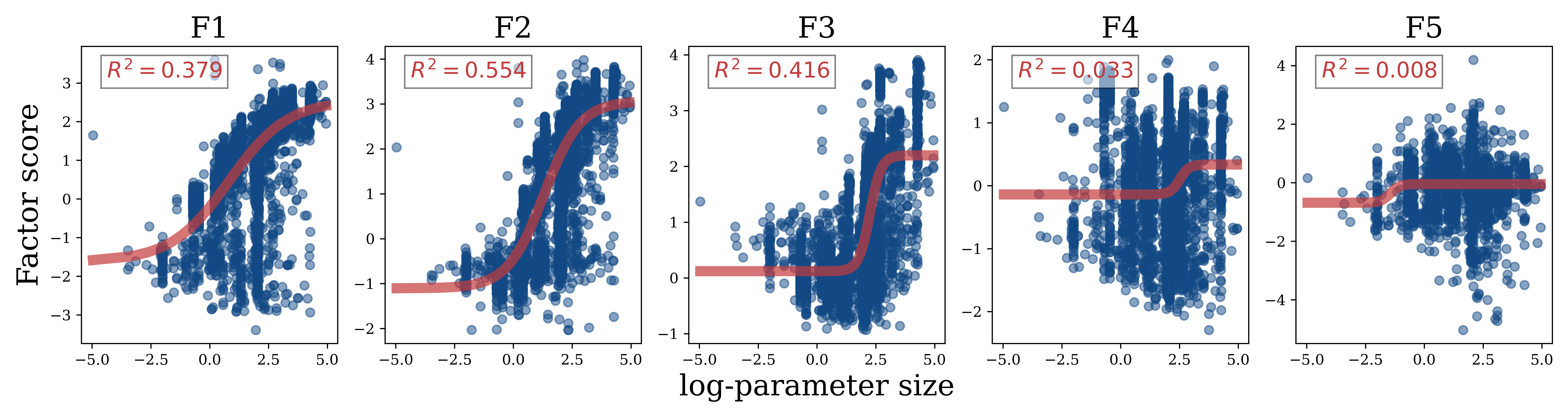}
    }
    \caption[Structured capabilities model results]{\textbf{Structured capabilities model results}. This model corresponds to \cref{eq:exp_a_4} from Experiment A. Factors 1 and 2 share appreciable proportions of the total variance ($0.41$ and $0.36$), followed by factor 3 ($0.14$). All three of these factors correlate strongly with parameter size ($R^2 = 0.379, 0.554, 0.416$).}
    \label{fig:factor_comparisons_4}
\end{figure}

\Cref{fig:factor_comparisons_1,fig:factor_comparisons_4} provide a visual comparison of the parameter estimates for models (\ref{eq:exp_a_1}) and (\ref{eq:exp_a_4}). \cref{eq:exp_a_1} fits a single dominant factor explaining $72\%$ of the communal variance. This factor has the largest loading coefficients for $15$ of the $19$ subtasks. Per model, factor scores for this factor correlate moderately with the $\log$-parameter size ($R^2 = 0.474$, according to a logistic relationship). The next largest factor has the largest coefficients for the remaining $4$ of $19$ subtasks, explains $16\%$ of the communal variance, and also correlates moderately with the $\log$-parameter size ($R^2 = 0.419$). In contrast, \cref{eq:exp_a_4} fits two large factors explaining $41\%$ and $36\%$ of the communal variance respectively. These factor scores also correlate with parameter size ($R^2 = 0.379$ and $R^2 = 0.554$). The third largest factor loads on similar subtasks to the secondary factor from model (\ref{eq:exp_a_1}). This third factor explains $14\%$ of communal variance and correlates moderately with model size ($R^2 = 0.416$).\footnote{The order of the factors is arbitrary, so as a visual aid, I order them by descending variance explained.}

Model (\ref{eq:exp_a_3}) is not depicted, though it learns a similar factor structure and coefficients to (\ref{eq:exp_a_4}), which is the same logistic model with the structural model size parameter. Model (\ref{eq:exp_a_3})'s three dominant factors explain $41\%$, $37\%$, and $14\%$ of the communal variance, and correlate with model size at $R^2 = 0.534, 0.372$, and $0.385$. Like model (\ref{eq:exp_a_4}), this model's two remaining factors have minimal correlation with model size and minimal communal variance explained.

\subsection{Comparison to observational scaling laws}\label{sec:results_scaling_law_comparison}

\begin{table}[t]
    \centering
    \small
    \begin{tabular}{l|rrr|rrr}
        \toprule
        & \multicolumn{3}{c|}{$\mathrm{MSE}_\mathrm{train}$} & \multicolumn{3}{c}{$\mathrm{MSE}_\mathrm{test}$} \\
        \bf Withheld subtask & \multicolumn{1}{c}{\bf SC} & \multicolumn{1}{c}{\bf OSL} & \multicolumn{1}{c|}{\bf Size} & \multicolumn{1}{c}{\bf SC} & \multicolumn{1}{c}{\bf OSL} & \multicolumn{1}{c}{\bf Size} \\
        \midrule
        Boolean expressions & \colorbox{cmap_blue}{\bf 1.056} & \colorbox{white}{1.153} & \colorbox{white}{2.487} & \colorbox{cmap_blue}{\bf 0.206} & \colorbox{white}{0.239} & \colorbox{white}{1.144} \\
        Causal judgement & \colorbox{white}{0.476} & \colorbox{cmap_blue}{\bf 0.469} & \colorbox{white}{0.846} & \colorbox{white}{0.338} & \colorbox{cmap_blue}{\bf 0.293} & \colorbox{white}{0.687} \\
        Date understanding & \colorbox{white}{0.464} & \colorbox{cmap_blue}{\bf 0.462} & \colorbox{white}{1.419} & \colorbox{white}{0.250} & \colorbox{cmap_blue}{\bf 0.149} & \colorbox{white}{1.022} \\
        Disambiguation QA & \colorbox{white}{1.536} & \colorbox{cmap_blue}{\bf 1.519} & \colorbox{white}{2.592} & \colorbox{cmap_blue}{\bf 0.608} & \colorbox{white}{0.680} & \colorbox{white}{1.764} \\
        Formal fallacies & \colorbox{cmap_blue}{\bf 0.961} & \colorbox{white}{0.984} & \colorbox{white}{1.490} & \colorbox{cmap_blue}{\bf 0.862} & \colorbox{white}{1.092} & \colorbox{white}{4.411} \\
        Geometric shapes & \colorbox{cmap_blue}{\bf 0.937} & \colorbox{white}{0.965} & \colorbox{white}{1.891} & \colorbox{cmap_blue}{\bf 0.370} & \colorbox{white}{0.467} & \colorbox{white}{1.277} \\
        Hyperbaton & \colorbox{white}{0.994} & \colorbox{cmap_blue}{\bf 0.993} & \colorbox{white}{1.921} & \colorbox{white}{2.472} & \colorbox{cmap_blue}{\bf 2.339} & \colorbox{white}{4.394} \\
        Logical deduction & \colorbox{cmap_blue}{\bf 0.147} & \colorbox{white}{0.156} & \colorbox{white}{0.945} & \colorbox{cmap_blue}{\bf 0.096} & \colorbox{white}{0.103} & \colorbox{white}{1.326} \\
        Movie recommendation & \colorbox{cmap_blue}{\bf 0.348} & \colorbox{white}{0.362} & \colorbox{white}{0.964} & \colorbox{cmap_blue}{\bf 0.958} & \colorbox{white}{0.966} & \colorbox{white}{1.328} \\
        Navigate & \colorbox{cmap_blue}{\bf 1.877} & \colorbox{white}{2.034} & \colorbox{white}{2.617} & \colorbox{cmap_blue}{\bf 0.504} & \colorbox{white}{1.260} & \colorbox{white}{3.926} \\
        Object counting & \colorbox{cmap_blue}{\bf 0.707} & \colorbox{white}{0.737} & \colorbox{white}{1.336} & \colorbox{white}{0.661} & \colorbox{cmap_blue}{\bf 0.470} & \colorbox{white}{0.783} \\
        Penguins in a table & \colorbox{cmap_blue}{\bf 0.423} & \colorbox{white}{0.435} & \colorbox{white}{1.455} & \colorbox{cmap_blue}{\bf 0.082} & \colorbox{white}{0.094} & \colorbox{white}{1.112} \\
        Reasoning about colored objects & \colorbox{cmap_blue}{\bf 0.324} & \colorbox{white}{0.331} & \colorbox{white}{1.241} & \colorbox{white}{0.146} & \colorbox{cmap_blue}{\bf 0.096} & \colorbox{white}{1.682} \\
        Ruin names & \colorbox{cmap_blue}{\bf 0.902} & \colorbox{white}{1.169} & \colorbox{white}{2.008} & \colorbox{cmap_blue}{\bf 0.351} & \colorbox{white}{0.974} & \colorbox{white}{1.816} \\
        Snarks & \colorbox{cmap_blue}{\bf 1.248} & \colorbox{white}{1.274} & \colorbox{white}{2.106} & \colorbox{white}{0.528} & \colorbox{cmap_blue}{\bf 0.527} & \colorbox{white}{3.260} \\
        Sports understanding & \colorbox{cmap_blue}{\bf 1.099} & \colorbox{white}{1.197} & \colorbox{white}{2.433} & \colorbox{white}{1.612} & \colorbox{cmap_blue}{\bf 1.379} & \colorbox{white}{2.041} \\
        Temporal sequences & \colorbox{cmap_blue}{\bf 1.691} & \colorbox{white}{1.746} & \colorbox{white}{3.068} & \colorbox{cmap_blue}{\bf 5.879} & \colorbox{white}{10.087} & \colorbox{white}{25.289} \\
        Tracking shuffled objects & \colorbox{cmap_blue}{\bf 0.681} & \colorbox{white}{0.717} & \colorbox{white}{0.907} & \colorbox{cmap_blue}{\bf 1.205} & \colorbox{white}{1.278} & \colorbox{white}{3.310} \\
        Web of lies & \colorbox{cmap_blue}{\bf 0.477} & \colorbox{white}{0.528} & \colorbox{white}{0.556} & \colorbox{cmap_blue}{\bf 4.835} & \colorbox{white}{5.500} & \colorbox{white}{7.084} \\
        \midrule
        \bf Average & \colorbox{cmap_blue}{\bf 0.860} & \colorbox{white}{0.907} & \colorbox{white}{1.699} & \colorbox{cmap_blue}{\bf 1.156} & \colorbox{white}{1.473} & \colorbox{white}{3.561} \\
        \bottomrule
    \end{tabular}
    \caption[Prediction errors for held-out subtasks]{\textbf{Prediction errors for held-out subtasks}. Each column shows the mean-squared error for predicting held-out subtasks with each different capability estimation method. \textbf{SC} refers to my structured capabilities model with latent factor capability estimates (\cref{eq:exp_b_latent}). \textbf{OSL} refers to the PCA decomposition approach from \citet{ruan2024observational}'s observational scaling laws (\cref{eq:exp_b_pca}). \textbf{Size} refers to \citet{owen2024predictable}'s logistic scaling laws using model parameter size (\cref{eq:exp_b_scale}). Highlighted boldface values indicate the lowest errors for each of $\mathrm{MSE}_\mathrm{train}$ and $\mathrm{MSE}_\mathrm{test}$.}
    \label{tab:heldout_prediction_accuracies}
\end{table}

\cref{tab:heldout_prediction_accuracies} lists the mean-squared prediction errors for the structured capability estimates (\cref{eq:exp_b_latent}), the observational scaling law capability estimates (\cref{eq:exp_b_pca}), and the pure scaling law model (\cref{eq:exp_b_scale}). We list the error on the training set $\vb_{p'}^\mathrm{train}$ as $\mathrm{MSE}_\mathrm{train}$, and on the test set $\vb_{p'}^\mathrm{test}$ as $\mathrm{MSE}_\mathrm{test}$. All three methods perform worse at predicting the held-out $\vb_{p'}^\mathrm{test}$ than they do on $\vb_{p'}^\mathrm{train}$. The pure scaling law model is the worst performing out-of-fold prediction model with an average $\mathrm{MSE}_\mathrm{train}$ of $1.699$ and an average $\mathrm{MSE}_\mathrm{test}$ of $3.561$. Between the structured capabilities and the observational scaling law capability estimates, the former approach outperforms the latter on $15$ of the $19$ heldout subtasks in terms of $\mathrm{MSE}_\mathrm{train}$. The same holds for $\mathrm{MSE}_\mathrm{test}$, where structured capability estimates achieve a lower prediction error on $12$ of $19$ held-out subtasks. The average $\mathrm{MSE}_\mathrm{train}$ is $0.860$ for the structured capabilities approach and $0.907$ for the observational scaling law approach, while the average $\mathrm{MSE}_\mathrm{test}$ is $1.156$ for the structured capability approach and $1.473$ for the observational scaling law approach. In both cases, the structured capabilities approach outperforms in terms of predictive accuracy, however, according to a two-sided $t$-test on these samples, the improvement is not statistically significant ($p = 0.776$ for $\mathrm{MSE}_\mathrm{train}$, $p = 0.637$ for $\mathrm{MSE}_\mathrm{test}$).

\begin{figure}[p]
    \centering
    \subfigure[Heatmap of principal component weights per subtask, including held-out Geometric shapes task. The weights for the held-out Geometric shapes task are estimated using the training set $\vb_{p'}^\mathrm{train}$.]{
        \includegraphics[width=\textwidth]{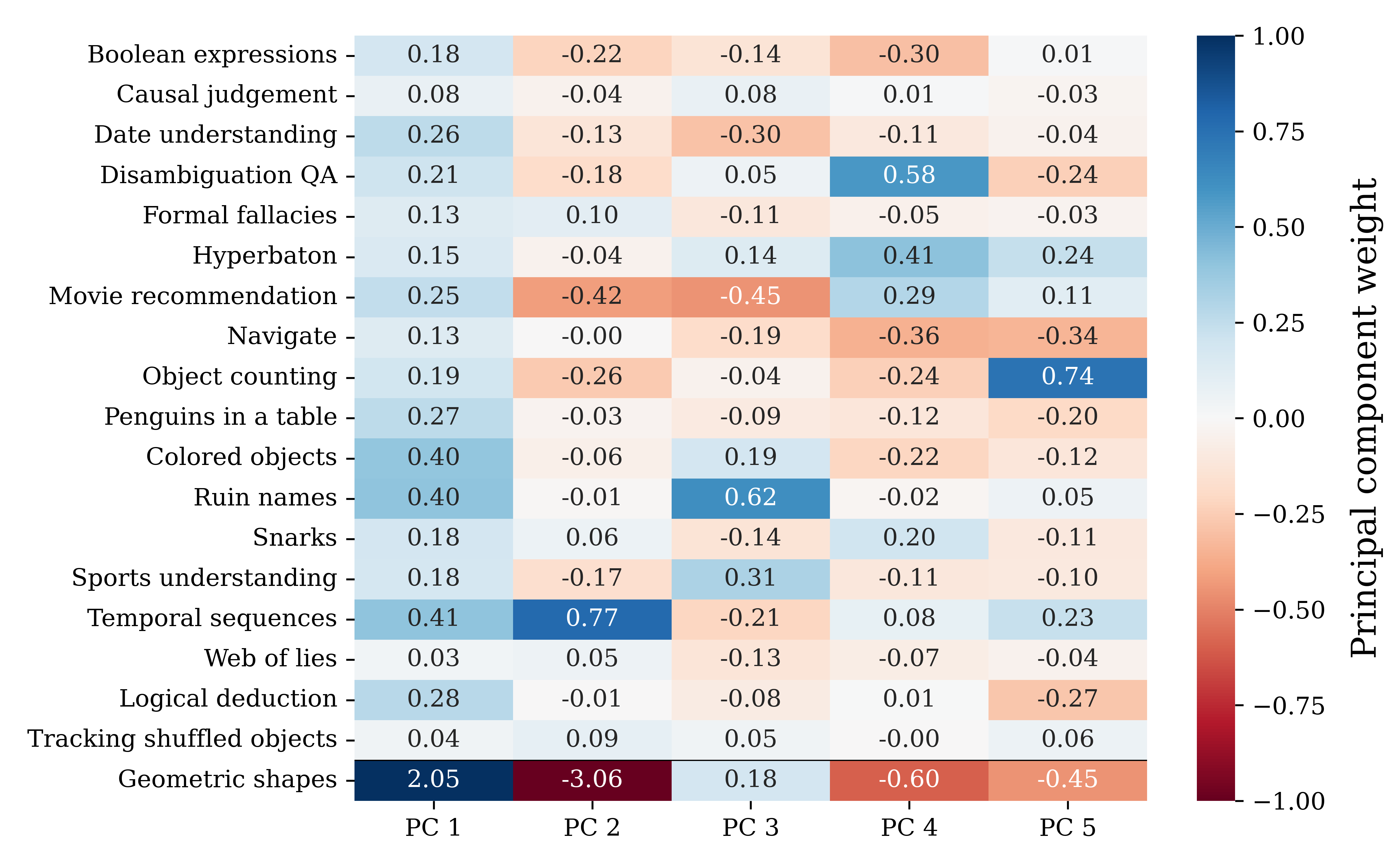}
    }
    \subfigure[Models' principal component scores against their raw Geometric Shapes subtask accuracy.]{
        \includegraphics[width=\textwidth]{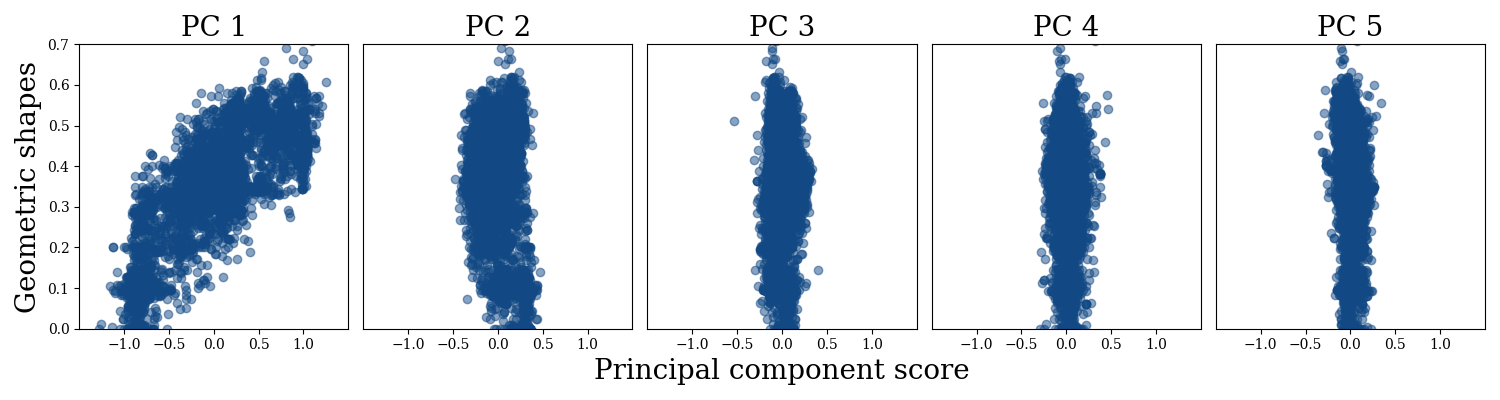}
    }
    \subfigure[Models' $\log$-parameter sizes against their principal component scores.]{
        \includegraphics[width=\textwidth]{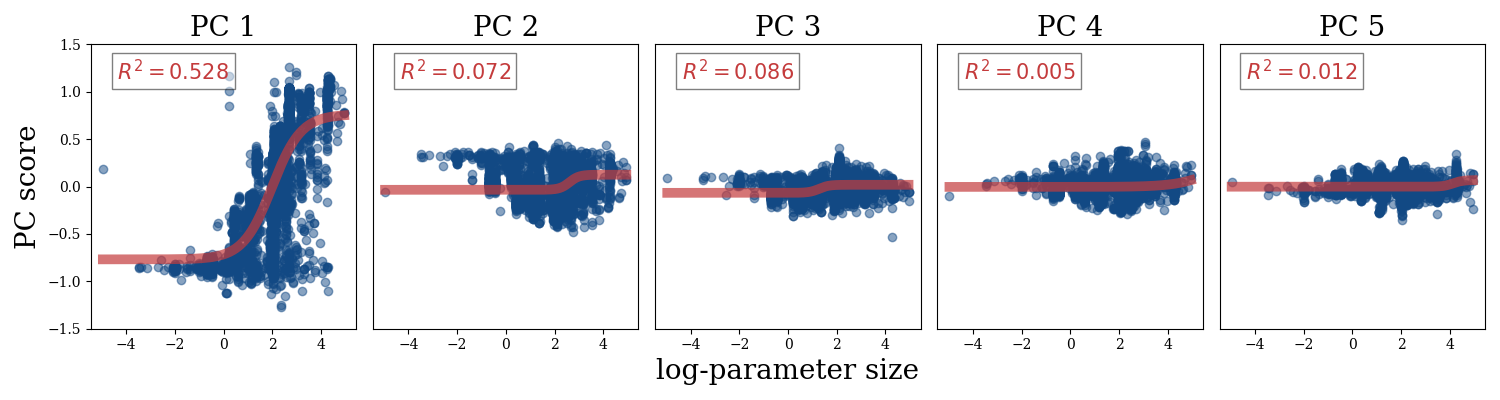}
    }
    \caption[Parameter estimates for the observational scaling law model]{\textbf{Parameter estimates for the observational scaling law model}. The heldout task to predict is Geometric shapes.}
    \label{fig:prediction_experiment_pca}
\end{figure}

\begin{figure}[p]
    \centering
    \subfigure[Heatmap of latent factor loadings per subtask. The factor loadings for the held-out Geometric shapes task are estimated using the training set $\vb_{p'}^\mathrm{train}$.]{
        \includegraphics[width=\textwidth]{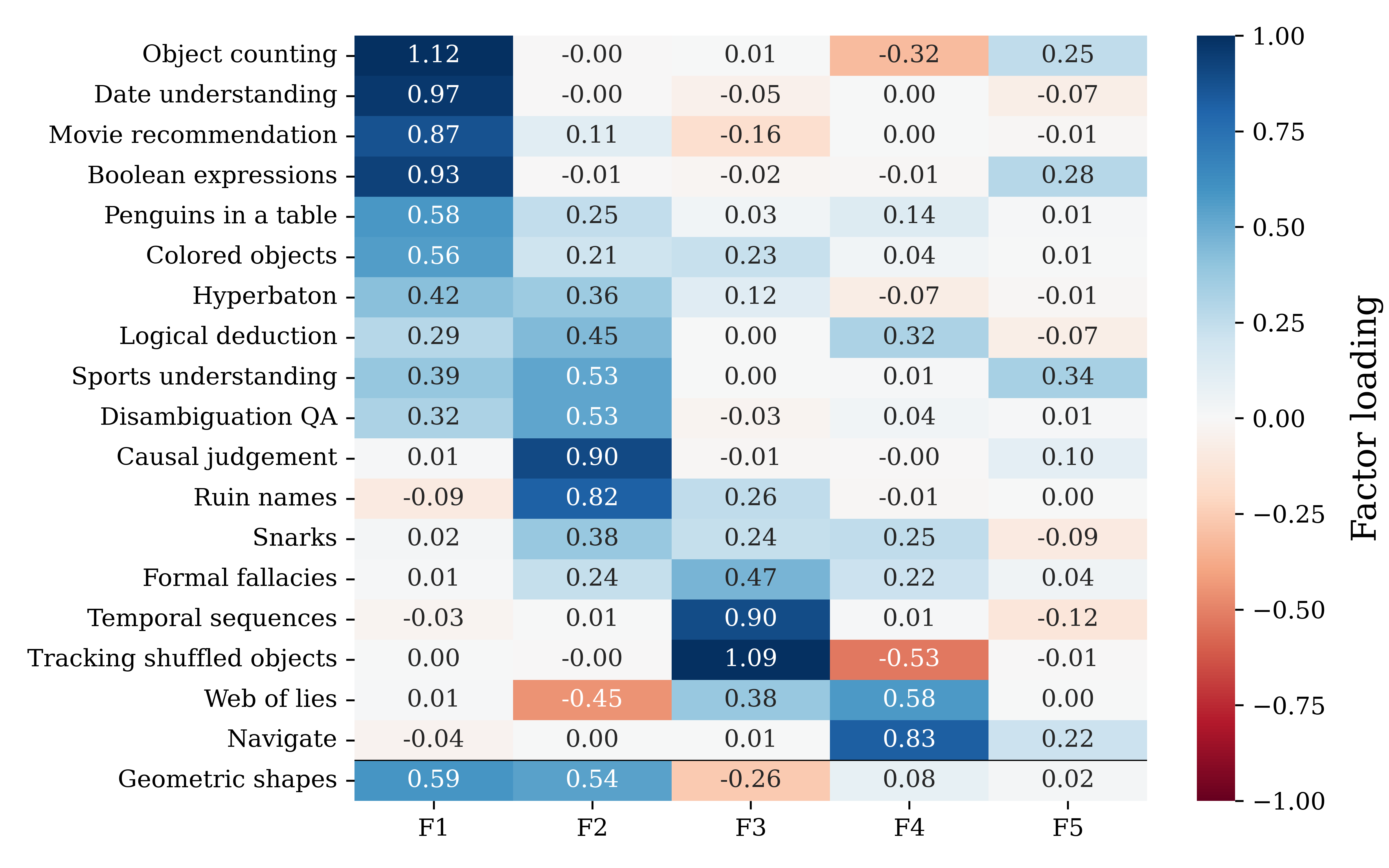}
    }
    \subfigure[Models' latent factor scores against their raw Geometric shapes subtask accuracy.]{
        \includegraphics[width=\textwidth]{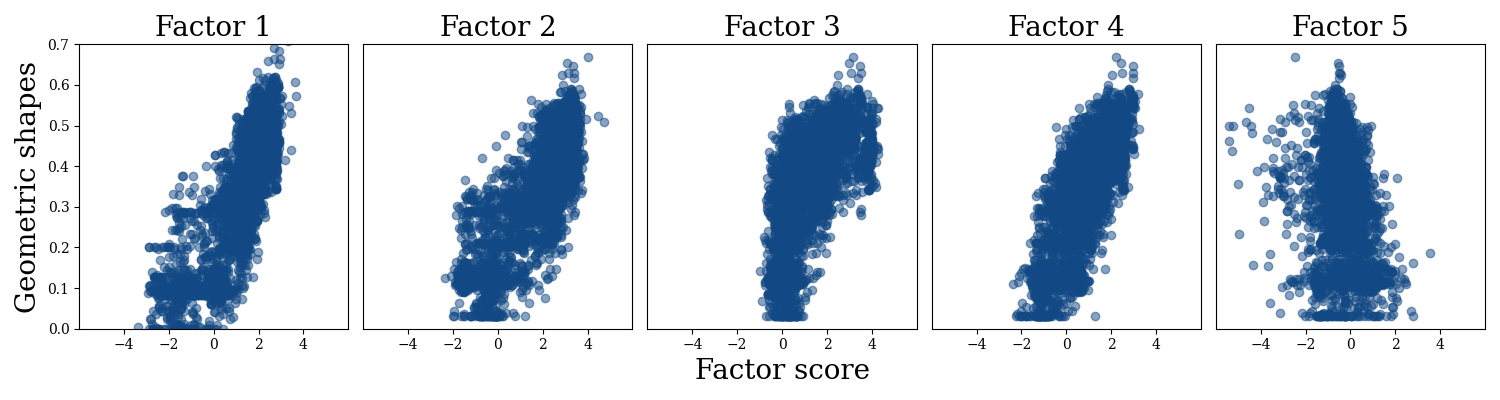}
    }
    \subfigure[Models' $\log$-parameter sizes against their latent factor scores.]{
        \includegraphics[width=\textwidth]{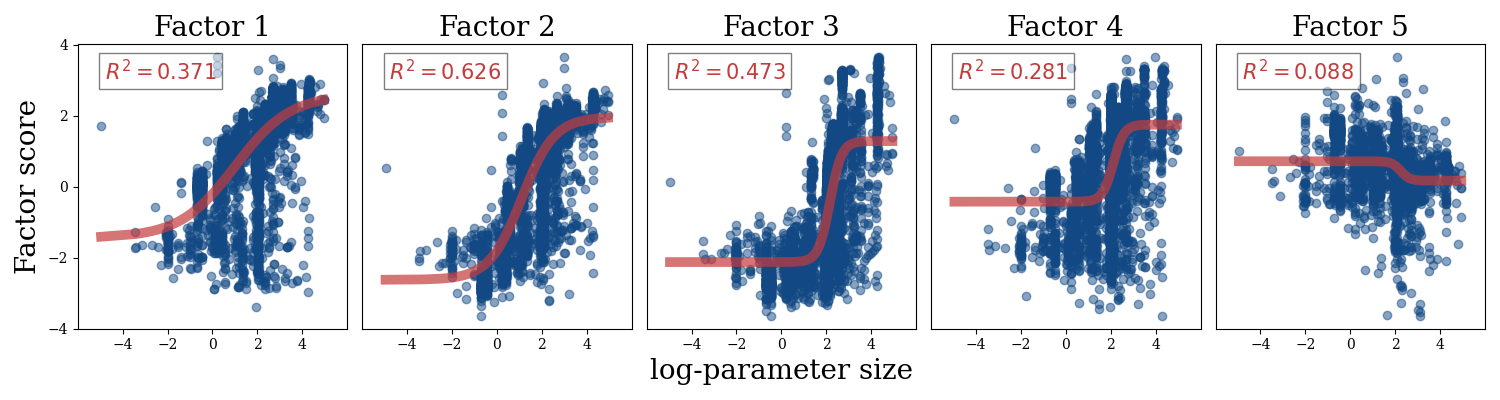}
    }
    \caption[Parameter estimates for the structured capabilities model]{\textbf{Parameter estimates for the structured capabilities model}. The held-out task to predict is Geometric shapes.}
    \label{fig:prediction_experiment_structural_regression}
\end{figure}

\Cref{fig:prediction_experiment_pca,fig:prediction_experiment_structural_regression} show the parameter estimates for each model when predicting the held-out Geometric shapes subtask. On this subtask, the structured capabilities model outperforms the observational scaling law model with $\mathrm{MSE}_\mathrm{test}$ values of $0.370$ and $0.467$, respectively. The observational scaling law model uses $k=5$ principal components as capability dimensions, though only the first principal component has a positive weight on all subtasks (\cref{fig:prediction_experiment_pca}~(a)), or exhibits meaningful variance compared to the Geometric shapes subtask accuracy (\cref{fig:prediction_experiment_pca}~(b)), or appears to bear a logistic relationship with $\log$-parameter size (\cref{fig:prediction_experiment_pca}~(c)). By comparison, the structured capabilities model estimates factors with clusters of positive loadings on distinct tasks (\cref{fig:prediction_experiment_structural_regression}~(a)). My model estimates that Geometric shapes loads similarly on the first and second factors, each of which shows a mild positive correlation with Geometric shapes scores (\cref{fig:prediction_experiment_structural_regression}~(b)). Finally, multiple factors show a logistic shape when compared with $\log$-parameter size (\cref{fig:prediction_experiment_structural_regression}~(c)).

\begin{figure}[t]
    \centering
    \subfigure[Proportion of variance explained per principal component.]{
        \includegraphics[width=\textwidth]{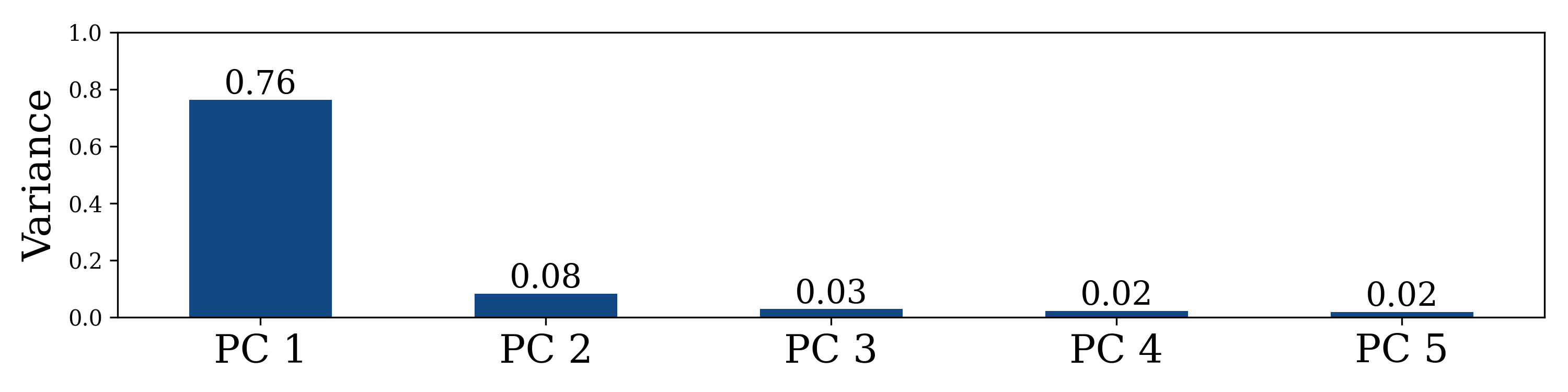}
    }
    \subfigure[Proportion of communal variance explained per latent factor, equivalent to the proportional sum of squared factor loadings.]{
        \includegraphics[width=\textwidth]{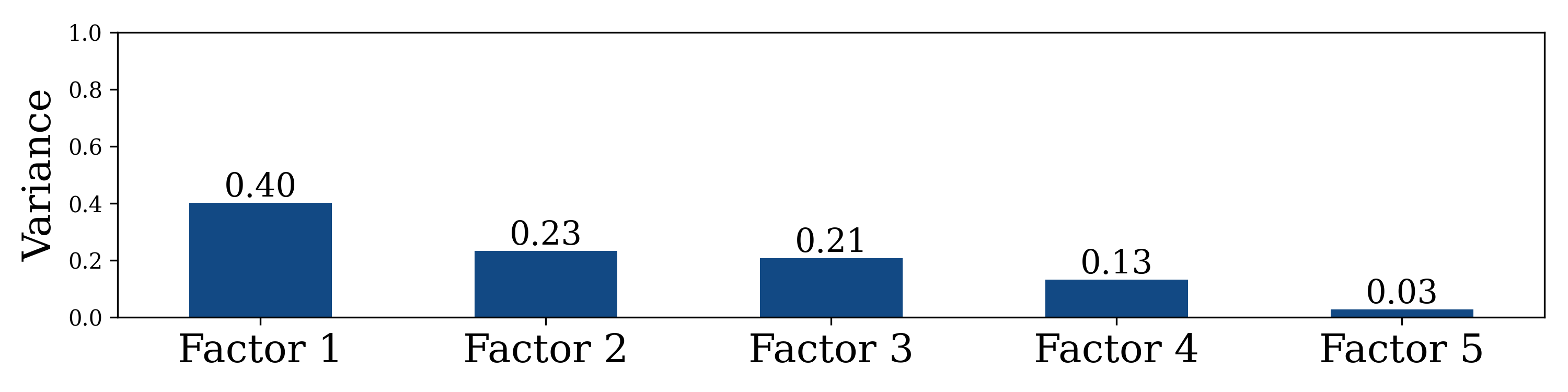}
    }
    \caption[Variance explained by estimated capabilities]{\textbf{Variance explained by estimated capabilities}. For both models the held-out task to predict is Geometric Shapes.}
    \label{fig:comparing_variance_explained}
\end{figure}

We also compare the proportions of variance explained in each model when predicting held-out Geometric shapes subtask scores (\cref{fig:comparing_variance_explained}). In the observational scaling law model, the first principal component dominates, explaining $76\%$ of the variance in the observed benchmark scores (\cref{fig:comparing_variance_explained}~(a)). Across all $18$ other out-of-fold experiments, this first principal component variance is never lower than $73\%$. In the structured capabilities model, the factor loadings are more evenly spread. The largest three factors together explain $84\%$ of the variance (\cref{fig:comparing_variance_explained}~(b)).

\clearpage

%% file: sections/5_discussion.tex
\section{Discussion}

\subsection{Principal findings}

My results highlight the shortcomings of existing models for LLM capabilities. Both latent factor models and observational scaling laws mistake LLM scale for a capability, and this mistake limits their explanatory and predictive power. In comparison, the novel structured capabilities model outperforms both alternatives, providing a more parsimonious fit than latent factor models and better predictive generalisation than observational scaling laws.

\subsection{Interpretation of results}

\paragraph{Latent factor models and observational scaling laws conflate scale with capabilities.} In two experiments, latent factor models and observational scaling laws both fit a dominant factor or principal component that correlates with LLMs' $\log$-parameter size. In experiment A, the pure latent factor model estimates a single dominant factor explaining $72\%$ of the communal variance across factor loadings. This factor bears a logistic relationship to the $\log$-parameter size, with $R^2 = 0.474$—a larger correlation than any other estimated factor. This correlation appears moderate, but it is close to the highest correlations between $\log$-parameter size and individual benchmark performance I observed in \cref{fig:subtask_performances_logistic_fit}, which are the basis for the effectiveness of \citet{owen2024predictable}'s scaling laws. In experiment B, the observational scaling law model consistently finds the first principal component to explain over $73\%$ of the total variance, and in one typical model, this principal component also bears the strongest logistic relationship to the $\log$-parameter size ($R^2 = 0.528$). In neither case is the model given explicit information about the parameter sizes of the LLM sample: they are only given the observed benchmark scores. However, the dynamics behind each model's parameter estimation lead them both to discover a proxy for model size. For the latent factor model, I suspect that likelihood maximisation discovers the strong positive correlation between benchmark scores and parameter size and consolidates this information in a single factor to improve the fit. For the observational scaling law model, PCA's variance-maximising decomposition method leads to the model capturing most of the variance due to parameter size in the largest principal component.

\paragraph{The structured capabilities model outperforms both alternatives.} My experimental results show that the two previous models of LLM capabilities are more than just theoretically inelegant—they are statistically suboptimal on the metrics that their designers care about. In experiment A, replacing the standard linear latent factor model with a logistic model improves the absolute fit indices, and adding a structural parameter for parameter size improves the Akaike and Bayesian information criteria. In simpler terms, the standard latent factor model is both a worse fit and a less parsimonious model of the LLM benchmark data. The same structural regression model that outperforms the latent factor model also beats observational scaling laws at predicting out-of-distribution performance. In experiment B, across $19$ out-of-fold prediction experiments for each of our \texttt{BBH} subtasks, the structured capabilities model beats observational scaling laws on the majority of subtasks and achieves lower average mean-squared errors on both training and test data.

\paragraph{The structured capabilities model captures capabilities with explanatory and predictive power.} Beyond objective improvements to model fit and predictive performance, the structured capabilities model gives an interpretable picture of how the capabilities behind \texttt{BBH} subtask performance emerge with scale. I consider these to be \textit{construct valid capability estimates} as far as we can estimate these from \texttt{BBH} benchmark tasks alone.

\cref{fig:factor_comparisons_4} shows the factor loadings and factor scores for $5$ capabilities across the \texttt{BBH} subtasks. There are three major factors that explain descending proportions of variance. These factors all correlate with parameter size, but they do so according to logistic curves with different difficulty and discrimination parameters. Specifically, each next factor shows a steeper curve that begins increasing at a larger parameter size than the previous factor. These factors also load on thematically-related clusters of \texttt{BBH} subtasks. The first capability, the quickest to increase with model scale, loads strongest on tasks involving simple enumeration (Object counting, Penguins in a table), simple formula parsing (Date understanding, Boolean expressions), and sentiment detection (Movie recommendation). The second capability arises later with scale, and loads moderately on tasks that require reasoning steps (Logical deduction, Formal fallacies) and strongly on tasks that seem to require some semantic intuition, like sarcasm or common-sense causal reasoning (Snarks, Ruin names, Causal judgement). The third capability is mostly present in the largest models, and loads on tasks requiring multiple steps of reasoning and state-tracking (Temporal sequences, Tracking shuffled objects).

These capabilities are able to categorise and predict how performance will translate to novel tasks. For example, the Geometric shapes task appears to require some formulaic parsing, as well as some spatial reasoning and common-sense understanding. One expects that the first two factors affect performance on this task, and indeed, these capabilities appear relevant for predicting the held-out Geometric shapes scores in \cref{fig:prediction_experiment_structural_regression}~(a).

\subsection{Relation to other studies}

My results comment directly on previous studies \citet{burnellRevealingStructureLanguage2023} and \citet{ilicEvidenceInterrelatedCognitivelike2024} that used latent factor models to estimate LLM capabilities. I replicated \citeauthor{burnellRevealingStructureLanguage2023}'s exploratory factor analysis method on the \texttt{BBH} benchmark dataset. I find their technique underperforms the structured capabilities model in terms of both model fit and parsimony. \citeauthor{burnellRevealingStructureLanguage2023}'s model achieves acceptable fit on our dataset. Their own reported model fit is poor, though I expect the improved result is due to the much larger sample size—$4,395$ in place of $29$ models. However, an analysis of our replicated model reveals the same deficiencies I hypothesised: the pure latent factor model fits a singular factor with a large proportion of variance explained, a high loading across all benchmarks, and a strong correlation with model size. In the structured capabilities model, the explicit inclusion of parameter size as a variable shows this factor to be illusory. This point is most significant when comparing these findings to \citet{ilicEvidenceInterrelatedCognitivelike2024}. A central claim of \citeauthor{ilicEvidenceInterrelatedCognitivelike2024}'s study is the identification of an ``Artificial General Ability'' factor loading on all indicator variables. My results suggest that this general factor is best explained as a proxy for model scale, and that contrary to \citeauthor{ilicEvidenceInterrelatedCognitivelike2024}'s finding, modelling a latent general capability actually harms the explanatory power of the model.

Compared to observational scaling laws, my results show that allowing for measurement error—which \emph{lowers} the variance explained by the capability estimates—can result in capabilities that are more interpretable and that generalise better to other tasks. This result should be taken only tenuously as a critique of observational scaling laws for their purpose as scaling laws for particular model families. In \citet{ruan2024observational}, a highlight of their approach is that they can fit more precise scaling laws for model families—extrapolating performance for larger versions of the same model—using a shared, low-dimensional capability space. However, I do show that \citeauthor{ruan2024observational}'s choice to use PCA for extracting these capabilities is flawed, and that a switch to latent factor estimates could improve their model's predictive generalisation.

This thesis also offers a concrete direction to answer calls made by AI position papers like \citet{wallachEvaluatingGenerativeAI2024}, \citet{salaudeen2025measurementmeaningvaliditycenteredframework}, and others. These papers advocate for the LLM research community to prioritise the construct validity of LLM benchmarks. Many suggestions for improving construct validity—limiting contamination through canary strings, or reporting standard errors, for example—target specific problems in LLM benchmarking outside the scope of capability modelling. Yet core aspects of construct validity, like convergent and discriminant validity, are best justified through a quantitative approach like the one I consider in this thesis.

\subsection{Limitations and future work}\label{sec:limitations}

This experimental setup has several limitations that I must discuss in context with the results.

First, I only estimate latent factors through exploratory factor analysis—I do not conduct a confirmatory factor analysis. This choice places notable limitations on the interpretability of my identified capabilities. In exploratory factor analysis, beyond deciding on the number of factors with Horn's parallel test (\cref{sec:data}), my factors are free to estimate any relationships supported by the data. \citet{ilicEvidenceInterrelatedCognitivelike2024}'s experiment instead uses confirmatory factor analysis, and this approach allows for a stricter kind of hypothesis testing that incorporates real theories about cognitive capabilities. If I want my latent factors to represent capabilities in a more theory-driven fashion, future work should move from an exploratory to a confirmatory approach following the best social scientific practices~\citep{simms2007construct}.

My modelling approach also presents limitations for the causal interpretation of my results. In some parts of this thesis I reference LLM capabilities `causing' the benchmark scores observed in the OpenLLM Leaderboard data. These scare quotes must stay in place—there is no rigorous basis on which I can claim causality from my results. My structured capabilities model is a purely observational model. Future studies could move beyond this limitation with interventional techniques, like mechanistic interpretability techniques for affecting the neural circuits linked to capabilities \citep{mayne2024can}.

My findings are also limited in scope by the choice to use only BIG-Bench Hard (\texttt{BBH}) questions. These questions share particular formatting constraints, like their use of multiple choice scoring, which can limit the experiments' relevance to other benchmarking domains, like open-ended response questions. Research shows that these different response formats can produce different interpretations of LLM performance~\citep{burnell_rethink_2023, schaeffer_are_2023, molfese-etal-2025-right}. BIG-Bench Hard subtasks also present quite contrived scenarios. These tasks test abstract skills, but they certainly fall short in terms of their resemblance to real-world scenarios—an aspect called \textit{ecological validity} in the literature on construct validity~\citep{schmuckler2001ecological}. For another relevant domain limitation, I should note that BIG-Bench Hard subtasks do nothing to assess the safety and alignment of AI systems to user intentions, another critical domain of LLM evaluation~\citep{reuel2025open}. Future work could extend these model comparisons to a more diverse range of benchmarks, including benchmarks assessing safety capabilities, where the need to understand the origins and the gaps in LLM performance is much more urgent.

\subsection{Implications}

In this thesis, I demonstrate a statistical model that can extract interpretable and generalisable capabilities from LLM benchmark results. This contribution has implications for anyone interested in solving the LLM evaluation crisis. My new method provides the first viable option for quantifying construct validity for LLM evaluations in a reputable fashion.

Given the fluency with which modern LLMs produce natural language, and the rich cognitive science theories we possess around `intelligence' and `reasoning' in humans, we may be tempted to apply our cognitive theories to LLMs in a kind of `top-down' fashion. We might think that a mathematics benchmark measures a `mathematical reasoning' capability in LLMs because we would recognise this proficiency in humans who performed well on the benchmark. Yet LLMs and humans are different cognitive systems. Humans are not trained on data in a way that risks contamination, and humans do not have scaling laws that predict their performance. For these reasons and others, a `top-down' approach to LLM capabilities is misguided. We cannot expect to port entire systems from cognitive science to LLMs, like the Cattell-Horn-Carroll model of cognitive capabilities~\citep{cattellHorn1978, carrollPsychometricsIntelligencePublic1997, schneiderCattellHornCarroll2018}, without distorting these models in the process. This limitation applies also to our measurement tools. We cannot expect that latent factor models apply directly to LLM benchmarks in the way that they apply to human standardised tests and personality surveys. Allowances should be made for the unique challenges posed by LLM evaluation. My experiments provide a modelling approach that incorporates one unique element of LLM evaluation—the impact of scaling laws—and shows the penalty of ignoring these allowances: existing models are misfit and they generalise worse to new benchmarks.

For researchers at the intersection of cognitive science and AI, my methods point towards a more constructive, `bottom-up' approach to LLM capabilities and the problem of construct validity in LLM evaluations. Given their predictive success, we know that we cannot ignore neural scaling laws. At the same time, we should retain the benefits that measurement theory affords models in the social sciences. The combination of these two, in the form of a structured capabilities model, provides a method that should be familiar enough to both cognitive scientists and computer scientists to provide a basis for formal models of construct validity. I advocate that future experiments answering the calls of \citet{wallachEvaluatingGenerativeAI2024} and \citet{salaudeen2025measurementmeaningvaliditycenteredframework} utilise similar approaches when trying to demonstrate construct validity in LLM benchmarks. Further, I hope that future benchmark designers take construct validity more seriously, and that they are inspired by the provided formal toolkit to design evaluations with better validity guarantees. Studies like \citet{polo_tinybenchmarks_2024} show the potential benefits of successful social science techniques applied to benchmarks: we can run smaller evaluations with better statistical guarantees, and we can provide stronger safety and alignment guarantees for frontier systems in healthcare, education, and judicial settings.

For AI policymakers, these experiments highlight the importance of cross-disciplinary collaboration when it comes to understanding large language models. From both a cognitive science and a computer science perspective, these new AI systems are unprecedented and exciting. As \citet{ganguli_predictability_2022} argue, their surprising behaviours should be cause for policy and governance concerns. If siloed approaches to understanding their capabilities will be suboptimal, it is essential that the scientific community's recommendations involve all relevant disciplines. My aim in this thesis is to provide one small example of what a cross-disciplinary approach can achieve. I hope to provoke larger and more ambitious crossovers for understanding artificial intelligence.

\clearpage

%% file: sections/6_conclusion.tex
\section{Conclusion}

LLM evaluation faces an ongoing crisis. To trust our evaluations—whether to understand LLM capabilities, or to predict how benchmarks will translate to the real world—we need formal models that identify general capabilities from our benchmark scores. This thesis presents the structured capabilities model, first model capable of extracting interpretable and generalisable capabilities.
Two experiments demonstrate the model's improvements over existing alternatives from both social science and computer science.

I sample the largest publicly available population of LLMs, $4,395$ models uploaded to the OpenLLM Leaderboard, and collect performances on $19$ subtasks from the \texttt{BBH} dataset. I use this collection of benchmark scores to fit three formal models of LLM capabilities. Two models represent the latest developments in separate subdisciplines of AI research—latent factor models, inspired by cognitive science techniques, and observational scaling laws, inspired by computer science techniques. The third model is the structured capabilities model, my novel hybrid of the previous two.

In one experiment, the structured capabilities model outperforms a latent factor model in terms of both model fit and parsimony. In a second experiment, the structured capabilities model outperforms an observational scaling law model at predicting out-of-distribution subtask accuracies. Put together, I argue that combining these two techniques improves our explanatory and predictive power when estimating LLM capabilities.

The AI research community believes that construct validity is essential for escaping the evaluation crisis. To ensure fair and effective AI governance, policymakers need clear descriptions of LLM capabilities from AI researchers. AI researchers need trustworthy measurements of these capabilities, and for that, they need access to formal methods that clarify the capabilities behind benchmark performances. My approach provides evidence that the best model for such a task combines insights across the social science and computer science disciplines.

%% file: appendices/benchmarks.tex
\section{List of mentioned benchmarks}\label{app:benchmarks}

In some places in the text, it aids readability to enumerate benchmark names without giving their citations. We style all benchmark names in \texttt{monospace} to make their mentions visually apparent. \cref{tab:mentioned_benchmarks} gives a list of all benchmarks mentioned in this paper along with their descriptions and citations.

\begin{center}
\begin{footnotesize}
\begin{longtable}{p{4.5cm}p{5.5cm}p{3.5cm}}
        \toprule
        \textbf{Benchmark} & \textbf{Dataset description} & \bf Citation \\\midrule
        \texttt{GSM8K} & Grade school-level arithmetic word problems. & \citet{cobbe_training_2021} \\
        \texttt{MATH} & Challenging high school-level competition mathematics problems.  & \citet{hendrycksMATH} \\
        \texttt{FrontierMath} & Unpublished, expert-level research mathematics problems. & \citet{glazer2024frontiermathbenchmarkevaluatingadvanced} \\
        \texttt{GPQA} & ``Google-Proof Q\&A.'' PhD-level biology, physics, and chemistry multiple choice questions. & \citet{rein2024gpqa} \\
        \texttt{MMLU-Pro} & Broad knowledge and reasoning benchmark with questions spanning 14 academic subjects. & \citet{wang2025mmlu} \\
        \texttt{Commonsense QA} & Crowd-sourced multiple choice questions targeting simple concepts. & \citet{talmor_commonsenseqa_2019} \\
        \texttt{Winogrande} & Adversarially-sourced ambiguous pronoun resolution problems. & \citet{sakaguchi_winogrande_2021} \\
        \texttt{HumanEval} & Docstring descriptions for generating functionally-correct software programs. & \citet{chen2021evaluatinglargelanguagemodels} \\
        \texttt{SWE-Bench} & GitHub issues requiring changes to python codebases. & \citet{jimenez2024swebench} \\
        \texttt{SuperGLUE} & General-purpose language understanding sourced from 8 NLP tasks. & \citet{wangSuperglueStickierBenchmark2019a} \\
        \texttt{Humanity's Last Exam} & Difficult multi-modal problems across many academic subjects. & \citet{humanitys_last_exam} \\
        \texttt{AIME 2024} & Questions from American invitational mathematics competiton. & \citep{mathematicalassociationofamericaAmericanInvitationalMathematics2024}\\
        \texttt{MathBench} & Broadly scoped mathematics questions in English and Chinese. & \citep{liu-etal-2024-mathbench} \\
        \texttt{HELM} & ``Holistic Evaluation of Language Models.'' & \citet{liang_holistic_2023} \\
        \texttt{BIG-bench} & ``Beyond the Imitation Game.'' Crowd-sourced collection of $204$ distinct tasks. & \citet{srivastava2023beyond} \\
        \texttt{TruthfulQA} & Q\&A targeting common misconceptions and falsehoods. & \citet{linTruthfulQAMeasuringHow2022} \\
        \texttt{BBQ} & ``Bias Benchmark for QA.'' Questions where one of nine social biases override the correct choice. & \citet{parrish-etal-2022-bbq} \\
        \texttt{HellaSwag} & Sentence completions testing common natural language inference. & \citet{zellers-etal-2019-hellaswag} \\
        \texttt{WikiFact} & Wikipedia content to test factual accuracy and relation extraction. & \citet{goodrich2019assessing} \\
        \texttt{NaturalQuestions} & Q\&A adapted from Google search queries. & \citet{kwiatkowski2019natural} \\
        \texttt{XSUM} & News summarisation using online articles from the BBC. & \citet{narayan-etal-2018-dont} \\
        \texttt{OpenbookQA} & Q\&A using elementary science facts. & \citet{mihaylov-etal-2018-suit} \\
        \texttt{bAbI} & Short stories about characters, followed by reading comprehension questions. & \citet{weston2015towards} \\
        \texttt{Dyck} & Completing Dyck sentences: properly nested sequences of characters \texttt{\{\}[]()}. & \citet{suzgun2019memory} \\
        \texttt{BBH} & ``BIG-Bench Hard.'' $23$ difficult tasks selected from \texttt{BIG-bench}. & \citet{BBH} \\
        \texttt{IFEval} & ``Instruction-Following Evaluation.'' Prompts with formatting requirements, word length, and other verifiable instructions. & \citet{zhou2023instructionfollowingevaluationlargelanguage} \\
        \texttt{MuSR} & ``Multistep Soft Reasoning.'' \textasciitilde$1,000$-word passages, like murder mysteries, testing reading comprehension. & \citet{sprague2024musr} \\
        \bottomrule
    \caption[All mentioned benchmark datasets, with citations]{\textbf{All mentioned benchmark datasets, with citations}.}
    \label{tab:mentioned_benchmarks}
\end{longtable}
\end{footnotesize}
\end{center}

%% file: appendices/mathematical_notation.tex
\section{Default mathematical notation}\label{app:math_notation}

We utilise the following mathematical notation throughout this work. This table of notation is copied from the ICLR 2024 submission guidelines, and appears originally in the textbook \textit{Deep Learning} \citep{goodfellow2016deep}, available at \url{https://github.com/goodfeli/dlbook_notation/}.

\centerline{\bf Numbers and Arrays}
\bgroup
\def\arraystretch{1.5}
\begin{tabular}{p{1in}p{3.25in}}
$\displaystyle a$ & A scalar (integer or real)\\
$\displaystyle \va$ & A vector\\
$\displaystyle \mA$ & A matrix\\
$\displaystyle \tA$ & A tensor\\
$\displaystyle \mI_n$ & Identity matrix with $n$ rows and $n$ columns\\
$\displaystyle \mI$ & Identity matrix with dimensionality implied by context\\
$\displaystyle \ve^{(i)}$ & Standard basis vector $[0,\dots,0,1,0,\dots,0]$ with a 1 at position $i$\\
$\displaystyle \text{diag}(\va)$ & A square, diagonal matrix with diagonal entries given by $\va$\\
$\displaystyle \ra$ & A scalar random variable\\
$\displaystyle \rva$ & A vector-valued random variable\\
$\displaystyle \rmA$ & A matrix-valued random variable\\
\end{tabular}
\egroup
\vspace{0.25cm}

\centerline{\bf Sets and Graphs}
\bgroup
\def\arraystretch{1.5}

\begin{tabular}{p{1.25in}p{3.25in}}
$\displaystyle \sA$ & A set\\
$\displaystyle \R$ & The set of real numbers \\
$\displaystyle \{0, 1\}$ & The set containing 0 and 1 \\
$\displaystyle \{0, 1, \dots, n \}$ & The set of all integers between $0$ and $n$\\
$\displaystyle [a, b]$ & The real interval including $a$ and $b$\\
$\displaystyle (a, b]$ & The real interval excluding $a$ but including $b$\\
$\displaystyle \sA \backslash \sB$ & Set subtraction, i.e., the set containing the elements of $\sA$ that are not in $\sB$\\
$\displaystyle \gG$ & A graph\\
$\displaystyle \parents_\gG(\ervx_i)$ & The parents of $\ervx_i$ in $\gG$
\end{tabular}
\vspace{0.25cm}

\centerline{\bf Indexing}
\bgroup
\def\arraystretch{1.5}

\begin{tabular}{p{1.25in}p{3.25in}}
$\displaystyle \eva_i$ & Element $i$ of vector $\va$, with indexing starting at 1 \\
$\displaystyle \eva_{-i}$ & All elements of vector $\va$ except for element $i$ \\
$\displaystyle \emA_{i,j}$ & Element $i, j$ of matrix $\mA$ \\
$\displaystyle \mA_{i, :}$ & Row $i$ of matrix $\mA$ \\
$\displaystyle \mA_{:, i}$ & Column $i$ of matrix $\mA$ \\
$\displaystyle \etA_{i, j, k}$ & Element $(i, j, k)$ of a 3-D tensor $\tA$\\
$\displaystyle \tA_{:, :, i}$ & 2-D slice of a 3-D tensor\\
$\displaystyle \erva_i$ & Element $i$ of the random vector $\rva$ \\
\end{tabular}
\egroup
\vspace{0.25cm}

\centerline{\bf Calculus}
\bgroup
\def\arraystretch{1.5}
\begin{tabular}{p{1.25in}p{3.25in}}
$\displaystyle\frac{d y} {d x}$ & Derivative of $y$ with respect to $x$\\ [2ex]
$\displaystyle \frac{\partial y} {\partial x} $ & Partial derivative of $y$ with respect to $x$ \\
$\displaystyle \nabla_\vx y $ & Gradient of $y$ with respect to $\vx$ \\
$\displaystyle \nabla_\mX y $ & Matrix derivatives of $y$ with respect to $\mX$ \\
$\displaystyle \nabla_\tX y $ & Tensor containing derivatives of $y$ with respect to $\tX$ \\
$\displaystyle \frac{\partial f}{\partial \vx} $ & Jacobian matrix $\mJ \in \R^{m\times n}$ of $f: \R^n \rightarrow \R^m$\\
$\displaystyle \nabla_\vx^2 f(\vx)\text{ or }\mH( f)(\vx)$ & The Hessian matrix of $f$ at input point $\vx$\\
$\displaystyle \int f(\vx) d\vx $ & Definite integral over the entire domain of $\vx$ \\
$\displaystyle \int_\sS f(\vx) d\vx$ & Definite integral with respect to $\vx$ over the set $\sS$ \\
\end{tabular}
\egroup
\vspace{0.25cm}

\centerline{\bf Probability and Information Theory}
\bgroup
\def\arraystretch{1.5}
\begin{tabular}{p{1.25in}p{3.25in}}
$\displaystyle P(\ra)$ & A probability distribution over a discrete variable\\
$\displaystyle p(\ra)$ & A probability distribution over a continuous variable, or over
a variable whose type has not been specified\\
$\displaystyle \ra \sim P$ & Random variable $\ra$ has distribution $P$\\
$\displaystyle  \E_{\rx\sim P} [ f(x) ]\text{ or } \E f(x)$ & Expectation of $f(x)$ with respect to $P(\rx)$ \\
$\displaystyle \Var(f(x)) $ &  Variance of $f(x)$ under $P(\rx)$ \\
$\displaystyle \Cov(f(x),g(x)) $ & Covariance of $f(x)$ and $g(x)$ under $P(\rx)$\\
$\displaystyle H(\rx) $ & Shannon entropy of the random variable $\rx$\\
$\displaystyle \KL ( P \Vert Q ) $ & Kullback-Leibler divergence of P and Q \\
$\displaystyle \mathcal{N} ( \vx ; \vmu , \mSigma)$ & Gaussian distribution %
over $\vx$ with mean $\vmu$ and covariance $\mSigma$ \\
\end{tabular}
\egroup
\vspace{0.25cm}

\centerline{\bf Functions}
\bgroup
\def\arraystretch{1.5}
\begin{tabular}{p{1.25in}p{3.25in}}
$\displaystyle f: \sA \rightarrow \sB$ & The function $f$ with domain $\sA$ and range $\sB$\\
$\displaystyle f \circ g $ & Composition of the functions $f$ and $g$ \\
  $\displaystyle f(\vx ; \vtheta) $ & A function of $\vx$ parametrized by $\vtheta$.
  (Sometimes we write $f(\vx)$ and omit the argument $\vtheta$ to lighten notation) \\
$\displaystyle \log x$ & Natural logarithm of $x$ \\
$\displaystyle \sigma(x)$ & Logistic sigmoid, $\displaystyle \frac{1} {1 + \exp(-x)}$ \\
$\displaystyle \zeta(x)$ & Softplus, $\log(1 + \exp(x))$ \\
$\displaystyle || \vx ||_p $ & $\normlp$ norm of $\vx$ \\
$\displaystyle || \vx || $ & $\normltwo$ norm of $\vx$ \\
$\displaystyle x^+$ & Positive part of $x$, i.e., $\max(0,x)$\\
$\displaystyle \1_\mathrm{condition}$ & is 1 if the condition is true, 0 otherwise\\
\end{tabular}
\egroup
\vspace{0.25cm}

%% file: appendices/bbh_task_descriptions.tex
\section{BIG-Bench Hard task descriptions}\label{app:bbh_tasks}

In the following list, we provide the task descriptions as given in the original BIG-Bench Hard paper by~\citet{BBH}. Our list below is a subset of the list in Appendix A of~\citet{BBH}. Each example is the first example in the OpenLLM Leaderboard dataset for that subtask.

\begin{enumerate}
    \item \textbf{Boolean Expressions}: Evaluate the truth value of a random Boolean expression consisting of Boolean constants (\texttt{True}, \texttt{False}) and basic Boolean operators (\texttt{and}, \texttt{or}, and \texttt{not}).

    \begin{quote}
        \textbf{Example}: Evaluate the result of a random Boolean expression.

        \vspace{1em}

        Q: not ( ( not not True ) ) is

        \vspace{1em}

        A: False
    \end{quote}
        
    \item \textbf{Causal Judgement}: Given a short story (involving moral, intentional, or counterfactual analysis), determine how a typical person would answer a causal question about the story.

    \begin{quote}
        \textbf{Example}: Answer questions about causal attribution.

        \vspace{1em}
        
        Q: How would a typical person answer each of the following questions about causation?
        
        Frank T., had an ongoing dispute with his neighbor over a stretch of land and one day decided to shoot his neighbor in the body. Frank T. had no experience with guns, his hand slipped on the barrel of the gun, and the shot went wild. Nonetheless, the bullet bounced off a large boulder several feet away and hit the neighbor's body, causing significant injury. Did Frank T. intentionally shoot his neighbor in the body?
        
        Options:
        
        - Yes
        
        - No

        \vspace{1em}
        
        A: No
    \end{quote}

    \item \textbf{Date Understanding}: Given a small set of sentences about a particular date, answer the provided question (e.g., “The concert was scheduled to be on 06/01/1943, but was delayed by one day to today. What is the date yesterday in \texttt{MM/DD/YYYY}?”).

    \begin{quote}
        \textbf{Example}: Infer the date from context.

        \vspace{1em}
        
        Q: Today is Christmas Eve of 1937. What is the date 10 days ago in MM/DD/YYYY?
        
        Options:
        
        (A) 12/14/2026
        
        (B) 12/14/1950
        
        (C) 12/14/2007
        
        (D) 12/14/1937
        
        (E) 07/14/1938
        
        (F) 12/14/1988

        \vspace{1em}
        
        A: (D)
    \end{quote}
    
    \item \textbf{Disambiguation QA}: Given a sentence with an “ambiguous” pronoun, either determine whether the sentence is inherently ambiguous (i.e., the thing that the pronoun refers to cannot be inferred by given information) or, if the pronoun can be implicitly deduced, state the antecedent of the pronoun (i.e., the noun to which the pronoun refers).

    \begin{quote}
        \textbf{Example}: Clarify the meaning of sentences with ambiguous pronouns.
        
        \vspace{1em}
        
        Q: In the following sentences, explain the antecedent of the pronoun (which thing the pronoun refers to), or state that it is ambiguous.
        
        Sentence: The chief told the counselor that they took the day off.
        
        Options:
        
        (A) The chief took the day off
        
        (B) The counselor took the day off
        
        (C) Ambiguous

        \vspace{1em}
        
        A: (A)
    \end{quote}

    \item \textbf{Formal Fallacies}: Given a context involving a set of statements (generated by one of the argument schemes), determine whether an argument—presented informally—can be logically deduced from the provided context.

    \begin{quote}
        \textbf{Example}: Distinguish deductively valid arguments from formal fallacies.

        \vspace{1em}
        
        Q: ``It is not always easy to see who is related to whom -- and in which ways. The following argument pertains to this question: To begin with, Lesley is a close friend of Fernando. Moreover, being a close friend of Fernando or a schoolmate of Lowell is sufficient for being a great-grandfather of Leroy. It follows that Lesley is a great-grandfather of Leroy.''
        
        Is the argument, given the explicitly stated premises, deductively valid or invalid?
        
        Options:
        
        - valid
        
        - invalid

        \vspace{1em}
        
        A: valid
    \end{quote}

    \item \textbf{Geometric Shapes}: Given a full SVG path element containing multiple commands, determine the geometric shape that would be generated if one were to execute the full path element.

    \begin{quote}
        \textbf{Example}: Name geometric shapes from their SVG paths.

        \vspace{1em}
        
        Q: This SVG path element <path d=``M 31.00,73.00 L 32.00,59.00 L 44.00,50.00 L 49.00,41.00 L 64.00,37.00 L 71.00,55.00 L 64.00,76.00 L 52.00,61.00 L 31.00,73.00''/> draws a\
        
        Options:
        
        (A) circle
        
        (B) heptagon
        
        (C) hexagon
        
        (D) kite
        
        (E) line
        
        (F) octagon
        
        (G) pentagon
        
        (H) rectangle
        
        (I) sector
        
        (J) triangle

        \vspace{1em}
        
        A: (F)
    \end{quote}
    
    \item \textbf{Hyperbaton}: Given two English-language sentences, determine the one with the correct adjective order.

    \begin{quote}
        \textbf{Example}: Order adjectives correctly in English sentences.
        
        \vspace{1em}
        
        Q: Which sentence has the correct adjective order:
        
        Options:
        
        (A) rubber terrible ship
        
        (B) terrible rubber ship

        \vspace{1em}
        
        A: (B)
    \end{quote}
    
    \item \textbf{Logical Deduction}: Deduce the order of a sequence of objects based on the clues and information about their spacial relationships and placements.

    \begin{quote}
        \textbf{Example}: A logical deduction task which requires deducing the order of a sequence of objects.
        
        \vspace{1em}
        
        Q: The following paragraphs each describe a set of three objects arranged in a fixed order. The statements are logically consistent within each paragraph. In a golf tournament, there were three golfers: Amy, Eli, and Eve. Eve finished above Amy. Eli finished below Amy.
        
        Options:
        
        (A) Amy finished last
        
        (B) Eli finished last
        
        (C) Eve finished last

        \vspace{1em}
        
        A: (B)
    \end{quote}
    
    \item \textbf{Movie Recommendation}: Given a list of movies a user might have watched and liked, recommend a new, relevant movie to the user out of the four potential choices user might have.

    \begin{quote}
        \textbf{Example}: Recommend movies similar to the given list of movies.

        \vspace{1em}
        
        Q: Find a movie similar to Star Wars Episode IV - A New Hope, Indiana Jones and the Last Crusade, Star Wars Episode V - The Empire Strikes Back, The Big Lebowski:
        
        Options:
        
        (A) Tetsuo
        
        (B) the Ironman
        
        (C) The Princess Bride
        
        (D) The Barkley Marathons The Race That Eats Its Young
        
        (E) Bug

        \vspace{1em}
    
        A: (C)
    \end{quote}
    
    \item \textbf{Navigate}: Given a series of navigation steps to an agent, determine whether the agent would end up back at its initial starting point.

    \begin{quote}
        \textbf{Example}: Given a series of navigation instructions, determine whether one would end up back at the starting point.
        
        \vspace{1em}
        
        Q: If you follow these instructions, do you return to the starting point? Turn left. Turn around. Turn left. Take 7 steps. Take 2 steps. Take 4 steps. Take 8 steps.
        
        Options:
        
        - Yes
        
        - No

        \vspace{1em}
        
        A: No
    \end{quote}
    
    \item \textbf{Object Counting}: Given a collection of possessions that a person has along with their quantities (e.g., three pianos, two strawberries, one table, and two watermelons), determine the number of a certain object/item class (e.g., fruits).

    \begin{quote}
        \textbf{Example}: Questions that involve enumerating objects and asking the model to count them.
        
        \vspace{1em}
        
        Q: I have a blackberry, a clarinet, a nectarine, a plum, a strawberry, a banana, a flute, an orange, and a violin. How many fruits do I have?

        \vspace{1em}
        
        A: 6
    \end{quote}

    \item \textbf{Penguins in a Table}: Given a unique table of penguins (and sometimes some new information), answer a question about the attributes of the penguins.

    \begin{quote}
        \textbf{Example}: Answer questions about a table of penguins and their attributes.
        
        \vspace{1em}
        
        Q: Here is a table where the first line is a header and each subsequent line is a penguin: name, age, height (cm), weight (kg) Louis, 7, 50, 11 Bernard, 5, 80, 13 Vincent, 9, 60, 11 Gwen, 8, 70, 15 For example: the age of Louis is 7, the weight of Gwen is 15 kg, the height of Bernard is 80 cm. We now add a penguin to the table:
        
        James, 12, 90, 12
        
        How many penguins are less than 8 years old?
        
        Options:
        
        (A) 1
        
        (B) 2
        
        (C) 3
        
        (D) 4
        
        (E) 5

        \vspace{1em}
        
        A: (B)
    \end{quote}

    \item \textbf{Reasoning about Colored Objects}: Given a context, answer a simple question about the color of an object on a surface.

    \begin{quote}
        \textbf{Example}: Answer extremely simple questions about the colors of objects on a surface.

        \vspace{1em}
        
        Q: On the nightstand, there is a red pencil, a purple mug, a burgundy keychain, a fuchsia teddy bear, a black plate, and a blue stress ball. What color is the stress ball?
        
        Options:
        
        (A) red
        
        (B) orange
        
        (C) yellow
        
        (D) green
        
        (E) blue
        
        (F) brown
        
        (G) magenta
        
        (H) fuchsia
        
        (I) mauve
        
        (J) teal
        
        (K) turquoise
        
        (L) burgundy
        
        (M) silver
        
        (N) gold
        
        (O) black
        
        (P) grey
        
        (Q) purple
        
        (R) pink

        \vspace{1em}
        
        A: (E)
    \end{quote}
    
    \item \textbf{Ruin Names}: Given an artist, band, or movie name, identify a one-character edit to the name that changes the meaning of the input and makes it humorous.

    \begin{quote}
        \textbf{Example}: Select the humorous edit that `ruins' the input movie or musical artist name.
        
        \vspace{1em}
        
        Q: Which of the following is a humorous edit of this artist or movie name: `whitesnake'?
        
        Options:
        
        (A) whitesnape
        
        (B) whitesnapke
        
        (C) whitesnuake
        
        (D) mwhitesnake

        \vspace{1em}
    
        A: (A)
    \end{quote}

    \item \textbf{Snarks}: Given two nearly-identical sentences, determine which one is sarcastic.

    \begin{quote}
        \textbf{Example}: Determine which of two sentences is sarcastic.
        
        According to Cambridge University Dictionary, sarcasm is ``the use of remarks that clearly mean the opposite of what they say, made in order to hurt someone's feelings or to criticize something in a humorous way.'' Sarcastic sentences often contain satirical or ironic utterances, hyperboles, ambivalent or witty remarks.
        
        \vspace{1em}
        
        Q: Which statement is sarcastic?
        
        Options:
        
        (A) Yes, because having interests and actively researching them is a huge waste
        
        (B) Yes, because having interests and actively researching them is a huge deal

        \vspace{1em}
        
        A: (A)
    \end{quote}

    \item \textbf{Sports Understanding}: Determine whether a factitious sentence related to sports is plausible.

    \begin{quote}
        \textbf{Example}: Determine whether an artificially constructed sentence relating to sports is plausible or not.
        
        \vspace{1em}
    
        Q: Is the following sentence plausible? ``Bam Adebayo scored a reverse layup in the Western Conference Finals.''

        \vspace{1em}
        
        A: yes
    \end{quote}
    
    \item \textbf{Temporal Sequences}: Given a series of events and activities a person has completed in the course of a day, determine what time, during the day, they might have been free to perform another activity.

    \begin{quote}
        \textbf{Example}: Task description: Answer questions about which times certain events could have occurred.
        
        \vspace{1em}
        
        Q: Today, Emily went to the museum. Between what times could they have gone?
        
        We know that:
        
        Emily woke up at 1pm.
        
        Elizabeth saw Emily reading at the library from 2pm to 4pm.
        
        Jessica saw Emily watching a movie at the theater from 4pm to 5pm.
        
        Leslie saw Emily waiting at the airport from 5pm to 6pm.
        
        William saw Emily buying clothes at the mall from 6pm to 7pm.
        
        The museum was closed after 7pm.
        
        Between what times could Emily have gone to the museum?
        
        Options:
        
        (A) 1pm to 2pm
        
        (B) 6pm to 7pm
        
        (C) 5pm to 6pm
        
        (D) 2pm to 4pm

        \vspace{1em}
        
        A: (A)
    \end{quote}
    
    \item \textbf{Tracking Shuffled Objects}: Given the initial positions of a set of objects and a series of transformations (namely, pairwise swaps) applied to them, determine the final positions of the objects.

    \begin{quote}
        \textbf{Example}: A task requiring determining the final positions of a set of objects given their initial positions and a description of a sequence of swaps.

        \vspace{1em}
        
        Q: Alice, Bob, and Claire are playing a game. At the start of the game, they are each holding a ball: Alice has a yellow ball, Bob has a blue ball, and Claire has a pink ball.
        
        As the game progresses, pairs of players trade balls. First, Claire and Alice swap balls. Then, Alice and Bob swap balls. Finally, Claire and Bob swap balls. At the end of the game, Bob has the
        
        Options:
        
        (A) yellow ball
        
        (B) blue ball
        
        (C) pink ball

        \vspace{1em}
    
        A: (A)
    \end{quote}
    
    \item \textbf{Web of Lies}: Evaluate the truth value of a random Boolean function expressed as a natural-language word problem.

    \begin{quote}
        \textbf{Example}: Evaluate a random boolean function expressed as a word problem.

        \vspace{1em}
        
        Q: Question: Fidel tells the truth. Jerry says Fidel tells the truth. Vina says Jerry tells the truth. Millicent says Vina lies. Raymond says Millicent lies. Does Raymond tell the truth?

        \vspace{1em}
        
        A: Yes
    \end{quote}
\end{enumerate}

%% file: appendices/derivation.tex
\section{Transformed matrix derivation}\label{app:derivation}

Beginning with
\begin{equation*}
    \mB_{i, :} = c_i + \frac{1 - c_i}{1 + \exp[-\alpha_i(\mL_{i, :}[\vw_n\log\vn] + \mE_{i, :}-\beta_i)]}
\end{equation*}
we set $\mB'_{i, :} = \mL_{i, :}[\vw_n\log\vn] + \mE_{i, :}$ and solve for this quantity:
\begin{subequations}
    \begin{align*}
        \mB_{i, :} &= c_i + \frac{1 - c_i}{1 + \exp\big(-\alpha_i(\mB'_{i, :}-\beta_i)\big)} \\
        \mB_{i, :} - c_i &= \frac{1 - c_i}{1 + \exp\big(-\alpha_i(\mB'_{i, :}-\beta_i)\big)} \\
        1 + \exp\big(-\alpha_i(\mB'_{i, :}-\beta_i)\big) &= \frac{1 - c_i}{\mB_{i, :} - c_i} \\
        e^{-\alpha_i(\mB'_{i, :}-\beta_i)} &= \frac{1 - c_i}{\mB_{i, :} - c_i} - 1 \\
        -\alpha_i(\mB'_{i, :}-\beta_i) &= \log\bigg[\frac{1 - c_i}{\mB_{i, :} - c_i} - 1\bigg] \\
        \mB'_{i, :}-\beta_i &= -\frac{1}{\alpha_i}\log\bigg[\frac{1 - c_i}{\mB_{i, :} - c_i} - 1\bigg] \\
        \mB'_{i, :} &= \beta_i - \frac{1}{\alpha_i}\log\bigg[\frac{1 - c_i}{\mB_{i, :} - c_i} - 1\bigg] \\
        \mB'_{i, :} &= \beta_i - \frac{1}{\alpha_i}\log\bigg[\frac{1 - c_i}{\mB_{i, :} - c_i} - \frac{\mB_{i, :} - c_i}{\mB_{i, :} - c_i}\bigg] \\
        \mB'_{i, :} &= \beta_i - \frac{1}{\alpha_i}\log\bigg[\frac{1 - c_i - (\mB_{i, :} - c_i)}{\mB_{i, :} - c_i}\bigg] \\
        \mB'_{i, :} &= \beta_i - \frac{1}{\alpha_i}\log\bigg[\frac{1 - \mB_{i, :}}{\mB_{i, :} - c_i}\bigg] \\
        \mB'_{i, :} &= \beta_i + \frac{1}{\alpha_i}\log\bigg[\frac{\mB_{i, :} - c_i}{1 - \mB_{i, :}}\bigg]
    \end{align*}
\end{subequations}

This equivalence means modelling something like the odds-ratio of expected success on individual subtask questions instead of the average score itself using a logistic response.

%% file: appendices/bbh_logistic_correlations.tex
\section{Parameter scaling laws for all \texttt{BBH} subtasks}\label{app:all_logistic_correlations}

\begin{figure}[t]
    \centering
    \includegraphics[width=\linewidth]{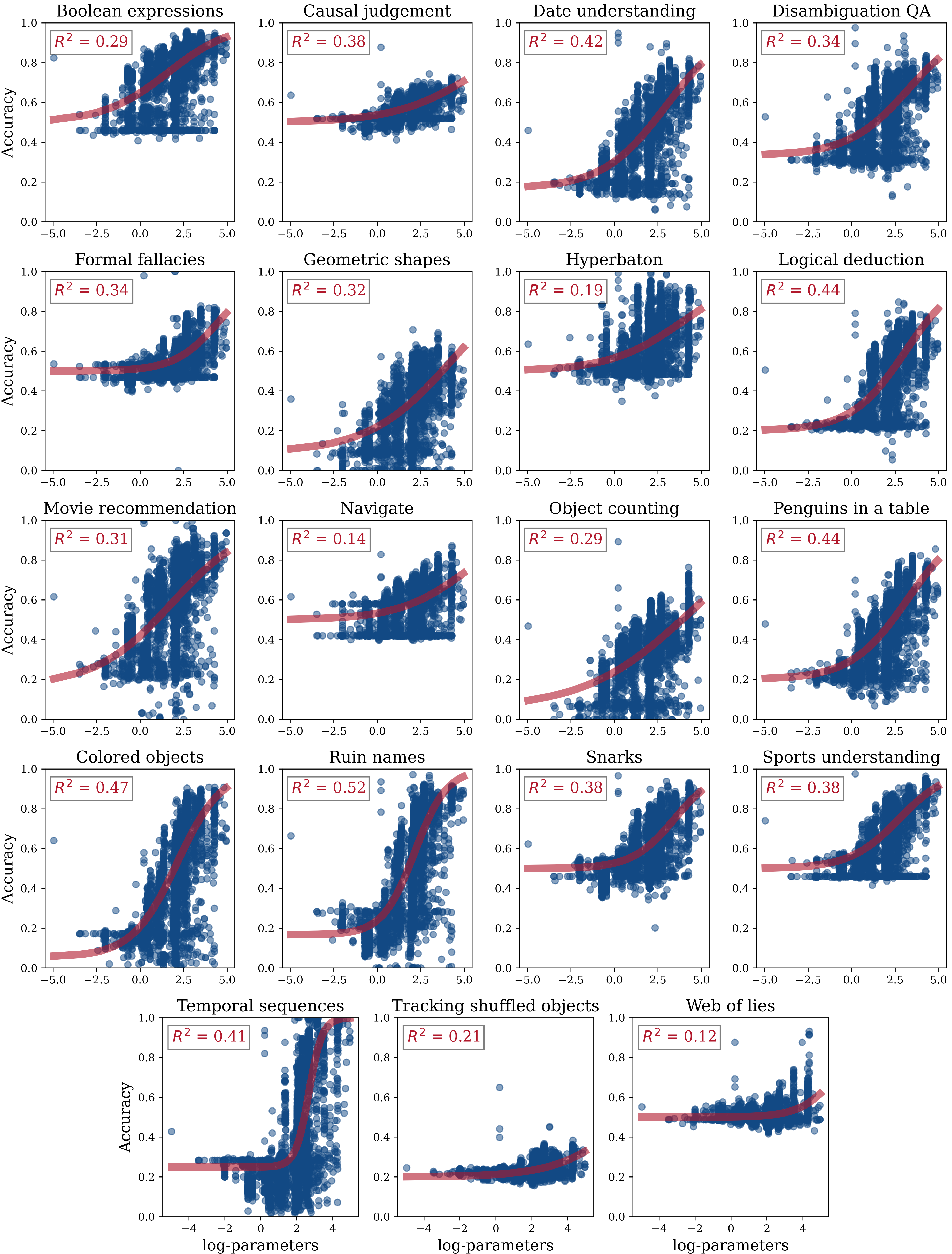}
    \caption[Raw \texttt{BBH} scores against $\log$-parameter size]{\textbf{Raw \texttt{BBH} scores against $\log$-parameter size}. The red line represents a logistic fit using the functional form of \cref{eq:logistic_scaling_law_model}. $R^2$ values range from $0.12$ (Web of lies) to $0.52$ (Ruin names).}
    \label{fig:subtask_performances_logistic_fit}
\end{figure}

\cref{fig:subtask_performances_logistic_fit} shows the result of fitting logistic scaling law model \cref{eq:logistic_scaling_law_model} for all $p=19$ \texttt{BBH} subtasks in my dataset.

